\documentclass{article} % For LaTeX2e
\usepackage{arxiv,times}

\usepackage{amsmath,amsfonts,bm}
\usepackage{amsmath, amsthm, amssymb}
\def\eqref#1{equation~\ref{#1}}
\def\1{\bm{1}}

\def\va{{\bm{a}}}
\def\vb{{\bm{b}}}

\def\vg{{\bm{g}}}

\def\vx{{\bm{x}}}

\def\vz{{\bm{z}}}

\DeclareMathAlphabet{\mathsfit}{\encodingdefault}{\sfdefault}{m}{sl}
\SetMathAlphabet{\mathsfit}{bold}{\encodingdefault}{\sfdefault}{bx}{n}

\usepackage{hyperref}
\usepackage{url}
\usepackage{graphicx}
\usepackage{multirow}
\usepackage{amssymb}
\usepackage{amsmath}
\usepackage{colortbl}
\usepackage{adjustbox}
\usepackage{makecell}
\usepackage{multicol}
\usepackage{booktabs}
\usepackage{bbding}
\usepackage{pifont}
\usepackage{wasysym}
\usepackage{utfsym}
\usepackage{fontawesome}
\usepackage{algorithm}
\usepackage{algorithmic}
\usepackage{colortbl} 
\usepackage{bm}
\usepackage{subcaption}
\usepackage{mathrsfs}
\usepackage{caption}
\usepackage{soul}
\usepackage{color}

\definecolor{mygray}{gray}{0.9}

\title{CLIPure: Purification in Latent Space \\ via CLIP for Adversarially Robust \\ Zero-Shot Classification}
\author{ Mingkun Zhang$^{1,3}$, Keping Bi$^{2,3,}$\thanks{corresponding authors}\ \ , Wei Chen$^{1,3,*}$, Jiafeng Guo$^{2,3}$, Xueqi Cheng$^{1,3}$ \\
  $^{1}$State Key Lab of AI Safety, CAS Key Lab of AI Safety, \\
  \ \ Institute of Computing Technology, Chinese Academy of Sciences, Beijing, China \\
  $^{2}$CAS Key Laboratory of Network Data Science and Technology, \\ 
  \ \ Institute of Computing Technology, Chinese Academy of Sciences, Beijing, China \\
  $^{3}$University of Chinese Academy of Sciences, Beijing, China \\
  \texttt{\{zhangmingkun20z, bikeping, chenwei2022, guojiafeng, cxq\}@ict.ac.cn} \\
}

\newtheorem{theorem}{Theorem}
\iclrfinalcopy % Uncomment for camera-ready version, but NOT for submission.
\begin{document}
 
% \definecolor{mycolor_blue}{RGB}{96,150,230}

\maketitle
\begin{abstract}
In this paper, we aim to build an adversarially robust zero-shot image classifier. We ground our work on CLIP, a vision-language pre-trained encoder model that can perform zero-shot classification by matching an image with text prompts ``a photo of a $<$class-name$>$.''. Purification is the path we choose since it does not require adversarial training on specific attack types and thus can cope with any foreseen attacks. We then formulate purification risk as the KL divergence between the joint distributions of the purification process of denoising the adversarial samples and the attack process of adding perturbations to benign samples, through bidirectional Stochastic Differential Equations (SDEs). The final derived results inspire us to explore purification in the multi-modal latent space of CLIP. We propose two variants for our CLIPure approach: \textit{CLIPure-Diff} which models the likelihood of images' latent vectors with the DiffusionPrior module in DaLLE-2 (modeling the generation process of CLIP's latent vectors), and \textit{CLIPure-Cos} which models the likelihood with the cosine similarity between the embeddings of an image and ``a photo of a.''. As far as we know, CLIPure is the first purification method in multi-modal latent space and CLIPure-Cos is the first purification method that is not based on generative models, which substantially improves defense efficiency. We conducted extensive experiments on CIFAR-10, ImageNet, and 13 datasets that previous CLIP-based defense methods used for evaluating zero-shot classification robustness. Results show that CLIPure boosts the SOTA robustness by a large margin, e.g., from 71.7\% to \textbf{91.1\%} on CIFAR10, from 59.6\% to \textbf{72.6\%} on ImageNet, and \textbf{108\%} relative improvements of average robustness on the 13 datasets over previous SOTA. The code is available at \href{https://github.com/TMLResearchGroup-CAS/CLIPure}{https://github.com/TMLResearchGroup-CAS/CLIPure}.
\end{abstract}

\section{Introduction}

% Neural networks have achieved impressive performance across various tasks but remain vulnerable to imperceptible perturbations, showing vulnerability even in the straightforward image classification task under adversarial attacks \citep{szegedy2013intriguing}.
% Previous efforts to enhance model robustness against adversarial examples mainly focus on two approaches: adversarial training \citep{mkadry2017towards, wang2023better} and purification \citep{song2017pixeldefend, nie2022diffusion}. Since adversarial examples are considered out-of-distribution samples \citep{song2017pixeldefend}, models perform poorly in these regions not encountered during training. Adversarial training aims to mitigate this by incorporating adversarial examples, generated by specific attacks, into the training set. 
% % This method helps improve model performance on these out-of-distribution samples. 
% In contrast, adversarial purification uses generative models to map adversarial examples back to the clean data distribution, where the model performs well.
% added by Keping
% The need for a zero-shot robust classifier. neural classifiers are not robust. zero-shot classifier CLIP is not robust. 

Image classifiers are usually trained in a supervised manner with training data and evaluated on the corresponding test data until recently several vision-language models have emerged as zero-shot classifiers \citep{li2023your,radford2021learning,li2022grounded}. Among them, CLIP \citep{radford2021learning} is an example that is popular, effective, and efficient. CLIP performs zero-shot classification by forming text prompts ``a photo of a $<$class-name$>$.'' of all the candidate categories, and selecting the class with the highest similarity with the image embedding. Despite its efficacy, when facing adversarial attacks, its accuracy can drop to zero, similarly vulnerable to other neural classifiers. 
 
Existing methods to enhance adversarial robustness follow two primary paths: adversarial training and purification. Adversarial Training (AT) \citep{mkadry2017towards, rebuffi2021fixing, wang2023better} incorporates adversarial examples into model training to boost robustness. It often achieves
% Adversarial defense: adversarial training and purification. 
% zero-shot classification or unseen attack. 
% Existing work based on CLIP. FARE and TeCoA
% their pros and cons. 
% Purification: certain types of datasets, worse than AT. 
% CLIP+Purification 
% Purification risk: purification p(x) of benign and adversarial examples. KL()
% smoothiness of log p(x). 
% is pixel space the best choice? 
% latent space (vision-language aligned). 
% We propose: two variants. 
% four versions comparisons. 
\begin{minipage}[c]{0.55\textwidth}
advanced performance in defending against the same attacks while failing to defend against unseen attacks \citep{chen2023robust}. FARE \citep{schlarmann2024robust} and TeCoA \citep{mao2022understanding} are two AT approaches integrated with CLIP, which enhance CLIP's zero-shot classification robustness while harming clean accuracy significantly and do not generalize to other types of attacks. Adversarial purification \citep{song2017pixeldefend} seeks to eliminate adversarial perturbations by optimizing samples to align them with the distribution of benign samples. Instead of adversarial samples, this approach often requires a generative model that can model the probability of benign samples. It can handle unforeseen attacks but often performs worse than AT methods on seen attacks and has lower inference efficiency. 
In this paper, we aim to produce an adversarially robust zero-shot classifier, so we opt for integrating purification with CLIP. To explore a better purification method, we formalize the purification risk and theoretically
\end{minipage}
\hfill % Horizontal fill to ensure the minipages are separated
\begin{minipage}[c]{0.44\textwidth}
    \includegraphics[width=\textwidth]{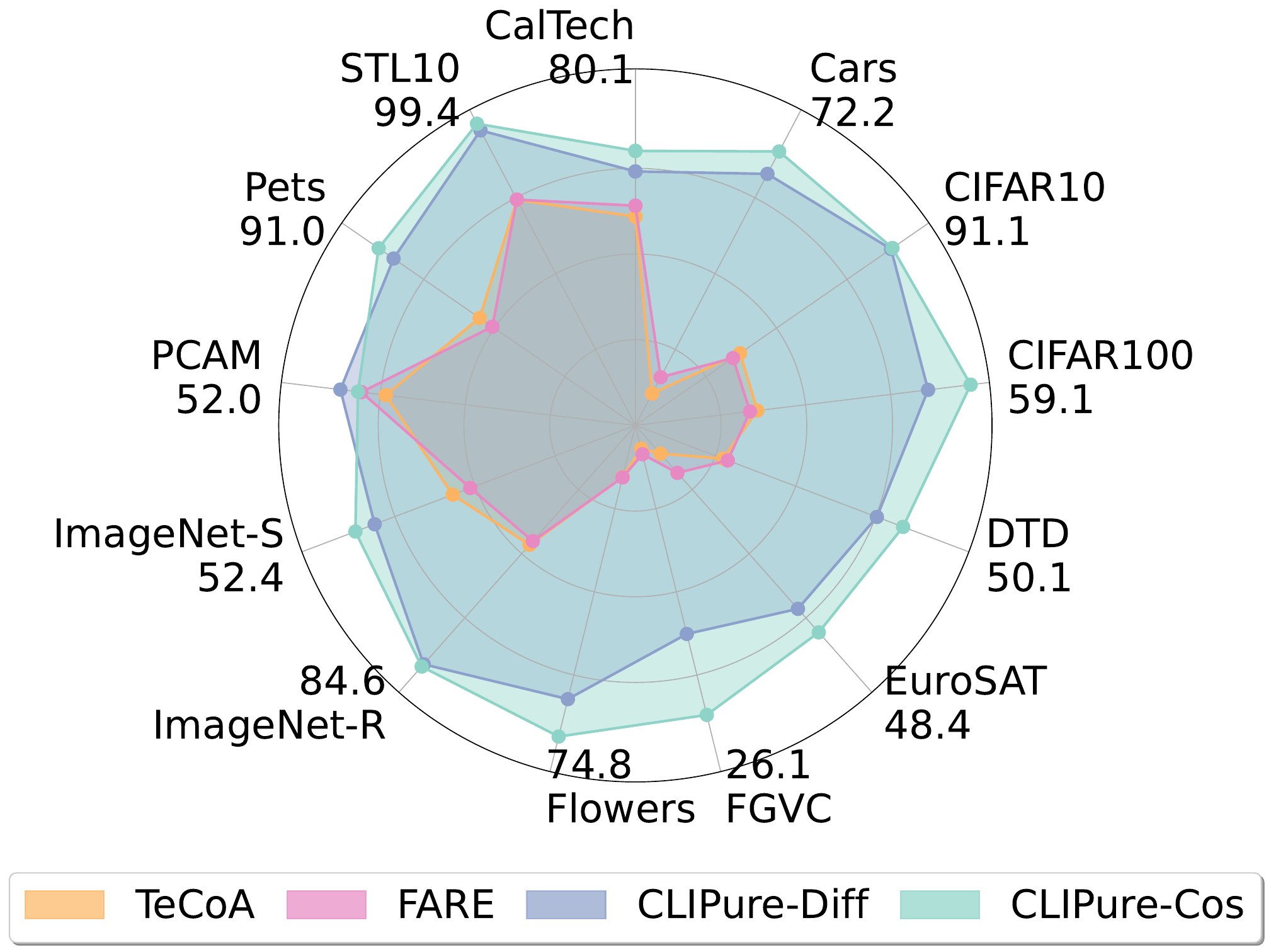} % Adjust the image path and dimensions as necessary
    \captionof{figure}{Adversarial robustness of two CLIPure versions versus adversarially trained CLIP models, evaluated against AutoAttack with $\ell_{\infty}=4/255$ across 13 zero-shot classification datasets.\\}
    \label{figure:13-zeroshot}
  \end{minipage}
analyze what may affect purification performance. Concretely, inspired by \citet{song2020score} that models the diffusion process with bidirectional Stochastic Differential Equations (SDEs) (\cite{anderson1982reverse}), we model the process of attack (i.e., adding perturbations to benign examples) with a forward SDE and purification (i.e., denoising adversarial examples) with a reverse SDE. Within this framework, we evaluate the risk of purification methods by measuring the KL divergence between the joint distributions of the purification and attack steps. After some derivation, we find that purification risk is related to 1) the negative KL divergence of the probability distributions of adversarial and benign examples (i.e., $-$KL$(p(\boldsymbol{x}_{\text{adv}}) || p(\boldsymbol{x}_{\text{ben}}))$), and 2) the $\ell_2$ norm of the gradients of adversarial samples' probability distribution 
% regarding $x_{\text{adv}}$
(i.e., $\nabla \log p(\boldsymbol{x}_{\text{adv}})$). This indicates that purification risk can be affected by: 1) the differences between $p(\boldsymbol{x}_{\text{adv}})$ and $p(\boldsymbol{x}_{\text{ben}})$; 2) the smoothness of $p(\boldsymbol{x}_{\text{adv}})$ and possibly its dimension.

To the best of our knowledge, existing adversarial purification methods are all conducted in pixel space. Given the above factors that can affect purification risk, it is natural to ask: \textit{are there purification approaches better than in pixel space?} As we know, pixel space is high-dimensional and sparse while latent embedding space is denser and smoother. Moreover, multi-modal latent representations are theoretically proven to have better quality than uni-modal \citep{huang2021makes}. CLIP, as a vision-language-aligned encoder model, has shown the superiority of its multi-modal embeddings on many tasks \citep{radford2021learning}. Accordingly, we propose to conduct purification in CLIP's latent space for adversarially robust zero-shot classification. 

Our method - CLIPure has two variants that model the likelihood of images' latent vectors with CLIP differently: 1) CLIPure-Diff: a generative version that employs the DiffusionPrior module of DaLLE-2 \citep{ramesh2022hierarchical} to model the likelihood of an image embedding with a diffusion model; 2) CLIPure-Cos: a discriminative version that models the likelihood with the cosine similarity between the image embedding and text embedding of a blank template ``a photo of a .''. Compared to likelihood modeling in pixel space and uni-modal latent space, we find that both our methods have several orders of magnitude larger KL divergence between the distributions of adversarial and benign samples than the former, and significantly higher than the latter (shown in Figure \ref{figure: distribution of clean and adv examples on px and pz}). Remarkably, CLIPure-Cos becomes the first purification method that does not rely on generative models and thus boosts defense efficiency by two or three orders of magnitude (See Table \ref{table: inference time}). 
%KL$(p(\boldsymbol{x}_{\text{adv}}) || p(\boldsymbol{x}_{\text{ben}}))$

% Disregard the effect of vector length. 
Since CLIP aligns image and text embeddings by their cosine similarities \citep{radford2021learning}, vector lengths are not important to reflect the vector relationship in CLIP's latent space. Hence, during likelihood maximization in CLIPure, we normalize the latent vectors to unit vectors to diminish the effect of vector length. This is critical to our approach, as our experiments indicate that the purification process could be obstructed by vector magnitude and ultimately fail.

We compare the robustness of CLIPure and SOTA methods against the strongest adaptive attacks, AutoAttack\citep{croce2020reliable}, with different $\ell_2$ or $\ell_{\infty}$ bounds on various image classification tasks, including the popular CIFAR10, ImageNet, and 13 other datasets (e.g., CIFAR100, ImagetNet-R) that evaluate zero-shot classification robustness in FARE \citep{schlarmann2024robust} and TeCoA \citep{mao2022understanding}. Note that CLIPure always conducts zero-shot classification and defense without the need for any dataset-specific training while the baselines can be any methods that are current SOTA. We are delighted to see that CLIPure boosts the SOTA robustness on all the datasets by a large margin, e.g., from 71.7\% to 91.1\% on CIFAR10 when $\ell_{\infty}=8/255$, from 59.6\% to 72.6\% on ImageNet when $\ell_{\infty}=4/255$. CLIPure achieves 45.9\% and 108\% relative improvements over previous SOTA - FARE\citep{schlarmann2024robust} regarding average robustness across the 13 zero-shot test datasets facing AutoAttack with $\ell_{\infty}=2/255$ and $4/255$, depicted in Figure \ref{figure:13-zeroshot}. Our work shows that purification in multi-modal latent space is promising for zero-shot adversarial robustness, shedding light on future research including but not limited to image classification.

\section{Related Work}
\textbf{Zero-Shot Image Classification.}
Unlike traditional models that are limited to predefined categories, vision-language models (VLMs) are trained on open-vocabulary data and align the embeddings of images and their captions into a common semantic space. This enables them to perform as zero-shot classifiers by matching the semantics of images to textual categories, offering superior generality and flexibility. CLIP \citep{radford2021learning}, trained on extensive internet image-text pairs, achieves advanced results in zero-shot classification tasks. Additionally, other VLMs including Stable Diffusion \citep{rombach2022high}, Imagen \citep{saharia2022palette}, and DaLLE-2 \citep{ramesh2022hierarchical} also possess zero-shot classification capabilities \citep{li2023your, clark2024text}.

\textbf{Adversarial Purification in Pixel Space.}
A prevalent paradigm of adversarial purification aims to maximize the log-likelihood of samples to remove perturbations in pixel space. Since purification has no assumption of the attack type, enabling it to defend against unseen attacks using pre-trained generative models such as PixelCNN \citep{song2017pixeldefend}, GANs \citep{samangouei2018defense}, VAEs \citep{li2020defense}, Energy-based models \citep{hill2020stochastic, yoon2021adversarial}, and Diffusion Models \citep{nie2022diffusion, chen2023robust, zhang2025causaldiff}. Owing to the capability of diffusion models, diffusion-based adversarial purification achieves state-of-the-art robustness among these techniques.

\textbf{CLIP-based Defense.}
While CLIP achieves impressive accuracy in zero-shot classification, it remains vulnerable to imperceptible perturbations \citep{Fort2021CLIPadversarial, mao2022understanding}. Adversarially training the CLIP model on ImageNet / Tiny-ImageNet \citep{schlarmann2024robust, mao2022understanding, wang2024pre} enhances its robustness but undermines its zero-shot capabilities and struggles against unseen attacks. \citet{choi2025adversarial} suggests smoothing techniques for certification. \citet{li2024one} advocates using robust prompts for image classification, but the defensive effectiveness is limited. Additionally, other research focuses on the out-of-distribution (OOD) robustness of the CLIP model \citep{tu2024closer, galindounderstanding}, which is orthogonal to our adversarial defense objectives.

% \section{Preliminary: Training and Inference Strategies of CLIP}
\section{Preliminary: CLIP as a Zero-shot Classifier}
\label{section: preliminary of CLIP}
In this section, 
% we introduce the training and inference methods of CLIP, which are fundamental to our CLIPure approach. 
we introduce how CLIP is trained and how CLIP acts as a zero-shot classifier. 
CLIP (Contrastive Language Image Pre-training) \citep{radford2021learning}, consists of an image encoder $\text{Enc}^i$ and a text encoder $\text{Enc}^t$. It is trained on 400 million image-text pairs from the internet, aiming to align image embeddings with their corresponding text captions through contrastive learning:
\begin{equation}
\begin{aligned}
\mathcal{L}_{\text{CLIP}} = -\frac{1}{2N} \sum_{n=1}^{N} \left[ \log \frac{\exp(\cos(\boldsymbol{z}^i_n, \boldsymbol{z}^t_n) / \tau)}{\sum_{m=1}^{N} \exp(\cos(\boldsymbol{z}^i_n, \boldsymbol{z}^t_m) / \tau)} + \log \frac{\exp(\cos(\boldsymbol{z}^i_n, \boldsymbol{z}^t_n) / \tau)}{\sum_{m=1}^{N} \exp(\cos(\boldsymbol{z}^i_m, \boldsymbol{z}^t_n) / \tau)} \right],
\end{aligned}
\end{equation}
where $N$ represents the number of image-caption pairs, $\boldsymbol{z}_n^i = \text{Enc}^i(\text{image$_n$})$ and $\boldsymbol{z}_n^t = \text{Enc}^t(\text{text$_n$})$ are the embeddings of the $n$-th image and text respectively, $\tau$ is a temperature parameter, and $\cos(\cdot, \cdot)$ denotes the cosine similarity function.

% This alignment enables CLIP to perform zero-shot classification by matching image embeddings to text embeddings corresponding to class descriptions. Class descriptions typically use a templated format such as 'a photo of a {class}', e.g., 'a photo of a dog' for categorizing dog images. The text embedding for each class $j$ is generated by $\boldsymbol{z}^t_j = \text{Enc}^t(\text{'a photo of a {\text{class}$_j$}'})$. The classification of an image $\boldsymbol{x}$ is then determined by: 
This alignment enables CLIP to perform zero-shot classification by matching image embeddings with text embeddings of a template ``a photo of a $<$class-name$>$.'', where $<$class-name$>$ iterates all the possible classes of a dataset. Without loss of generality, given an image, let $\boldsymbol{z}^i$ denote its CLIP-encoded image embedding, and $\boldsymbol{z}^t_c$ be the text embedding of a possible class description, i.e.,  $\boldsymbol{z}^t_c=\text{Enc}^t(\text{``a photo of a {\text{class} $c$}.''})$. The predicted class $\hat{y}$ is determined by:
\begin{equation} 
\label{equation:zero-shot-classification}
\begin{aligned} 
\hat{y} = \arg \max_c \cos(\boldsymbol{z}^i, \boldsymbol{z}^t_c).
\end{aligned} 
\end{equation} 
% where $\boldsymbol{z}^i = \text{Enc}^i(\boldsymbol{x})$ and $\boldsymbol{z}^t_j$ is the text embedding of the class description. 
For enhanced classification stability, as in \citet{radford2021learning}, we use 80 templates of diverse descriptions in combination with class names, such as ``a \textit{good} photo of $<$class-name$>$''. In our experiments, each class $c$'s embedding, $\boldsymbol{z}^t_c$ in Eq. \ref{equation:zero-shot-classification} is the average text embedding of $c$ paired with all the templates.

\section{CLIPure: Adversarial Purification in Latent Space via CLIP}
In this section, we outline the methodology of our CLIPure, focusing on adversarial purification within CLIP's latent space. We first define purification risk through a Stochastic Differential Equation (SDE) perspective and derive its lower bound in Section~\ref{section: adversarial purification risk}. Section~\ref{section: purification in CLIP latent space} introduces the rationale for CLIPure to potentially achieve a smaller purification risk and two variants of modeling sample likelihood. In Section~\ref{section: purification on direction}, we propose normalizing latent vectors to diminish the effect of vector length during purification to align with CLIP's latent space modeled using cosine similarity. 
% purifying in CLIP's latent space by disregarding magnitude and solely modifying the embedding direction, which aligns with how CLIP models its latent space.

% In this section, 首先定义了purificatin risk through an SDE view and推导了purification risk 的lower bound in Section~\ref{section: adversarial purification risk}. Section~\ref{section: purification in CLIP latent space} introduce the rationale of CLIP's latent space for smaller purification risk for potential for better adversarial purification. Section~\ref{section: purification on direction} we propose 在CLIP隐空间的净化时消除magnitue 的影响只modify embedding direction property, which alighs with the modeling of CLIP’s latent space.

% our CLIPure including:

% \begin{itemize}
%     \item highlights the smoothness of log-likelihood in latent space may reduce purification risk leading to better robustness.
%     \item 
    
%     \textbf{Which} latent space is better for adversarial purification? Section~\ref{} outlines the multimodal model's (i.e., CLIP) latent space better distinguishes clean from adversarial distributions, thus enhancing purification effectiveness.
%     \item \textbf{How} to purify in CLIP latent space? n Section~\ref{section: purification on direction}, we propose to modify the direction of image embeddings in CLIP’s latent space, focusing on semantic representation rather than magnitude.
% \end{itemize}

\subsection{Adversarial Purification Risk}
\label{section: adversarial purification risk}
Considering that adversarial attacks progressively add perturbations to an image while purification gradually removing noise to restore the original image, we formulate both the attack and purification processes through the lens of Stochastic Differential Equations (SDEs). This framework allows us to propose a measure of purification risk based on the divergence between the attack and purification processes, providing insights into what affects purification effectiveness. 
% designing more effective purification strategies.

We formulate the attack process as a transformation from the benign distribution $p_{\text{ben}}(\boldsymbol{x})$ to an adversarial example distribution $p_{\text{adv}}(\boldsymbol{x})$ by an attack algorithm. Note that for simplicity we use $p(\boldsymbol{x})$ to represent $p_{\text{ben}}(\boldsymbol{x})$ in this paper. 
Take untargeted PGD-attack \citep{mkadry2017towards} for instance, the adversarial attack behavior on a benign sample $\boldsymbol{x}_0$ can be described as:
\begin{equation}
\begin{aligned}
\mathrm{d} \boldsymbol{x} = \alpha \mathrm{sign} (\nabla_{\boldsymbol{x}} \mathcal{L}(\theta; \boldsymbol{x}_t, y_{\text{true}})) \mathrm{d} t + \sigma \mathrm{d} \boldsymbol{w}_t, \quad \boldsymbol{x}_0 \sim p(\boldsymbol{x}), \quad \text{s.t.,} \quad \| \boldsymbol{x}_T - \boldsymbol{x}_0 \|_{\rho} \leq \epsilon,
\label{equation: forward sde of untargeted attack}
\end{aligned}
\end{equation}
where $\alpha$ represents the attack step size, $p(\boldsymbol{x})$ denotes the distribution of benign samples, $\mathcal{L}(\theta; \boldsymbol{x}_t, y_{\text{true}})$ denotes the loss of $\boldsymbol{x}_t$ classified by the model with parameters $\theta$ as the ground truth category $y_{\text{true}}$ at attack step $t$ (where $t \in [0, T]$), $\mathrm{d} \boldsymbol{w}_t$ denotes the Wiener process (Brownian motion). The constant $\sigma$ serves as a scaling factor for the noise component and the adversarial example $\boldsymbol{x}_T$ is bounded by $\epsilon$ in $\ell_{\rho}$ norm.

The corresponding reverse-time SDE \citep{anderson1982reverse} of Eq.~\ref{equation: forward sde of untargeted attack} describes the process from the adversarial example distribution $p_{\text{adv}}$ to the purified sample distribution $p_{\text{pure}}$, and is expressed as:
\begin{equation}
\begin{aligned}
d\boldsymbol{x} = [\alpha \mathrm{sign} (\underbrace{\nabla_{\boldsymbol{x}} \mathcal{L}(\theta; \boldsymbol{x}_t, y_{\text{true}}}_{\text{classifier guidance}})) - \sigma^2 \underbrace{\nabla \log p(\boldsymbol{x}_t}_{\text{purification}})] \mathrm{d} t + \sigma \mathrm{d} \tilde{\boldsymbol{w}}_t, \quad \boldsymbol{x}_T \sim p_{\mathrm{adv}}(\boldsymbol{x}),
\label{equation: reverse sde of untargeted attack}
\end{aligned}
\end{equation}
where $\log p(\boldsymbol{x}_t)$ represents the log-likelihood of $\boldsymbol{x}_t$ concerning the distribution of clean samples, analogous to the score function described in Score SDE \citep{song2020score}, $\mathrm{d} \tilde{\boldsymbol{w}}_t$ represents the reverse-time Wiener process. A detailed discussion on the form of the reverse-time SDE can be found in Appendix~\ref{appendix: discussion of the reverse SDE}.

Note that in the reverse-time SDE, $t$ progresses from $T$ to $0$, implying that $\mathrm{d}t$ is negative. According to Eq.~\ref{equation: reverse sde of untargeted attack}, the reverse SDE aims to increase the sample's log-likelihood while simultaneously decreasing the loss of classifying $\boldsymbol{x}$ to $y_{\text{true}}$. In Eq.~\ref{equation: reverse sde of untargeted attack}, the purification term is related to the common objective of adversarial purification $\boldsymbol{x}_{\text{pure}} = \arg \max_{\boldsymbol{x}} \log p(\boldsymbol{x})$. The classifier guidance term has been employed to enhance purification \citep{zhang2024classifier}, and their objective aligns well with Eq. \ref{equation: reverse sde of untargeted attack}. We will incorporate this guidance term with CLIPure in Appendix ~\ref{section: combination} and see its impact.

% Regarding the classifier guidance term, previous work \citep{zhang2024classifier} has discussed enhancing adversarial robustness by increasing the classifier's confidence, and their purification objective aligns well with the reverse-time SDE framework described in Eq.~\ref{equation: reverse sde of untargeted attack}. Since the ground truth category $y_{\text{true}}$ is unknown, estimating $y_{\text{true}}$ constitutes an additional topic that does not impact our exploration of purification strategies. We will discuss this further in Section~\ref{section: combination}.

Then, we define the joint distribution of the attack process described by the forward SDE in Eq.~\ref{equation: forward sde of untargeted attack} as $\mathcal{P}_{0:T} = p(\boldsymbol{x}_0=\boldsymbol{x}_{\text{ben}}, \boldsymbol{x}_1, ..., \boldsymbol{x}_T=\boldsymbol{x}_{\text{adv}}) \in \mathbb{R}^{(T+1)\times d}$, where each $\boldsymbol{x}_t \in \mathbb{R}^d$. For the purification process defined by the reverse-time SDE in Eq.~\ref{equation: reverse sde of untargeted attack}, we denote the joint distribution as $\mathcal{Q}_{0:T} = p(\boldsymbol{x}_{0}=\boldsymbol{x}_{\text{pure}}, \boldsymbol{x}_{1}, ..., \boldsymbol{x}_{T}=\boldsymbol{x}_{\text{adv}})$. Here, $\boldsymbol{x}_{\text{ben}}$, $\boldsymbol{x}_{\text{adv}}$, and $\boldsymbol{x}_{\text{pure}}$ denotes the benign sample, adversarial example, and purified sample respectively. We define the purification risk $\mathcal{R}(\mathcal{Q})$ by the KL divergence between the reverse SDE (corresponding to the purification process) and the joint distribution of forward SDE (representing the attack process):
\begin{equation}
\begin{aligned}
\mathcal{R}(\mathcal{Q}) & := \mathrm{KL}(\mathcal{Q}_{0:T} \| \mathcal{P}_{0:T}) = \mathrm{KL}(\mathcal{Q}_{0, T} \| \mathcal{P}_{0,T}) + \mathbb{E}_{\mathcal{Q}_{0,T}}[\mathrm{KL}(\mathcal{Q}_{1:T-1|0,T} \| \mathcal{P}_{1:T-1|0,T})] \\
& \geq \  \mathrm{KL}(\mathcal{Q}_{0, T} \| \mathcal{P}_{0,T}).
\end{aligned}
\end{equation}

Then, we focus solely on the purified example, i.e., $\mathrm{KL}(\mathcal{Q}_{0, T} \| \mathcal{P}_{0,T})$ rather than the entire purification trajectory. The forward SDE in Eq.~\ref{equation: forward sde of untargeted attack} describes the transformation from $p(\boldsymbol{x}_{\text{ben}})$ to $p(\boldsymbol{x}_{\text{adv}})$, enabling us to obtain the conditional probability $p(\boldsymbol{x}_{\text{adv}}|\boldsymbol{x}_{\text{ben}})$. Simultaneously, the reverse SDE in Eq.~\ref{equation: reverse sde of untargeted attack} supports the transformation from $p(\boldsymbol{x}_{\text{adv}})$ back to $p(\boldsymbol{x}_{\text{pure}})$, helping us to quantify $p(\boldsymbol{x}_{\text{pure}}|\boldsymbol{x}_{\text{adv}})$. Then we can derive that:
\begin{equation}
\begin{aligned}
\mathcal{R}(\mathcal{Q}) \geq & \  \mathrm{KL}(\mathcal{Q}_{0, T} \| \mathcal{P}_{0,T}) = \mathrm{KL}(p(\boldsymbol{x}_{\text{pure}}, \boldsymbol{x}_{\text{adv}}) \| p(\boldsymbol{x}_{\text{ben}}, \boldsymbol{x}_{\text{adv}}))\\
% = & \ \mathbb{E}_{\boldsymbol{x}_{\text{adv}}} \left[ \mathrm{KL}(p(\boldsymbol{x}_{\text{pure}} | \boldsymbol{x}_{\text{adv}}) \| p(\boldsymbol{x}_{\text{adv}} | \boldsymbol{x}_{\text{ben}})) \right] - \mathrm{KL}(p(\boldsymbol{x}_{\text{adv}}) \| p(\boldsymbol{x}_{\text{ben}})) \\
= & \ \frac{1}{2} \mathbb{E}_{\boldsymbol{x}_{\text{adv}}} \left[ \nabla \log p(\boldsymbol{x}_{\text{adv}})^T \nabla \log p(\boldsymbol{x}_{\text{adv}}) \sigma^2 \Delta t \right] - \mathrm{KL}(p(\boldsymbol{x}_{\text{adv}}) \| p(\boldsymbol{x}_{\text{ben}})),
% \geq & - \mathrm{KL}(p(\boldsymbol{x}_{\text{adv}}) \| p(\boldsymbol{x}_{\text{ben}}))
\label{equation: purification risk}
\end{aligned}
\end{equation}
where $\Delta t$ denotes a small time interval for attack and purification, related to the perturbation magnitude. A detailed proof of the result is provided in Appendix~\ref{appendix: proof of purification quality function}.

% This quantifies the adjustments to the probability density brought about by the purification process from $\boldsymbol{x}_{\text{adv}}$ to $\boldsymbol{x}_{\text{purified}}$.

The result in Eq.~\ref{equation: purification risk} highlights that the lower bound of the purification risk is influenced by two factors: 1) the smoothness of the log-likelihood function at adversarial examples and possibly the sample dimension, as indicated by the $\ell_2$ norm of $\nabla \log p(\boldsymbol{x}_{\text{adv}})$, 2) the differences between the likelihood of clean and adversarial samples in the benign example space.  

\begin{figure}[t]
\centering
% \label{figure:px-pz-multi-uni-modal}
\newlength{\commonheight}
\setlength{\commonheight}{2.7cm}
\begin{subfigure}[b]{0.23\linewidth}  % Width set to one third of text width
    \includegraphics[width=\linewidth, height=\commonheight]{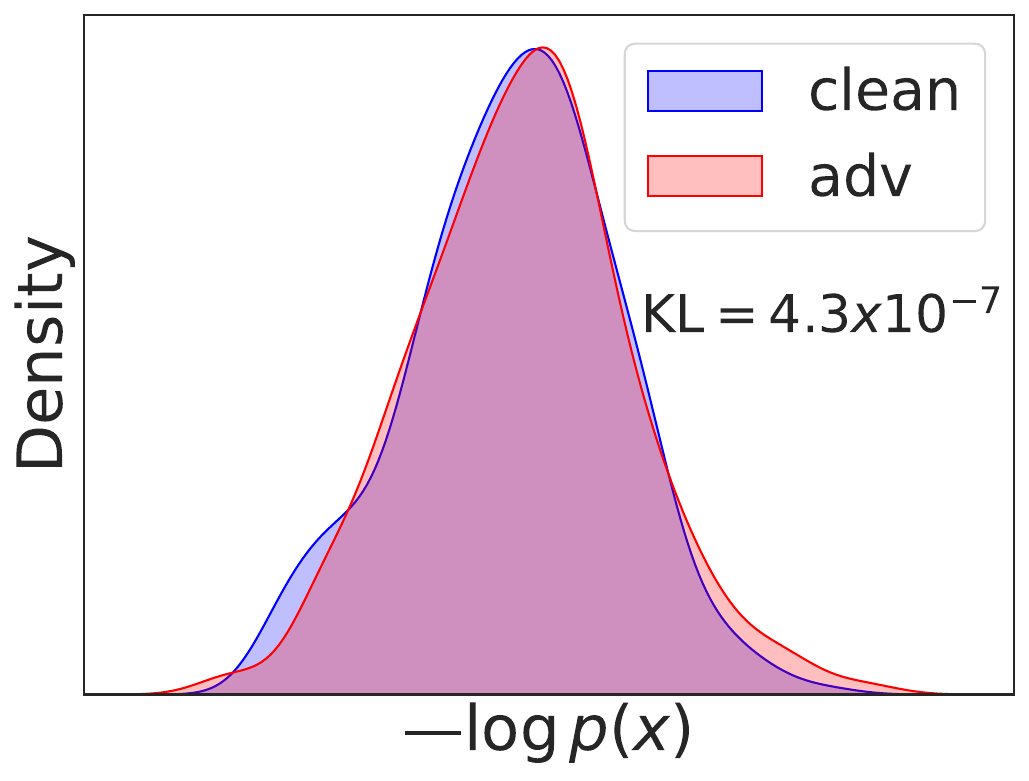}
    \caption{Pixel Space}
    \label{subfigure: log-likelihood distribution of px}
\end{subfigure}
\hfill  % Add space between subfigures
\begin{subfigure}[b]{0.23\linewidth}
    \includegraphics[width=\linewidth, height=\commonheight]{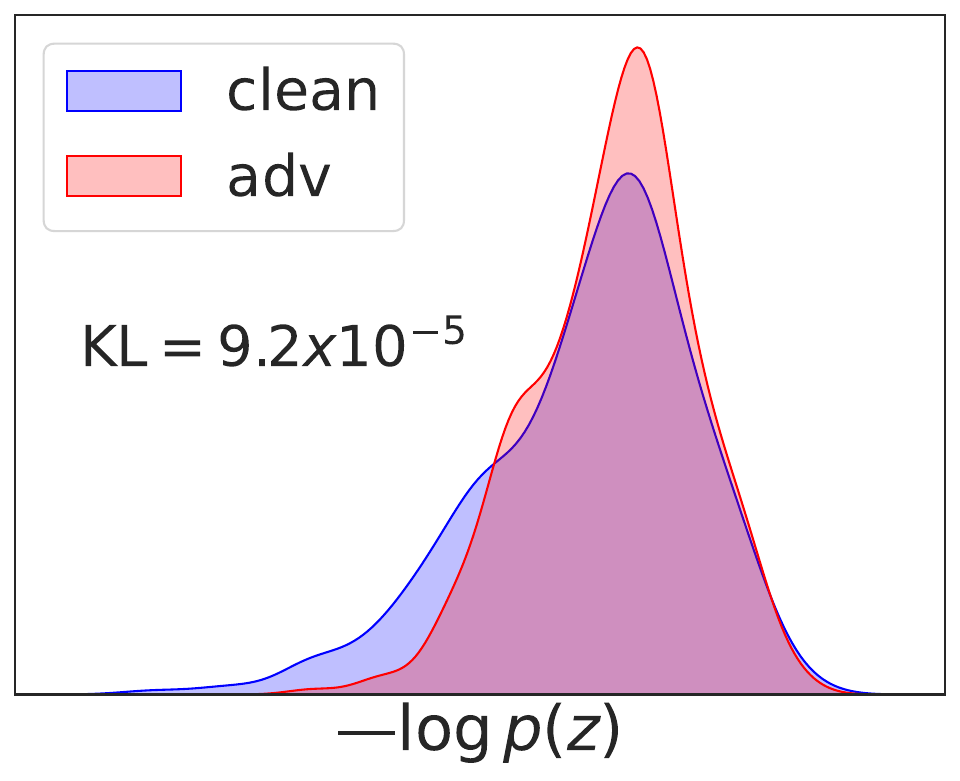}
    \caption{Vision Latent Space}
    \label{subfigure: log-likelihood distribution of pz by Stable Diffusion}
\end{subfigure}
\hfill  % Add space between subfigures
\begin{subfigure}[b]{0.23\linewidth}
    \includegraphics[width=\linewidth, height=\commonheight]{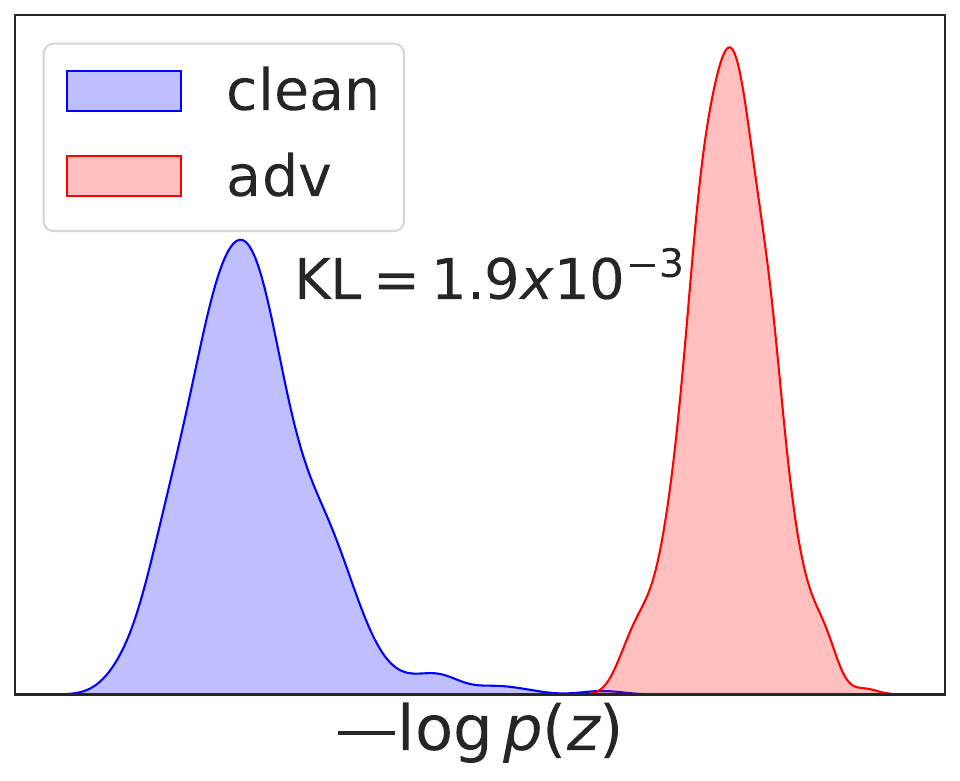}
    \caption{Multi-modal(Diff)}
    \label{subfigure: log-likelihood distribution of pz by DiffusionPrior}
\end{subfigure}
\hfill
\begin{subfigure}[b]{0.23\linewidth}
    \includegraphics[width=\linewidth, height=\commonheight]{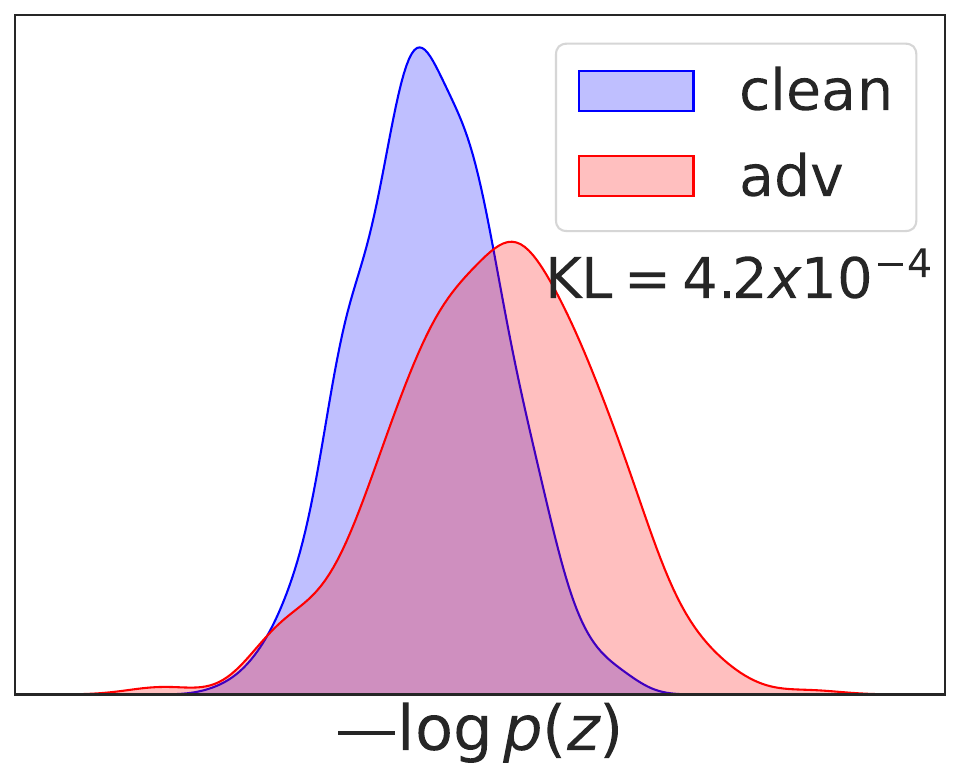}
    \caption{Multi-modal(Cos)}
    \label{subfigure: log-likelihood distribution of pz by CLIP}
\end{subfigure}
\caption{Negative log-likelihood estimated by diffusion models on (a) pixel space via EDM, (b) uni-modal latent space via VQVAE, (c) multi-modal latent space via DiffusionPrior, and (d) multi-modal latent space via CLIP (using cosine similarity for log-likelihood estimation). KL represents the value of $\mathrm{KL}(p(\boldsymbol{x}_{\text{adv}}) \| p(\boldsymbol{x}_{\text{ben}}))$ discussed in Section~\ref{section: purification in CLIP latent space} indicating the difference between clean and adversarial example distribution.}
\label{figure: distribution of clean and adv examples on px and pz}
\end{figure}

\subsection{Adversarial Purification in CLIP's Latent Space}
\label{section: purification in CLIP latent space}

In this section, we further explore how to achieve a smaller purification risk $\mathcal{R}(\mathcal{Q})$.
Considering $\mathbb{E}_{\boldsymbol{x}_{\text{adv}}} \left[ \nabla \log p(\boldsymbol{x}_{\text{adv}})^T \nabla \log p(\boldsymbol{x}_{\text{adv}}) \sigma^2 \Delta t \right]$ within $\mathcal{R}(\mathcal{Q})$ in Eq. \ref{equation: purification risk}, standard pixel space purification may lead to higher purification risks due to its sparsity and possibly peaked gradient distribution in high dimensionality. Thus, we are curious to investigate purification in latent space where the distribution of sample densities is more uniform and smoother. 
% the sparse and high-dimensional nature of pixel space often results in large gradients of likelihood. This leads to a high purification risk, potentially causing significant differences between purified and benign images. 
% Thus, we propose purifying adversarial examples in latent space, where the distribution of sample densities is more uniform and the characteristics are smoother. 

$\mathrm{KL}(p(\boldsymbol{x}_{\text{adv}}) \| p(\boldsymbol{x}_{\text{ben}}))$ in Eq. \ref{equation: purification risk} implies
% the ability to distinguish between clean and adversarial distributions significantly impacts the purification risk. 
representations that excel at detecting out-of-distribution adversarial examples are likely to carry a lower risk of purification errors. \citet{huang2021makes} suggest that multi-modal latent spaces offer superior quality compared to uni-modal counterparts. Inspired by this observation, CLIP's well-aligned multi-modal latent space, where image embeddings are guided by the semantics of finer-grained words in an open vocabulary, may provide a foundation for purification with a lower risk.  
% Textual data, offering fine-grained guidance through an open vocabulary, enhances the quality of image embeddings in multimodal models, as supported by \citet{huang2021makes}. 
% Given the advanced capabilities and extensive training of the CLIP model on diverse image-text datasets, we propose leveraging CLIP's latent space for lower purification risk, expecting this approach to yield a better adversarial purification. 

% Regarding image classification, embeddings represent image semantics, aiming to extract key information for categorization. While uni-modal models like classifiers or VAEs lack good semantic supervision, making adversarial purification challenging. In contrast, multimodal models like CLIP use image-corresponding captions during training. This approach aligns image content with textual semantics, providing a higher-quality latent space that enhances adversarial purification.

To validate the efficacy of different spaces in modeling sample likelihood for adversarial purification, we focus on the term $\mathrm{KL}(p(\boldsymbol{x}_{\text{adv}}) \| p(\boldsymbol{x}_{\text{ben}}))$ in the lower bound of the purification risk $\mathcal{R}(\mathcal{Q})$.
% This focus is due to the challenge in fairly comparing the first term across pixel and latent spaces, as it tends to be smaller and less variable.
% To validate this idea, we prioritize understanding how well adversarial and benign distributions can be differentiated in each space to evaluate their potential for effective purification. We specifically examine the term $\mathrm{KL}(p(\boldsymbol{x}_{\text{adv}}) \| p(\boldsymbol{x}_{\text{ben}}))$, which represents the lower bound of the purification risk $\mathcal{R}(\mathcal{Q})$. This is due to the difficulty in fairly comparing the first term $\mathbb{E}_{\boldsymbol{x}_{\text{adv}}} \left[ \nabla \log p(\boldsymbol{x}_{\text{adv}})^T \nabla \log p(\boldsymbol{x}_{\text{adv}}) \sigma^2 \Delta t \right]$ between pixel space and latent space, as this term is usually smaller and less variable.
As illustrated in Figure~\ref{figure: distribution of clean and adv examples on px and pz}, we extract 512 samples from ImageNet \citep{deng2009imagenet} and generate adversarial examples by AutoAttack \citep{croce2020reliable} on the CLIP \citep{radford2021learning} classifier. We explore four types of likelihood modeling approaches:

\textbf{In Pixel Space:}
Sample likelihood of the joint distribution of image pixels is estimated by a generative model (we use an advanced diffusion model - EDM\citep{karras2022elucidating}). Figure~\ref{subfigure: log-likelihood distribution of px} indicates that even EDM struggles to distinguish between clean and adversarial sample distributions at the pixel level. We use the Evidence Lower Bound (ELBO) to estimate log-likelihood, expressed as $\log p_{\theta}(\boldsymbol{x}) \geq -\mathbb{E}_{\boldsymbol{\epsilon},t}[\|\epsilon_{\theta}(\boldsymbol{x}_{t}, t) - \boldsymbol{\epsilon} \|_2^2] + C$, where $\boldsymbol{\epsilon} \sim \mathcal{N}(\boldsymbol{0}, \boldsymbol{I})$, and $C$ is typically negligible \citep{ho2020denoising, song2020score}.

\textbf{In Vision Latent Space:}
The likelihood of the joint distribution of an image embedding in a uni-modal space is estimated with the VQVAE component of the Stable Diffusion (SD) model \citep{rombach2022high}. Note that although SD is multi-modal, its VQVAE component is trained solely on image data and keeps frozen while training the other parameters, making it a vision-only generative model of latent vectors. 
Compared to Figure~\ref{subfigure: log-likelihood distribution of px}, Figure~\ref{subfigure: log-likelihood distribution of pz by Stable Diffusion} demonstrates improved capability in distinguishing clean and adversarial sample distributions. 
% , as shown In Figure~\ref{subfigure: log-likelihood distribution of pz by Stable Diffusion}. 
% As for log-likelihood estimation in vision latent space, we leverage the VQVAE component of the Stable Diffusion (SD) model \citep{rombach2022high}. 
% This VQVAE is trained solely on image data while remains parameter-frozen during the training of the latent diffusion model conditioned on the image-corresponding caption. This makes the VQVAE a purely vision-based component, even though SD is a vision-language model. 

\textbf{In Vision-Language Latent Space:}
For CLIP's latent space, we present two approaches to estimate the log-likelihood of image embeddings, i.e., $\log p(\boldsymbol{z}^i)$:
% , detailed as follows: 

(1) \textbf{Diffusion-based} Likelihood Estimation (corresponds to \textit{CLIPure-Diff}): The DiffusionPrior module in DaLLE-2 (based on CLIP) \citep{ramesh2022hierarchical} models the generative process of image embeddings conditioned on text embeddings. Employing DiffusionPrior, we estimate the log-likelihood $\log p_{\theta}(\boldsymbol{z}^i)$ by conditioning on a blank template ``a photo of a .'': \begin{equation} \log p_{\theta}(\boldsymbol{z}^i) \approx \log p_{\theta}(\boldsymbol{z}^i | \bar{\boldsymbol{z}}^t) = -\mathbb{E}_{\boldsymbol{\epsilon},t}[|\epsilon_{\theta}(\boldsymbol{z}^i_{t}, t, \bar{\boldsymbol{z}}^t) - \boldsymbol{\epsilon} |_2^2] + C, \end{equation} where $\epsilon_{\theta} (\boldsymbol{z}^i, t, \boldsymbol{z}^t)$ is the UNet \citep{ronneberger2015u} in DiffusionPrior \citep{ramesh2022hierarchical}, parameterized by $\theta$, with the noised image embedding $\boldsymbol{z}^i_{t}$ at timestep $t$ under the condition of text embedding $\boldsymbol{z}^t$, and $C$ is a constant typically considered negligible \citep{ho2020denoising}. 

(2) \textbf{Cosine Similarity-based} Likelihood Estimation (corresponds to \textit{CLIPure-Cos}): We estimate $\log p_{\theta}(\boldsymbol{z}^i)$ by computing the cosine similarity between image embedding $\boldsymbol{z}^i$ and the blank template’s text embedding:
\vspace{-1mm}
   \begin{equation}
   \log p_{\theta}(\boldsymbol{z}^i) \propto \cos(\boldsymbol{z}^i, \bar{\boldsymbol{z}}^t).
   % \approx \log \frac{\cos(\boldsymbol{z}^i, \bar{\boldsymbol{z}}^t) + 1}{2}.
   \end{equation}
% Consider the value range, we estimate $p_{\theta}(\boldsymbol{z}^i)$ by $(\cos(\boldsymbol{z}^i, \bar{\boldsymbol{z}}^t) + 1)/2$. For simplicity, we directly optimize $\cos(\boldsymbol{z}^i, \bar{\boldsymbol{z}}^t)$ in our experiments.
Note that by modeling likelihood without using generative models, the defense efficiency can be significantly boosted. The inference time of CLIPure-Cos is only 1.14 times of the vanilla CLIP for zero-shot classification, shown in Table \ref{table: inference time}. 

Figure \ref{subfigure: log-likelihood distribution of pz by DiffusionPrior} and Figure~\ref{subfigure: log-likelihood distribution of pz by CLIP} show that CLIPure-Diff and CLIPure-Cos have several orders larger magnitude KL divergence between clean and adversarial samples than modeling likelihood in pixel space. Modeling likelihood in uni-modal latent space also leads to larger KL divergence than pixel space but is smaller than multi-modal space. It indicates that purification in CLIP's latent space is promising to have lower purification risk and enhance adversarial robustness.

% As shown in Figure~\ref{subfigure: log-likelihood distribution of pz by CLIP}, clean samples exhibit higher semantic similarity with a blank template than adversarial samples, reflecting the alignment of text and image embeddings during training compared with adversarial examples. This method thus provides a feasible way to estimate log-likelihood for adversarial purification in CLIP's latent space, underpinning our CLIPure-Cos approach detailed in Algorithm~\ref{algorithm: latent purification by CLIP and DaLLE2.DiffusionPrior}.

% These approaches showcase how different spaces contribute to the effectiveness of adversarial purification, emphasizing the potential of vision-language latent space in distinguishing between benign and adversarial distributions more effectively than traditional pixel or single-modality latent spaces. Consequently, the latent space of CLIP offers a lower purification risk and holds greater potential for enhancing adversarial robustness.

% Notably, our method of adversarial purification operates in a plug-and-play manner, requiring no additional training of the models.

\begin{algorithm*}[t]
  \caption{\textbf{CLIPure}: Adversarial Purification in Latent Space via CLIP}
  \label{algorithm: latent purification by CLIP and DaLLE2.DiffusionPrior}
  \textbf{Required:}{ An off-the-shelf CLIP model including an image encoder $\text{Enc}^{i}$ and a text encoder $\text{Enc}^t$, textual embedding $\bar{\boldsymbol{z}}^t$ of the blank templates (without class names, e.g., 'a photo of a.'),  purification step $N$, step size $\eta$, and a DiffusionPrior model $\epsilon_{\theta}$ from DaLLE-2 (for generative version).} \\
  \textbf{Input: }{Latent embedding of input example $\boldsymbol{z}^i$} \\
  % \textbf{Output: }{Purified latent embedding}
  \textbf{Output: }{label $y$}
  % \vspace{1mm} 
  
  \textbf{for} $i = 1$ \textbf{to} $N$, \textbf{do}
    
    \quad \textbf{step1: Obtain latent embedding}

    \vspace{0.5mm}
    \quad \quad direction $\boldsymbol{u} = \boldsymbol{z}^i / \|\boldsymbol{z}^i \|_2^2$, \  magnitude $r = \boldsymbol{z}^i / \boldsymbol{u}$

    % \vspace{0.5mm}
    % \quad \quad magnitude $r = \boldsymbol{z}^i / \boldsymbol{u}$

    \vspace{0.6mm} 
    \quad \textbf{step2: Estimate log-likelihood}
    \vspace{0.6mm}
    
  \noindent\begin{minipage}[t]{.49\textwidth}
    \quad \quad \textbf{CLIPure-Diff} via DiffusionPrior $\epsilon_{\theta}$

    % \vspace{0.4mm}
    % \quad \quad \quad $\bar{\boldsymbol{z}}^t = \text{Enc}^{t}(\mathcal{T})$

    \vspace{0.4mm}
    \quad \quad \quad sample $\boldsymbol{\epsilon} \sim \mathcal{N}(\mathbf{0}, \mathbf{1})$

    \vspace{0.4mm}
    \quad \quad \quad $ \log p(\boldsymbol{z}^i) \leftarrow -\|\epsilon_{\theta}(\boldsymbol{z}^i_{t}, t, \bar{\boldsymbol{z}}^t) - \boldsymbol{\epsilon} \|_2^2 $

  \end{minipage}%
  \hfill\vline\hfill
  \begin{minipage}[t]{.49\textwidth}
    \textbf{CLIPure-Cos} via CLIP
    
    % \quad \quad \textbf{Discriminative version} via CLIP

    % \vspace{1.5mm}
    % \quad $\bar{\boldsymbol{z}}^t = \text{Enc}^{t}(\mathcal{T})$

    \vspace{1.8mm}
    \quad $ \log p(\boldsymbol{z}^i) \leftarrow \cos(\boldsymbol{z}^i, \bar{\boldsymbol{z}}^t)
     % \approx \log \frac{\cos(\boldsymbol{z}^i, \bar{\boldsymbol{z}}^t) + 1}{2}
     $

  \end{minipage}

    \vspace{1.2 mm} 
    \quad \textbf{step3: Update latent variable}
    \vspace{0.2mm}
    
    % \noindent\begin{minipage}[t]{.49\textwidth}
    
    % \quad \quad $\boldsymbol{u} \leftarrow \boldsymbol{u} + \eta \nabla_{\boldsymbol{z}^i} \log p(\boldsymbol{z}^i) \cdot \nabla_{\boldsymbol{u}} \boldsymbol{z}^i$
    
    \quad \quad $\boldsymbol{u} \leftarrow \boldsymbol{u} + \eta \frac{\partial \log p(\boldsymbol{z}^i)}{\partial \boldsymbol{z}^i} \cdot \frac{\partial \boldsymbol{z}^i}{\partial \boldsymbol{u}}$
    % , \ $\boldsymbol{z}^i \leftarrow r \cdot \boldsymbol{u}$
    % \quad \quad $\boldsymbol{u} \leftarrow \boldsymbol{u} + \eta \frac{\partial \log p(\boldsymbol{z}^i) }{\partial \boldsymbol{z}^i} \cdot \frac{\partial \boldsymbol{z}^i}{\partial \boldsymbol{u}}$
    % \vspace{0.2mm}
    \ , \  $\boldsymbol{z}^i \leftarrow r \cdot \boldsymbol{u}$
  %   \end{minipage}%
  % \hfill\vline\hfill
  % \begin{minipage}[t]{.49\textwidth}
  % \vspace{1.2 mm} 
  %   \quad 
  %   \vspace{0.2mm}
    
  %   \quad $\boldsymbol{u} \leftarrow \boldsymbol{u} + \eta \frac{\partial \log p(\boldsymbol{z}^i)}{\partial \boldsymbol{z}^i} \cdot \frac{\partial \boldsymbol{z}^i}{\partial \boldsymbol{u}}$
  %   \ , \  $\boldsymbol{z}^i \leftarrow r \cdot \boldsymbol{u}$
  % \end{minipage}

    \quad \textbf{step4: Classification based on purified embedding}

    \vspace{0.2mm}
    \quad \quad predict label $y$ across candidate categories according to Eq.~\ref{equation:zero-shot-classification}
    
    \color{black}{}
    \textbf{end for}
    % \quad return purified latent $\boldsymbol{z}^i$
    
    return predicted label $y$

\end{algorithm*}

% \subsection{Adversarial Purification on Direction Proporties}

\subsection{Adversarial Purification Based on Normalized Unit Vectors}
\label{section: purification on direction}
Typically, purification in pixel space is conducted through gradient ascent on the sample $\boldsymbol{x}$ by using the derivative of the log-likelihood $\log p(\boldsymbol{x})$: $\boldsymbol{x} \leftarrow \boldsymbol{x} + \alpha \nabla \log p(\boldsymbol{x})$, where $\alpha$ denotes the step size for updates. However, given that the cosine similarities between vectors in CLIP's latent space are the criterion for alignment where vector lengths do not take effect, it is inappropriate to directly apply the typical purification update manner to this space. Thus, we normalize the image vectors to unit vectors at each purification step to diminish the effect of vector length. 
% given CLIP's method of modeling semantic characteristics of image embeddings through contrastive learning, directly modifying image embeddings in the same manner is inappropriate. Instead, we propose purifying the image embedding by maximizing log-likelihood focusing on its directional properties, disregarding vector magnitude which may disturb purification stability. This approach also aligns well with the CLIP model's reliance on the embedding direction for zero-shot classification, as discussed in Section~\ref{section: preliminary of CLIP}.
Specifically, we first normalize the vector $\boldsymbol{z}^i = \text{Enc}^i(\boldsymbol{x})$, obtained from the CLIP model image encoder for an input sample $\boldsymbol{x}$ (potentially an adversarial sample), to a unit vector $\boldsymbol{u} = \boldsymbol{z}^i / \|\boldsymbol{z}^i \|_2^2$.
Then we calculate the sample's log-likelihood $\log p(\boldsymbol{z}^i)$ and compute the gradient $\boldsymbol{g}_{\boldsymbol{u}}$ by the chain rule:
\begin{equation}
\begin{aligned}
\boldsymbol{g}_{\boldsymbol{u}} = \frac{\partial \log p(\boldsymbol{z}^i)}{\partial \boldsymbol{u}} =  \frac{\partial \log p(\boldsymbol{z}^i)}{\partial \boldsymbol{z}^i} \cdot \frac{\partial \boldsymbol{z}^i}{\partial \boldsymbol{u}}.
\end{aligned}
\end{equation}
This gradient $\boldsymbol{g}_{\boldsymbol{u}}$ is then used to update the direction $\boldsymbol{z}^i$ for adversarial purification, detailed in Algorithm~\ref{algorithm: latent purification by CLIP and DaLLE2.DiffusionPrior}. Note that CLIPure is based on the original CLIP and does not need any extra training.
% For a detailed description of this process, please refer to Algorithm~\ref{algorithm: latent purification by CLIP and DaLLE2.DiffusionPrior}.

We also attempted adversarial purification by directly optimizing vectors instead of the normalized version. We experimented extensively with various steps, parameters, and momentum-based methods, but found it challenging to achieve robustness over 10\% on ImageNet, indicating that it is difficult to find an effective purification path with vector lengths taking effect.  

\section{Experiments}
\textcolor{blue}{The results presented below are obtained under the experimental setting with \texttt{create\_graph=False}. In this setup, no new computational graph is generated during the purification process; consequently, this does not constitute a genuine adaptive attack. Instead, it leads to gradient obfuscation. A detailed discussion is provided in Appendix~\ref{appendix: supplementary for attack evaluation}.} 

\subsection{Experimental Settings}
\label{section: experimental settings}
\textbf{Datasets.}
Following the RobustBench \citep{croce2020reliable} settings, we assess robustness on CIFAR-10 and ImageNet. To compare against CLIP-based zero-shot classifiers with adversarial training \citep{schlarmann2024robust, mao2022understanding}, we conduct additional tests across 13 image classification datasets (detailed in Appendix~\ref{appendix: zero-shot datasets}). In line with \citet{schlarmann2024robust}, we randomly sampled 1000 examples from the test set for our evaluations. 
% Notably, our tests utilize the off-the-shelf CLIP model without further training, making all dataset evaluations zero-shot for CLIPure.

\textbf{Baselines.}
We evaluate the performance of \textbf{pixel space purification} strategies employing generative models, including Purify-EBM \citep{hill2020stochastic} and ADP \citep{yoon2021adversarial} based on Energy-Based Models; DiffPure based on Score SDE \citep{nie2022diffusion} and DiffPure-DaLLE2.Decoder based on Decoder of DaLLE2 \citep{ramesh2022hierarchical}; GDMP based on DDPM \citep{ho2020denoising}; and likelihood maximization approaches such as LM-EDM \citep{chen2023robust}     based on the EDM model \citep{karras2022elucidating} and LM-DaLLE2.Decoder which adapts LM to the Decoder of DaLLE-2. Furthermore, we perform an ablation study adapting LM to the latent diffusion model to achieve \textbf{ uni-modal latent space purification} using the Stable Diffusion Model \citep{rombach2022high}, denoted as LM-StableDiffusion. Furthermore, we also compare with the state-of-the-art \textbf{adversarial training} methods such as AT-ConvNeXt-L \citep{singh2024revisiting} and AT-Swin-L \citep{liu2024comprehensive}, along with innovative training approaches such as MixedNUTS \citep{bai2024mixednuts} and MeanSquare \citep{amini2024meansparse}, and methods that utilize DDPM and EDM to generate adversarial samples for training dataset expansion: AT-DDPM \citep{rebuffi2021fixing} and AT-EDM \citep{wang2023better}. Moreover, we also compare CLIPure with adversarially trained CLIP model according to TeCoA \citep{mao2022understanding} and FARE \citep{schlarmann2024robust}. Additionally, we also evaluate the performance of \textbf{classifiers without defense} strategies, including CLIP, WideResNet (WRN), and Stable Diffusion.
% Our baselines include purification in \textbf{pixel space} and in \textbf{uni-modal latent space}, as well as \textbf{adversarial training} strategies.
Due to space constraints, we list the details of baselines in Section~\ref{sec: detailed baselines in appendix} in Appendix~\ref{appendix: experimental settings}.

\textbf{Adversarial Attack.}
Following the setup used by FARE \citep{schlarmann2024robust}, we employ \textbf{AutoAttack}'s \citep{croce2020reliable} strongest white-box APGD for both targeted and untargeted attacks across 100 iterations, focusing on an $\ell_{\infty}$ threat model (typically $\epsilon=4/255$ or $\epsilon=8/255$) as well as $\ell_2$ threat model (typically $\epsilon=0.5$) for evaluation.
% We leverage \textbf{adaptive attack} with full access to the model parameters and inference strategies, including the purification mechanism to expose the model's vulnerabilities thoroughly. It means attackers can compute gradients against the entire CLIPure process according to the chain rule:
% % \begin{equation}
% % \begin{aligned}
% $\frac{\partial \mathcal{L}}{\partial \boldsymbol{x}} = \frac{\partial \mathcal{L}}{\partial \boldsymbol{z}^i_{\text{pure}}} \cdot \frac{\partial \boldsymbol{z}^i_{\text{pure}}}{\partial \boldsymbol{z}^i} \cdot \frac{\partial \boldsymbol{z}^i}{\partial \boldsymbol{x}}.$
We also evaluate CLIPure against the \textbf{BPDA} (short for Backward Pass Differentiable Approximation) with \textbf{EOT} (Expectation Over Transformation) \citep{hill2020stochastic} and latent-based attack \citep{rombach2022high} detailed in Appendix~\ref{sec: detailed setting of attacks}.

% We evaluate our CLIPure and baselines against adaptive attacks with AutoAttack \citep{croce2020reliable}, BPDA with EOT \citep{athalye2018obfuscated}, and latent-based attack \citep{shukla2023generating}, detailed in Section~\ref{sec: detailed setting of attacks} in Appendix~\ref{appendix: experimental settings}.

% \begin{table}[t]
% \centering
% \caption{Comparison with the top-ranked models in their respective settings on RobustBench \citep{croce2020reliable} for CIFAR-10 dataset against $\ell_{\infty}$ threat model ($\epsilon=8/255$) and $\ell_2$ threat model ($\epsilon=0.5$), as well as ImageNet against $\ell_{\infty}$ threat model ($\epsilon=4/255$).}
% \label{table: robustbench results}
% \begin{tabular}{lccccccc}
% \toprule
% \multirow{2}{*}{Method} & \multicolumn{2}{c}{CIFAR10} &  \multicolumn{2}{c}{CIFAR10} &  \multicolumn{2}{c}{ImageNet} \\
% \cmidrule(lr){2-3} \cmidrule(lr){4-5} \cmidrule(lr){6-7}
%  & Clean acc & $\ell_{\infty}=8/255$ & Clean acc & $\ell_{2}=0.5$ &  Clean acc & $\ell_{\infty}=4/255$ \\
% \midrule 
% Rank \#1 & 93.7 & 73.7 & 95.5 & 85.0 & 78.0 & 59.6 \\
% \bf LLM-DP & 95.2 &  88.0 & 95.2 &  91.3 & 73.1 & 65.0 \\
% \bf LLM-CLIP & 95.6 & \bf 91.1 & 95.6 & \bf 91.9 & 76.3 & \bf 72.6 \\
% \bottomrule
% \end{tabular}
% \end{table}

\begin{table}[t]
\caption{Comparison of performance against AutoAttack under $\ell_{\infty}$ ($\epsilon=8/255$) and $\ell_2$ ($\epsilon=0.5$) threat model on CIFAR-10 dataset, showcasing various defense methods including adversarial training and purification mechanisms. We highlight defenses that operate in pixel or latent space (``defense space''), and indicates the modal information used in defense strategies with ``V'' for Vision and ``V-L'' for Vision-Language multimodal representations. We use underlining to highlight the best robustness for baselines, and bold font to denote the state-of-the-art (SOTA) across all methods.}
\label{table: main result on CIFAR10}
\begin{center}
\setlength{\tabcolsep}{2.3pt}
\begin{tabular}{clcc|ccc}
% \hline 
% \multirow{4}{*}{\makecell[c]{\text{Adv.} \\ \text{Train}}} 
\toprule
& \multirow{2}{*}{\textbf{Method}} & \multirow{2}{*}{\makecell[c]{\textbf{Defense} \\ \textbf{Space}}} & \multirow{2}{*}{\textbf{\makecell[c]{\text{Latent} \\ \text{Modality}}}} & \multirow{2}{*}{\textbf{\makecell[c]{\text{Clean} \\ \text{Acc (\%)}}}} & \multicolumn{2}{c}{\textbf{Robust Acc (\%)}} \\
& & & & & $\ell_\infty=8/255$ & $\ell_2=0.5$ \\ 
\midrule
\multirow{3}{*}{\makecell[c]{\text{w/o} \\ \text{Defense}}} & WRN-28-10 \citep{zagoruyko2016wide}& - & V & 94.8 & 0.0&0.0\\
& StableDiffusion \citep{li2023your} & - &V & 87.8 & 0.0 & 38.8 \\
& CLIP \citep{radford2021learning} & - & V-L & 95.2 & 0.0 & 0.0 \\
\hline
\multirow{4}{*}{\makecell[c]{\text{Adv.} \\ \text{Train}}} & AT-DDPM-$\ell_{2}$ \citep{rebuffi2021fixing}& Pixel & V & 93.2 & 49.4 & 81.1 \\
& AT-DDPM-$\ell_{\infty}$ \citep{rebuffi2021fixing} & Pixel & V & 88.8 & 63.3 & 64.7 \\
& AT-EDM-$\ell_{2}$ \citep{wang2023better}& Pixel & V & 95.9 & 53.3 & \underline{84.8} \\
& AT-EDM-$\ell_{\infty}$ \citep{wang2023better} & Pixel & V & 93.4 & 70.9 & 69.7 \\
\hline
\multirow{2}{*}{Other}& TETRA \citep{blau2023classifier} & Pixel & V & 88.2 & 72.0 & 75.9 \\
& RDC \citep{chen2023robust} & Pixel & V & 89.9 & \underline{75.7} &82.0 \\
% ARL \citep{bartoldson2024adversarial} & Pixel & Vision & 93.7 & 73.7 &  \\
% MeanSparse \citep{amini2024meansparse}& Pixel & Vision & 93.2 & 72.1 & \\
% FARE \citep{schlarmann2024robust} & Latent & Vision-Language & 77.7 & 5.1 & 30.6 \\
% TeCoA \citep{mao2022understanding} & Pixel & Vision-Language & 79.6 & 8.1 & 32.4 \\
\hline
\multirow{5}{*}{Purify} & LM - StableDiffusion & Latent & V & 37.9 & 6.9 & 8.6 \\
& DiffPure \citep{nie2022diffusion} & Pixel & V & 90.1 & 71.3 & 80.6 \\
& LM - EDM \citep{chen2023robust} & Pixel & V & 87.9 & 71.7 & 75.0 \\
% LM - Diff & Pixel & Vision & & & \\
& \textbf{Our CLIPure - Diff} & Latent & V-L & 95.2 & \bf 88.0 & \bf 91.3 \\
& \textbf{Our CLIPure - Cos} & Latent & V-L & 95.6 & \bf 91.1 & \bf 91.9 \\
\bottomrule
\end{tabular}
% \vspace{-2mm}
\end{center}
\end{table}

\subsection{Main Results}
In this section, we compare CLIPure-Diff and CLIPure-Cos with SOTA methods and examine the model robustness from various perspectives. 
% In this section, we compare our CLIPure approach, which includes both the CLIPure-Diff and CLIPure-Cos versions, with top-ranked methods on RobsutBench and baselines.
% Importantly, our approach achieves efficient inference capabilities without relying on generative models. Moreover, our CLIPure uses the off-the-shelf CLIP model, fully preserving its zero-shot capabilities and simultaneously significantly improving its robustness, thus marking a significant advancement in addressing the long-standing issue of neural network vulnerability to adversarial attacks.

% \textbf{Ranking on RobustBench.} As for overall performance, we benchmark against the top-ranked models on the well-recognized RobustBench \citep{croce2020reliable} for CIFAR-10 and ImageNet datasets in Table~\ref{table: robustbench results}. Both versions of CLIPure demonstrate remarkable performance. Notably, CLIPure-Cos reduces the gap between robustness and clean accuracy to only 4.4\% under the $\ell_{\infty}$ attack setting on CIFAR-10. On the ImageNet dataset, CLIPure-Cos significantly narrows this gap to just 4.3\% from the clean accuracy benchmark of 76.3\%. 

\textbf{Discussion of CLIPure-Diff and CLIPure-Cos.} We compare the performance of CLIPure-Diff and CLIPure-Cos against AutoAttack across CIFAR-10, ImageNet, and 13 datasets in Tables~\ref{table: main result on CIFAR10}, \ref{table: main result on imagenet}, and \ref{table: zero-shot results}, respectively.
% , as well as defense against BPDA+EOT and latent-based attack in Table~\ref{table: BPDA+EOT on CIFAR10} and \ref{table: latent attack on CIFAR10} in Appendix~\ref{appendix: more expxperimental results}. 
Results indicate that both models have achieved new SOTA performance on all the datasets and CLIPure-Cos uniformly outperforms CLIPure-Diff in clean accuracy and robustness. It is probably because the DiffusionPrior component used in CLIPure-Diff models the generation process by adding noise to the original image embeddings encoded by CLIP without diminishing the effect of vector magnitude. 
% This underperformance of CLIPure-Diff is likely due to its latent diffusion model (LDM) applying noise directly to image embeddings without eliminating the inference of vector magnitude, unlike CLIPure-Cos, which optimizes semantic direction, aligning with CLIP's cosine similarity approach. 
Specifically, the noise is added as $\boldsymbol{z}^i_{t} = \sqrt{\bar{\alpha}} \boldsymbol{z}^i_{0} + \sqrt{1-\bar{\alpha}} \boldsymbol{\epsilon}, \boldsymbol{\epsilon} \sim \mathcal{N}(\boldsymbol{0}, \boldsymbol{I})$. We expect that the performance will be boosted if the generation process also eliminates the effect of vector length.  
In contrast, CLIPure-Cos has no such issue and it models the likelihood with cosine similarities that are consistent with CLIP. 

\textbf{Comparisons with Purification in Pixel Space.}
Compared to the SOTA purification methods
% of adversarial samples has traditionally been conducted in pixel space via a generative model, with the current state-of-the-art method 
, DiffPure \citep{nie2022diffusion} and LM-EDM \citep{chen2023robust} (the likelihood maximization approach based on the advanced diffusion model EDM \citep{karras2022elucidating}), our CLIPure achieves better adversarial robustness (shown in Table \ref{table: main result on CIFAR10} and \ref{table: main result on imagenet}) as well as superior inference efficiency (shown in Table \ref{table: inference time}). Table~\ref{table: main result on CIFAR10} shows that on CIFAR10 under the $\ell_\infty$ and $\ell_2$ threat models, our CLIPure-Cos showed improvements of 27.1\% and 22.5\% over LM-EDM and improvements of 27.8\% and 14.0\% over DiffPure. On the ImageNet dataset, shown in Table~\ref{table: main result on imagenet}, CLIPure-Cos achieves a relative increment of 288.2\% over LM-EDM and 63.5\% over DiffPure. Additionally, the inference efficiency of our CLIPure is multiple orders of magnitude higher than DiffPure and LM-EDM (see Table \ref{table: inference time}). 
% , details of which can be seen in Table~\ref{table: inference time} in Section~\ref{section: inference efficiency}.

Moreover, in Table~\ref{table: main result on imagenet}, we also compared the pixel space purification based on the Decoder component of DaLLE-2 \citep{ramesh2022hierarchical} (LM-DaLLE2.Decoder).
% that models that generation of an image based on the latent vector output by DiffusionPrior component.
% where the Decoder modeled the pixel distribution of images under the guidance of image semantics.
% We utilized an unconditional diffusion model to estimate the log-likelihood $\log p(\boldsymbol{x})$ by dropping the guidance signal for purification in pixel space. 
Results show that the LM-DaLLE2.Decoder suffers from a drop in accuracy, possibly because the diffusion architecture ADM \citep{dhariwal2021diffusion} it employs is slightly inferior to EDM in terms of generation quality.
% The drop in clean accuracy is a common issue faced by pixel space methods, but our LLM approach has shown significant mitigation, even achieving clean accuracy on the ImageNet dataset that is comparable to discriminative models (WideResNet-50).

\textbf{Comparisons with Purification in Uni-modal Latent Space.}
We evaluate latent space purification using Stable Diffusion \citep{rombach2022high} (i.e., LM-StableDiffusion) on the CIFAR-10 dataset, with results detailed in Table~\ref{table: main result on CIFAR10}. As discussed in Section~\ref{section: purification in CLIP latent space}, Stable Diffusion (SD) models an uni-modal image latent space through its VQVAE \citep{rombach2022high}. Direct comparisons are challenging due to SD and CLIP utilizing different training data and fundamentally distinct backbones (diffusion model versus discriminative model). As noted in \citet{li2023your}, SD, as an generative model, is designed for generation rather than classification tasks, so its zero-shot classification performance is not good enough, which limits its potential for robustness. 
% Consequently, while LM-StableDiffusion experiences a significant drop in clean accuracy, it indicates potential for improvement.

\textbf{Comparisons with CLIP-based Baselines.}
Figure \ref{figure:13-zeroshot} and Table~\ref{table: zero-shot results} show the zero-shot adversarial robustness on 13 datasets, compared to CLIP-based baselines enhanced with adversarial training, i.e., FARE \citep{schlarmann2024robust} and TeCoA \citep{mao2022understanding}, CLIPure-Diff and CLIPure-Cos surpass their best-reported average robustness by significant margins, 39.4\% and 45.9\% when $\ell_{\infty}=2/255$, 95.4\% and 108\% when the attacks are stronger with $\ell_{\infty}=4/255$. FARE and TeCoA often have much lower clean accuracy than vanilla CLIP due to their adversarial training on ImageNet, which harms zero-shot performance on other datasets.  
Additionally, Table \ref{table: main result on imagenet} shows that even on ImageNet against the attacks they have been trained with, CLIPure still outperforms them by a huge margin (65\% and 72.6\% versus 33.3\% and 44.3\%). 
The robustness of CLIPure against unseen attacks is much better than their performance on seen attacks, showing that we are on the right path of leveraging the power of pre-trained vision-language models.

\textbf{More Experiments and Analysis.} Due to space constraints, in the Appendix, we include a detailed case study, showcasing the visualization of image embeddings during the purification process using a diffusion model in Figure~\ref{figure: purification process shown by generated images} in Appendix~\ref{section: case study}. Figure~\ref{subfigure: combination} in Appendix~\ref{section: combination} illustrates the effects of combining our approach with adversarial training and pixel space purification methods, while Figure~\ref{subfigure: guidance} displays the outcomes of integrating classifier guidance. Additionally, we employ T-SNE to visualize the distribution of image and text embeddings in CLIP's latent space in Figure~\ref{subfigure: polar plot of embeddings} and analyze the impact of step size on performance in Figure~\ref{subfigure: hyperparam}.

\begin{table}[t]
% \vspace{-2mm}
\caption{Performance comparison of defense methods on ImageNet against AutoAttack with $\ell_{\infty}$ threat model ($\epsilon=4/255$). Indicates whether the methods use ImageNet training set for training as zero-shot. ``V'' stands for Vision, and ``V-L'' for Vision-Language multimodal representations.}
\label{table: main result on imagenet}
\begin{center}
\setlength{\tabcolsep}{4pt}
\begin{tabular}{clccc|cc}
% \hline 
% \multirow{4}{*}{\makecell[c]{\text{Adv.} \\ \text{Train}}} 
% \multirow{2}{*}{\textbf{Method}} & \multirow{2}{*}{\makecell[c]{\text{Defense} \\ \text{Space}}} & \multirow{2}{*}{\textbf{\makecell[c]{\text{Latent} \\ \text{Modality}}}} & \multirow{2}{*}{\textbf{\makecell[c]{\text{Clean} \\ \text{Acc (\%)}}}} & \multicolumn{2}{c}{\textbf{Robust Acc (\%)}} \\
% & & & & \textbf{AA }\bm{$\ell_\infty$} & \textbf{AA }\bm{$\ell_2$} \\ 
\toprule
&\bf Method & \bf \makecell[c]{\text{Defense} \\ \text{Space}} & \bf \makecell[c]{\text{Latent} \\ \text{Modality}}  & \bf \makecell[c]{\text{Zero} \\ \text{-Shot}} & \bf \textbf{\makecell[c]{\text{Clean} \\ \text{Acc (\%)}}} & \textbf{\makecell[c]{\text{Robust} \\ \text{Acc (\%)}}} \\
\midrule
\multirow{2}{*}{\makecell[c]{\text{w/o} \\ \text{Defense}}} & WideResNet-50 \citep{zagoruyko2016wide} & - & V& \ding{55} & 76.5 & 0.0 \\
& CLIP \citep{radford2021learning} & - & V-L &\checkmark & 74.9 & 0.0 \\
\hline
% ARL \citep{bartoldson2024adversarial} & Pixel & Vision & &  \\
\multirow{4}{*}{\makecell[c]{Adv.\\ Train}}& FARE \citep{schlarmann2024robust} & Latent & V-L &\ding{55} & 70.4 & 33.3 \\
& TeCoA \citep{mao2022understanding} & Latent & V-L &\ding{55} & 75.2 & 44.3 \\
& AT-ConvNeXt-L \citep{singh2024revisiting} & Pixel & V &\ding{55} & 77.0 & 57.7\\
& AT-Swin-L \citep{liu2024comprehensive} & Pixel & V &\ding{55} & 78.9 & \underline{59.6} \\
\hline
\multirow{2}{*}{Others} & MixedNUTS \citep{bai2024mixednuts}& Pixel & V &\ding{55} & 81.5 & 58.6 \\
& MeanSparse \citep{amini2024meansparse}& Pixel & V &\ding{55} & 78.0 & \underline{59.6} \\
\hline
\multirow{5}{*}{Purify} & LM - DaLLE2.Decoder & Pixel & V-L & \checkmark &36.9 & 9.2 \\
& DiffPure - DaLLE2.Decoder & Pixel & V-L & \checkmark & 31.2 & 9.0 \\
& LM - EDM \citep{chen2023robust} & Pixel & V &\ding{55} & 69.7 & 18.7\\
& DiffPure \citep{nie2022diffusion} & Pixel & V &\ding{55} & 71.2 & 44.4\\
% LM - Diff & Pixel & Vision & & & \\
& \textbf{Our CLIPure - Diff} & Latent & V-L &\checkmark & 73.1 & \bf 65.0 \\
& \textbf{Our CLIPure - Cos} & Latent & V-L &\checkmark & 76.3 & \bf 72.6 \\
\bottomrule
\end{tabular}
\vspace{-4mm}
\end{center}
\end{table}

% \subsection{Analysis}
% \label{section: analysis}

% \vspace{-2mm}
\subsection{Inference Efficiency}
\label{section: inference efficiency}
We evaluate the inference efficiency of our CLIPure and baseline models by measuring the inference time of an averaged over 100 examples from the CIFAR-10 dataset on a single 4090 GPU. The results are displayed in Table~\ref{table: inference time}. Our CLIPure-Cos innovatively uses a discriminative approach for adversarial sample purification, achieving inference times comparable to those of discriminative models.
Traditional purification methods typically leverage a generative model like diffusion and significantly involve complex inference procedures. Such complexity limits their applicability in efficiency-sensitive tasks, such as autonomous driving. 
% It also demonstrates robust performance under actual adversarial attacks and does not require additional training, thus offering superior advantages in terms of efficient and robust inference.

\subsection{CLIP Variants}
\begin{figure}[t]
\centering
% \captionsetup{labelfont={color=blue}}
% \newlength{\commonheight}
% \setlength{\commonheight}{3.5cm}
\begin{subfigure}[b]{0.49\linewidth}
    \includegraphics[width=\linewidth]{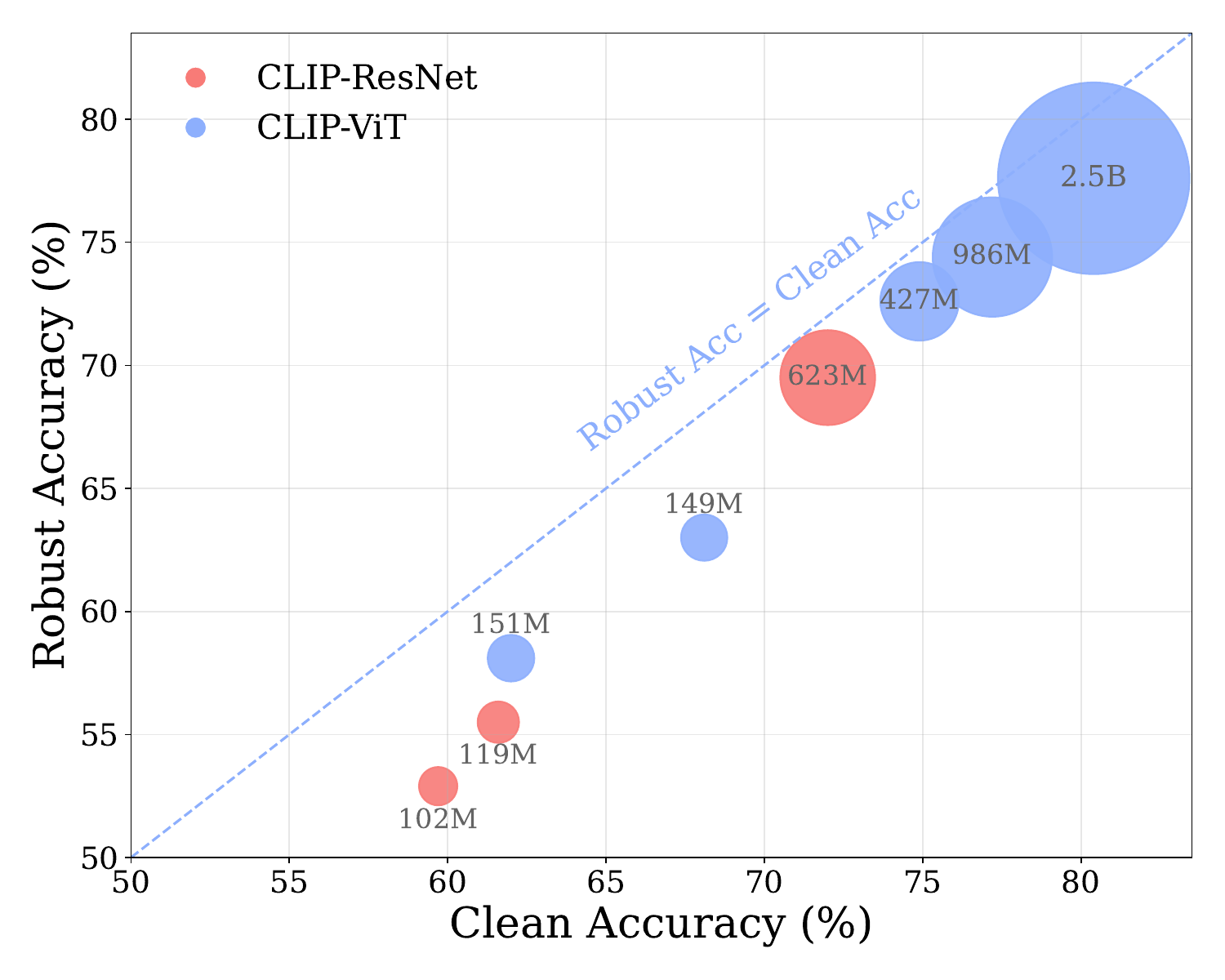}
    % \caption{\color{blue}{Different Version of CLIP}}
    \label{subfigure: backbone_CLIP}
\end{subfigure}
% \hfill  % Add space between subfigures
\begin{subfigure}[b]{0.49\linewidth}  % Width set to one third of text width
    \includegraphics[width=\linewidth]{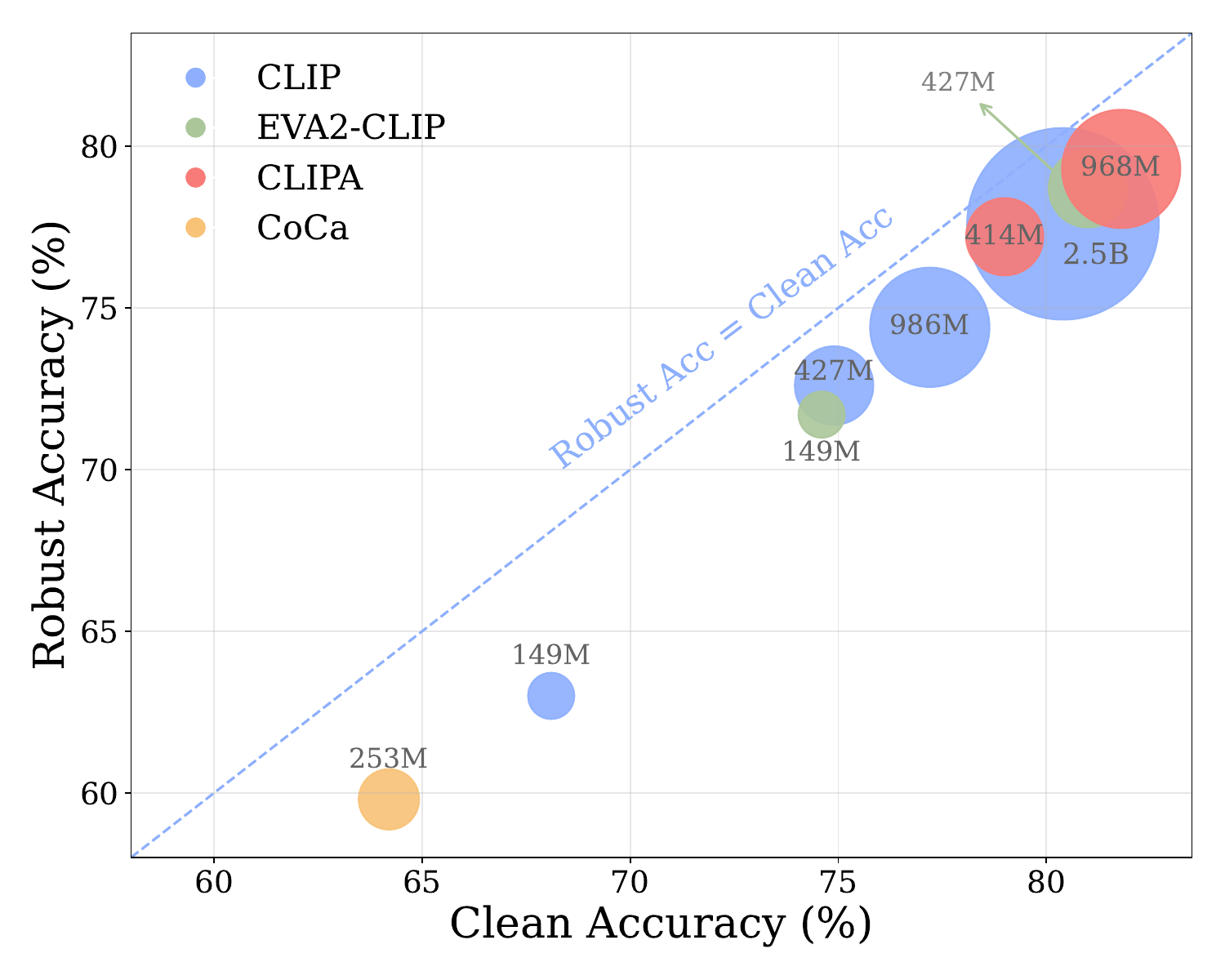}
    % \caption{\color{blue}{Different backbone of CLIPure}}
    \label{subfigure: backbones}
\end{subfigure}
% \vspace{-3mm}
\caption{Accuracy and robustness (detailed in Table~\ref{table: performance across backbones}) of CLIPure-Cos for (Left) various versions of CLIP and (Right) different backbone models. The bubble size represents the number of parameters, which is also indicated alongside each bubble. The left figure presents CLIPure-Cos based on ResNet-based models (including RN50, RN101, RN50x64, marked in red) and ViT-based models (ViT-B-16, ViT-B-32, ViT-L-14, ViT-H-14, and ViT-bigG-14, marked in blue). The right figure depicts CLIPure-Cos based on CLIP (including ViT-B-16, ViT-L-14, ViT-H-14, ViT-bigG-14, marked in blue), EVA2-CLIP (including ViT-B-16 and ViT-L-14, marked in green), CLIPA (including ViT-L-14 and ViT-H-14, marked in red), and CoCa (ViT-B-32, marked in yellow). 
The blue dashed line represents the point where robust accuracy equals clean accuracy, serving as the upper bound of robustness, since successful defense against adversarial attacks hinges on accurate classification.
}
\label{figure: backbones for CLIPure}
% \vspace{-2mm}
\end{figure}

To investigate the influence of backbone models and their scale, we evaluate the performance of CLIPure-Cos across various versions of CLIP \citep{radford2021learning} and its variants, including EVA2-CLIP \citep{sun2023eva}, CLIPA \citep{li2024inverse}, and CoCa \citep{yu2205coca}. The results in Table~\ref{table: performance across backbones} and Figure~\ref{figure: backbones for CLIPure} indicate: 
1) Larger models generally exhibit stronger capabilities (i.e., clean accuracy), which in turn enhances the robustness of CLIPure. Notably, the CLIPure-Cos with CLIPA with ViT-H-14 achieves the best performance, with robustness reaching 79.3\% on ImageNet—a truly remarkable achievement.
2) CLIPA-based CLIPure-Cos outperforms other CLIP models of comparable size in both accuracy and robustness.
3) ViT-based CLIPure-Cos models show better results than those based on ResNet.

\begin{table}[t]
\centering
% \caption{Comparison of different methods in terms of inference time. Dis. denotes the purification with a discriminative model, and Gen. denotes that with a generative model.}
\caption{Comparison of different methods in terms of average inference time over 100 samples on CIFAR10. ``Relative Time'' expresses each method's inference time as a multiple of the CLIP model's time for classification. Dis. denotes purification with a discriminative model, and Gen. denotes that with a generative model.}
\label{table: inference time}
\begin{tabular}{lccccc}
\toprule
Method& CLIPure-Cos & CLIPure-Diff & DiffPure & LM-EDM & CLIP  \\ \midrule
Gen. or Dis. & Dis. & Gen. & Gen. & Gen. & Dis.  \\ 
Inference Time (s)& 4.1 x $10^{-4}$ & 0.01 & 2.22 & 0.25 & 3.6 x $10^{-4}$ \\
Relative Time  & 1.14x & 27.78x & 6166.67x & 694.44x & 1x\\
\bottomrule
\end{tabular}
\end{table}

% \textbf{Comparison with Euclidean Space Purification.}
% We compare purification in Cartesian coordinate latent space for both the LLM-Diff and LLM-CLIP models. Extensive experimental searches covered various purification steps and step size parameters, alongside trials of momentum-based optimization methods. We found it challenging to devise an effective strategy for achieving robustness on the ImageNet dataset to exceed 10\%, echoing the difficulties encountered with stable diffusion. We do not deny the feasibility of purification in Euclidean space but we found that purification in polar coordinates easily facilitates effective purification.

% \section{Discussion}

\section{Conclusion}
We develop CLIPure, a novel adversarial purification method that operates within the CLIP model's latent space to enhance adversarial robustness on zero-shot classification without additional training. CLIPure consists of two variants: CLIPure-Diff and CLIPure-Cos, both achieving state-of-the-art performance across diverse datasets including CIFAR-10 and ImageNet. CLIPure-Cos, notably, does not rely on generative models, significantly enhancing defense efficiency. Our findings reveal that purification in a multi-modal latent space holds substantial promise for adversarially robust zero-shot classification, pointing the way for future research that extends beyond image classification.

\subsection*{Acknowledgments}
This work is supported by the Strategic Priority Research Program of the Chinese Academy of Sciences, Grant No. XDB0680101, CAS Project for Young Scientists in Basic Research under Grant No. YSBR-034, the National Key Research and Development Program of China under Grants No. 2023YFA1011602, the National Natural Science Foundation of China (NSFC) under Grants No. 62302486, the Innovation Project of ICT CAS under Grants No. E361140, the CAS Special Research Assistant Funding Project, the project under Grants No. JCKY2022130C039, the Strategic Priority Research Program of the CAS under Grants No. XDB0680102, and the NSFC Grant No. 62441229.

% \clearpage

\bibliography{arxiv}
\bibliographystyle{arxiv}

\clearpage
\appendix

\section{Discussion of the Reverse-Time SDE}
\label{appendix: discussion of the reverse SDE}
% discussion of the form of reverse SDE

\begin{figure}[t]
\centering
\includegraphics[width=\linewidth]{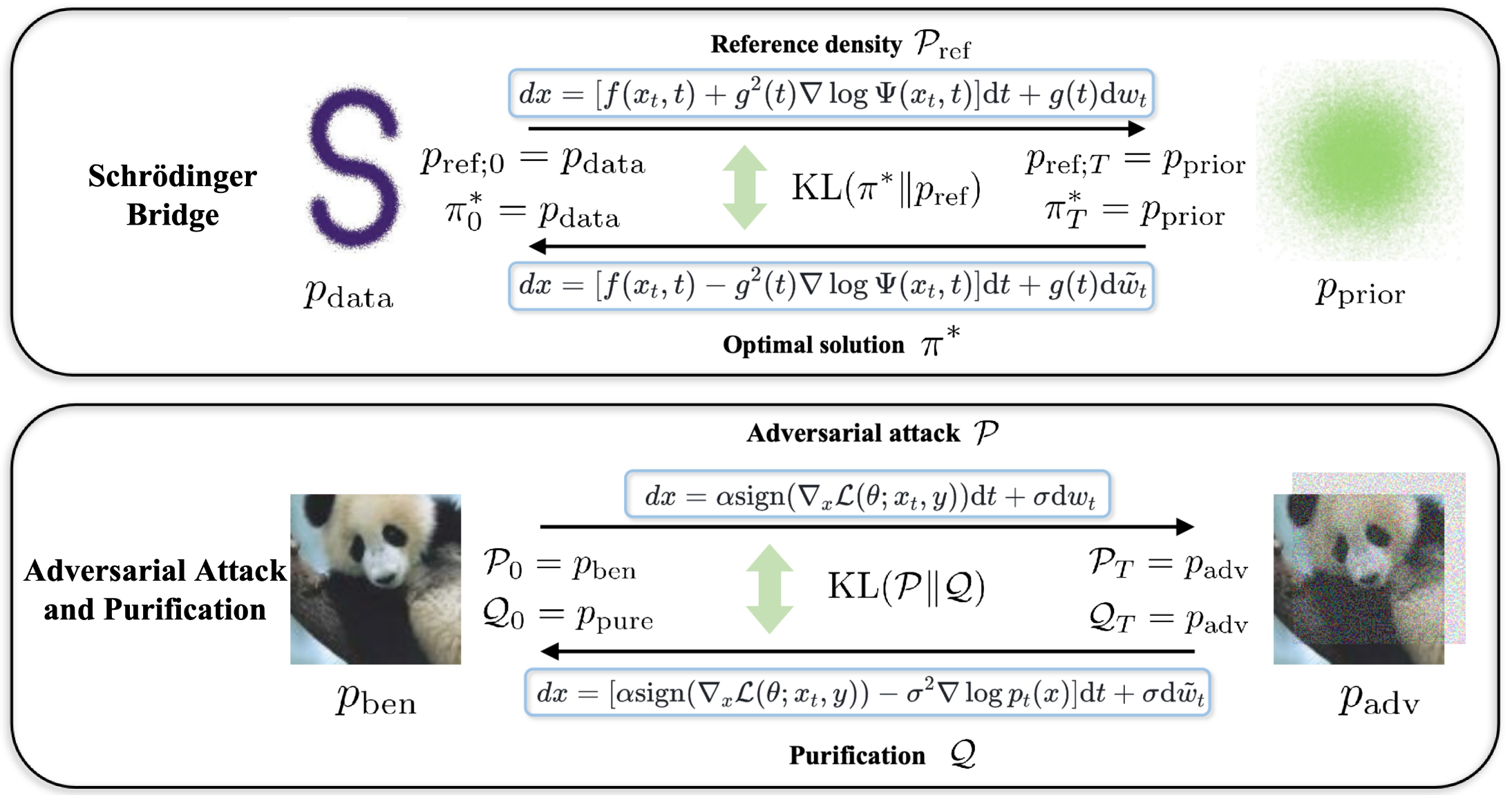} % Adjust the path and options as needed
\caption{Comparison of the Schrödinger Bridge \citep{schrodinger1932theorie} framework with our adversarial attack and purification process}
\label{figure: schrodinger bridge compared with attack and purification process}
\end{figure}

Motivated by the Schrödinger Bridge theory \citep{schrodinger1932theorie, tang2024simplified} which models the transition between two arbitrary distributions, we utilize this theoretical framework to model the transformation between benign and adversarial sample distributions. 

Within the Schrödinger Bridge framework, the forward Stochastic Differential Equation (SDE) describes a pre-defined transition from a data distribution $p_{\text{data}}$ to a prior distribution $p_{\text{prior}}$, noted as a reference density $p_{\text{ref}} \in \mathscr{P}_{T+1}$, the space of joint distributions on $\mathbb{R}^{(T+1)\times d} $, where $T$ represents the timesteps and $d$ denotes the data dimension. Consequently, we have $p_{\text{ref};0}=p_{\text{data}}$ and $p_{\text{ref};T}=p_{\text{prior}}$. Both  $p_{\text{data}}$ and $p_{\text{prior}}$ are defined over 
the data space $\mathbb{R}^d$. The optimal solution $\pi^*$, starting from $p_{\text{prior}}$ and transferring to $p_{\text{data}}$, is termed the Schrödinger Bridge, described by the reverse SDE.

Specifically, the forward SDE is expressed as:
\begin{equation}
\begin{aligned}
\mathrm{d} \boldsymbol{x} = [f(\boldsymbol{x}_t, t) + g^2(t) \nabla \log \Psi(\boldsymbol{x}_t, t)] \mathrm{d} t + g(t) \mathrm{d} \boldsymbol{w}_t, \quad x_0 \sim p_{\text{data}},
\end{aligned}
\end{equation}
where $f(\boldsymbol{x}_t, t)$ is the drift function and $g(t)$ is the diffusion term. $\boldsymbol{w}_t$ is the standard Wiener process, and $\Psi$ and $\tilde{\Psi}$ are time-varying energy potentials constrained by the following Partial Differential Equations (PDEs):
\begin{equation}
\begin{aligned}
& \frac{\partial \Psi}{\partial t} = -\nabla_{\boldsymbol{x}} \Psi^T f - \frac{1}{2} \operatorname{Tr}(g^2 \nabla_{\boldsymbol{x}}^2 \Psi) \\
& \frac{\partial \tilde{\Psi}}{\partial t} = -\nabla_{\boldsymbol{x}} \tilde{\Psi}^T f - \frac{1}{2} \operatorname{Tr}(g^2 \nabla_{\boldsymbol{x}}^2 \tilde{\Psi}),
\label{equation in appendix: SDE constraints}
\end{aligned}
\end{equation}
such that $\Psi(\boldsymbol{x}, 0) \tilde{\Psi}(\boldsymbol{x}, 0) = p_{\text{data}}, \Psi(\boldsymbol{x}, T) \tilde{\Psi}(\boldsymbol{x}, T) = p_{\text{prior}}$. 

Similarly, we define the mutual transformation between clean and adversarial sample distributions using coupled SDEs. The forward SDE describes the transformation process from benign to adversarial example distributions corresponding to an attacker's process. For an untargeted PGD attack, this is represented as:
\begin{equation}
\begin{aligned}
\mathrm{d} \boldsymbol{x} = \alpha \mathrm{sign} (\nabla_{\boldsymbol{x}} \mathcal{L}(\theta; \boldsymbol{x}_t, y_{\text{true}})) \mathrm{d} t + \sigma \mathrm{d} \boldsymbol{w}_t , \quad \boldsymbol{x}_0 \sim p_{\text{ben}},
\label{equation in appendix: forward SDE of untargeted PGD attack}
\end{aligned}
\end{equation}
where $\alpha$ represents the attack step size, $\mathcal{L}(\theta; \boldsymbol{x}_t, y_{\text{true}})$ denotes the loss when the adversarial example $\boldsymbol{x}_t$ is classified by the model with parameters $\theta$ as the ground truth category $y_{\text{true}}$ at attack step $t$ (where $t \in [0, T]$), $\mathrm{d} \boldsymbol{w}_t$ denotes the Wiener process (Brownian motion) to express more general cases, and $\sigma$ is a constant. The adversarial example $\boldsymbol{x}_T$ is bound by $\epsilon$ in $l_p$ norm, i.e., $\|\boldsymbol{x}_T - \boldsymbol{x}_0 \|_p \leq \epsilon$.

According to Eq.~\ref{equation in appendix: forward SDE of untargeted PGD attack}, we obtain formulation of the drift function and diffusion term is
\begin{equation}
\begin{aligned}
f(\boldsymbol{x}_t, t) & = \alpha \mathrm{sign} (\nabla_{\boldsymbol{x}} \mathcal{L}(\theta; \boldsymbol{x}_t, y_{\text{true}})) \\
g(t) & = \sigma
\end{aligned}
\end{equation}

According to \citep{anderson1982reverse, song2020score}, the corresponding reverse-time SDE is:
\begin{equation}
\begin{aligned}
\mathrm{d}\boldsymbol{x} = [f(\boldsymbol{x}_t, t) - g^2(t) \nabla_{\boldsymbol{x}} \log p(\boldsymbol{x}) ] \mathrm{d}t + \sigma \mathrm{d} \tilde{\boldsymbol{w}}_t.
\end{aligned}
\end{equation}

Thus, we can derive the corresponding reverse SDE of Eq.~\ref{equation in appendix: forward SDE of untargeted PGD attack} as
\begin{equation}
\begin{aligned}
dx = [\alpha \mathrm{sign} (\nabla_{\boldsymbol{x}} \mathcal{L}(\theta; \boldsymbol{x}_t, y_{\text{true}})) - \sigma^2 \nabla \log p(\boldsymbol{x}_t)] \mathrm{d} t + \sigma \mathrm{d} \tilde{w}_t, \quad \boldsymbol{x}_T \sim p_{\mathrm{adv}},
\label{equation in appendix: reverse SDE of untargeted PGD attack}
\end{aligned}
\end{equation}
where $\log p(\boldsymbol{x})$ is the log-likelihood of the marginal distribution of benign samples, also known as the score function in Score SDE \citep{song2020score}. $\mathrm{d}\tilde{w}_t$ denotes the reverse-time Wiener process as time flows from $t=T$ to $t=0$.

Note that the coupled SDEs described in Eq.~\ref{equation in appendix: forward SDE of untargeted PGD attack} and Eq.~\ref{equation in appendix: reverse SDE of untargeted PGD attack} are special cases of Eq.~\ref{equation in appendix: SDE constraints} when $\nabla \log \Psi(\boldsymbol{x}_t, t) = 0 $ and $\nabla \log \tilde{\Psi}(\boldsymbol{x}_t, t) = \nabla \log p(\boldsymbol{x}_t)$.

Without loss of generality, other adversarial attack methods can typically be described using the forward SDE. Therefore, our proposed approach of modeling adversarial attacks and purification processes using coupled SDEs is universally applicable.

% definition

% schordinger bridge

% forward and reverse SDE

% special case

% targeted attack

% discussion of targeted attack, other type of attack as well as OOD transformation

\section{Detailed Proof of Purification Risk Function}
\label{appendix: proof of purification quality function}

We first define the joint distribution of the attack process described by the forward SDE in Eq.~\ref{equation: forward sde of untargeted attack} as $\mathcal{P}_{T+1}$, which spans $\mathbb{R}^{(T+1)\times d}$. Thus we have $\mathcal{P}_{0} = p_{\text{ben}}$ and $\mathcal{P}_{T} = p_{\text{adv}}$. For the purification process, we designate the joint distribution as $\mathcal{Q}_{T+1}$, also defined over $\mathbb{R}^{(T+1)\times d}$, with $\mathcal{Q}_{0} = p_{\text{adv}}$ and $\mathcal{Q}_{T} = p_{\text{pure}}$. $T$ represents the timestep, $p_{\text{ben}}$, $p_{\text{adv}}$ and $p_{\text{pure}}$ denote the distribution of benign, adversarial, and purified samples.

According to the results of \citet{leonard2014some}, we have 
\begin{equation}
\begin{aligned}
\mathrm{KL}(\mathcal{Q}_{0:T} | \mathcal{P}_{0:T}) & = \mathrm{KL}(\mathcal{Q}_{0, T} | \mathcal{P}_{0,T}) + \mathbb{E}_{\mathcal{Q}_{0,T}}[\mathrm{KL}(\mathcal{Q}_{1:T-1|0,T} | \mathcal{P}_{1:T-1|0,T})]. \\
% & \geq \mathrm{KL}(\mathcal{Q}_{0, T} | \mathcal{P}_{0,T}),
\end{aligned}
\end{equation}

Since we focus on the differences between the purified and benign examples rather than the entire purification path, we concentrate on $\mathrm{KL}(\mathcal{Q}_{0, T} \| \mathcal{P}_{0,T})$. Thus, we continue to derive:
\begin{equation}
\begin{aligned}
& \mathrm{KL}(\mathcal{Q}_{0, T} \| \mathcal{P}_{0,T}) \\
= & \ \mathrm{KL}(p(\boldsymbol{x}_{\mathrm{adv}}, \boldsymbol{x}_{\mathrm{pure}}) | p_{\mathrm{attack}}(\boldsymbol{x}_{\text{ben}}, \boldsymbol{x}_{\mathrm{adv}}))\\
= & \ \int \int p(\boldsymbol{x}_{\text{adv}}) p(\boldsymbol{x}_{\text{pure}} | \boldsymbol{x}_{\text{adv}}) \log \frac{p(\boldsymbol{x}_{\text{adv}}) p(\boldsymbol{x}_{\text{pure}} | \boldsymbol{x}_{\text{adv}})}{p(\boldsymbol{x}_{\text{ben}}) p(\boldsymbol{x}_{\text{adv}} | \boldsymbol{x}_{\text{ben}})} \, d\boldsymbol{x}_{\text{pure}} \, d\boldsymbol{x}_{\text{adv}} \\
= & \ \int p(\boldsymbol{x}_{\text{adv}}) \left( \int p(\boldsymbol{x}_{\text{pure}} | \boldsymbol{x}_{\text{adv}}) \log \frac{p(\boldsymbol{x}_{\text{pure}} | \boldsymbol{x}_{\text{adv}})}{p(\boldsymbol{x}_{\text{adv}} | \boldsymbol{x}_{\text{ben}})} \, d\boldsymbol{x}_{\text{pure}} \right) d\boldsymbol{x}_{\text{adv}} - \int p(\boldsymbol{x}_{\text{adv}}) \log \frac{p(\boldsymbol{x}_{\text{adv}})}{p(\boldsymbol{x}_{\text{ben}})} \, d\boldsymbol{x}_{\text{adv}} \\
= & \ \mathbb{E}_{\boldsymbol{x}_{\text{adv}}} \left[ \mathrm{KL}(p(\boldsymbol{x}_{\text{pure}} | \boldsymbol{x}_{\text{adv}}) \| p(\boldsymbol{x}_{\text{adv}} | \boldsymbol{x}_{\text{ben}})) \right] - \mathrm{KL}(p(\boldsymbol{x}_{\text{adv}}) \| p(\boldsymbol{x}_{\text{ben}})),
\end{aligned}
\end{equation}
where $p(\boldsymbol{x}_{\text{adv}}|\boldsymbol{x}_{\text{ben}})$ represents the conditional probability transitioning from $\boldsymbol{x}_{\text{ben}}$ to $\boldsymbol{x}_{\text{adv}}$ as described by the forward SDE, and $q(\boldsymbol{x}_{\text{pure}}|\boldsymbol{x}_{\text{adv}})$ represents the conditional probability transitioning from $\boldsymbol{x}_{\text{adv}}$ to $\boldsymbol{x}_{\text{pure}}$ as described by the reverse SDE. This formulation underscores the effectiveness of the purification process by measuring how well the adversarial transformations are reversed.

Given that the perturbations added by attackers are typically imperceptible, over a small time interval $\Delta t$, the solution of the forward SDE described by Eq.~\ref{equation: forward sde of untargeted attack} approximates a normal distribution:
\begin{equation}
\begin{aligned}
p({\boldsymbol{x}_{\text{adv}}|\boldsymbol{x}_{\text{ben}}}) \sim \mathcal{N}(\boldsymbol{x}_{\text{ben}} + \alpha \operatorname{sign}(\nabla_x \mathcal{L}) \Delta t, \sigma^2 \Delta t).
\end{aligned}
\end{equation}
Similarly, the reverse SDE process described by Eq.~\ref{equation: reverse sde of untargeted attack} approximates a normal distribution:
\begin{equation}
\begin{aligned}
p(\boldsymbol{x}_{\text{pure}}|\boldsymbol{x}_{\text{adv}}) \sim \mathcal{N}(\boldsymbol{x}_{\text{adv}} - (\alpha \operatorname{sign}(\nabla_{\boldsymbol{x}_{\text{adv}}} \mathcal{L}(\theta; \boldsymbol{x}_{\text{adv}}, y_{\text{true}}))) - \sigma^2 \nabla \log p(\boldsymbol{x}_{\text{adv}})) \Delta t, \sigma^2 \Delta t).
\end{aligned}
\end{equation}

The KL divergence formula for two normal distributions is:
\begin{equation}
\begin{aligned}
KL(\mathcal{N}(\mu_0, \Sigma_0) \parallel \mathcal{N}(\mu_1, \Sigma_1)) = \frac{1}{2} \left(\operatorname{tr}(\Sigma_1^{-1} \Sigma_0) + (\mu_1 - \mu_0)^T \Sigma_1^{-1} (\mu_1 - \mu_0) - k + \log\frac{\det \Sigma_1}{\det \Sigma_0}\right).
\end{aligned}
\end{equation}
When $\Sigma_0 = \Sigma_1 = \sigma^2 \Delta t$, the KL divergence reduced as
\begin{equation}
\begin{aligned}
KL(\mathcal{N}(\mu_0, \Sigma_0) \parallel \mathcal{N}(\mu_1, \Sigma_1)) = \frac{1}{2 \sigma^2 \Delta t} \|\mu_1 - \mu_0\|^2.
\end{aligned}
\end{equation}
Thus we have
\begin{equation}
\begin{aligned}
& \mathbb{E}_{\boldsymbol{x}_{\text{adv}}} \left[ \mathrm{KL}(p(\boldsymbol{x}_{\text{pure}} | \boldsymbol{x}_{\text{adv}}) \| p(\boldsymbol{x}_{\text{adv}} | \boldsymbol{x})) \right] \\
= & \ \mathbb{E}_{\boldsymbol{x}_{\text{adv}}} \left[ \nabla \log p(\boldsymbol{x}_{\text{ben}})^T \nabla \log p(\boldsymbol{x}_{\text{ben}}) \sigma^2 \Delta t \right].
\end{aligned}
\end{equation}
Thus we have proved the result in Eq.~\ref{equation: purification risk} that
\begin{equation}
\begin{aligned}
& \mathrm{KL}(\mathcal{Q}_{0, T} | \mathcal{P}_{0,T}) 
= \mathrm{KL}(p(\boldsymbol{x}_{\mathrm{adv}}, \boldsymbol{x}_{\text{pure}}) | p(\boldsymbol{x}, \boldsymbol{x}_{\mathrm{adv}}))\\
= & \ \mathbb{E}_{\boldsymbol{x}_{\text{adv}}} \left[ \mathrm{KL}(q(\boldsymbol{x}_{\text{pure}} | \boldsymbol{x}_{\text{adv}}) \| p(\boldsymbol{x}_{\text{adv}} | \boldsymbol{x}_{\text{ben}})) \right] - \mathrm{KL}(p(\boldsymbol{x}_{\text{adv}}) \| p(\boldsymbol{x})) \\
= & \frac{1}{2} \mathbb{E}_{\boldsymbol{x}_{\text{adv}}} \left[ \nabla \log p(\boldsymbol{x}_{\text{adv}})^T \nabla \log p(\boldsymbol{x}_{\text{adv}}) \sigma^2 \Delta t \right] - \mathrm{KL}(p(\boldsymbol{x}_{\text{adv}}) \| p(\boldsymbol{x}_{\text{ben}})).
\end{aligned}
\end{equation}

% \section{Proof of Purification Risk in Latent Space}
% \label{appendix: proof of purification risk in latent space}
% In this section, we make a further comparison of the purification risk between pixel space and latent space.

% According to the result in Eq.~\ref{equation: purification risk} that the risk of purification process $\mathcal{Q}$ is defined as
% \begin{equation}
% \begin{aligned}
% \mathcal{R}(\mathcal{Q}) = \mathbb{E}_{\boldsymbol{x}_{\text{adv}}} \left[ \nabla \log p(\boldsymbol{x})^T \nabla \log p(\boldsymbol{x}) \sigma^2 \Delta t \right],
% \end{aligned}
% \end{equation}
% where $\boldsymbol{x}_{\text{adv}}$ represents the adversarial examples, $\log p(\boldsymbol{x})$ is the log-likelihood of $\boldsymbol{x}$, $\sigma$ denotes the scale factor of Wiener process in Eq.~\ref{equation: forward sde of untargeted attack}, $\Delta t $ is the small time interval related to the perturbation.

% % 借用flow-based model对隐空间与原始空间

\begin{figure}[t]
\centering
% \captionsetup{labelfont={color=blue}}
\includegraphics[width=\linewidth]{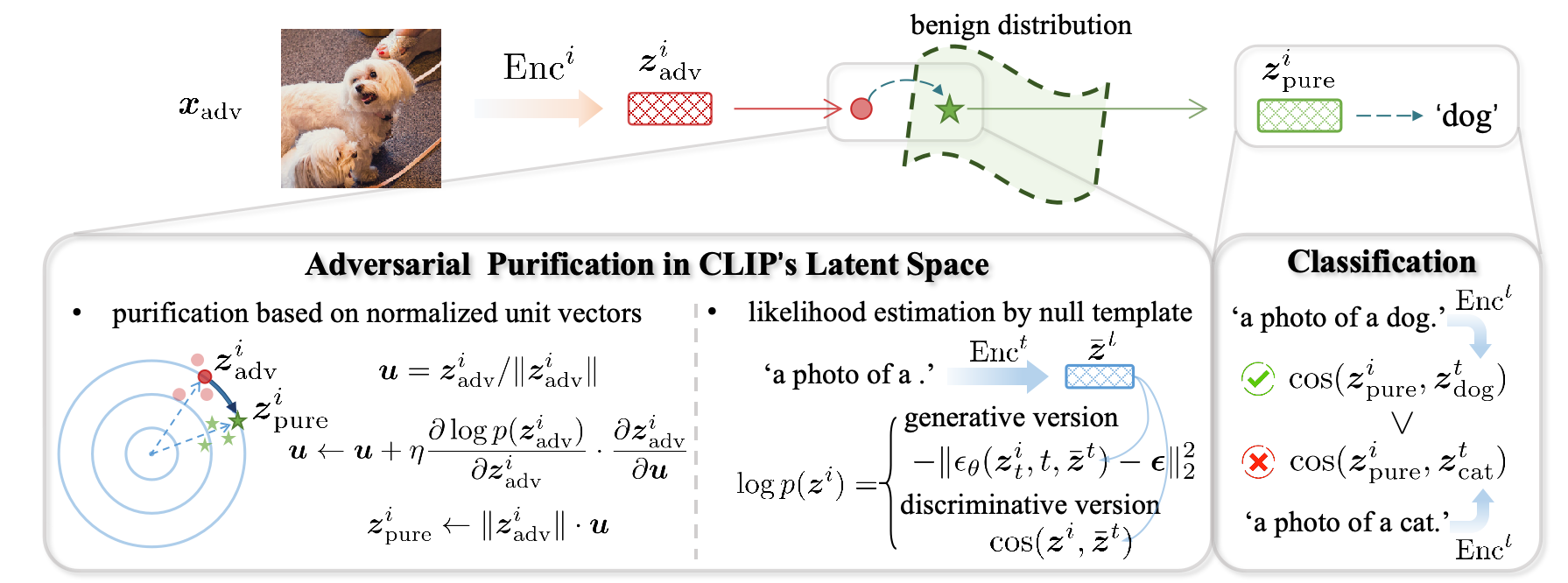} % Adjust the path and options as needed
\caption{Illustration of the process of CLIPure including the purification in latent space and zero-shot classification, detailed in Algorithm~\ref{algorithm: latent purification by CLIP and DaLLE2.DiffusionPrior}.}
\label{figure: CLIPure process}
\end{figure}

\section{Experimental Setup}
\label{appendix: experimental settings}

\subsection{Implementation Detail}
\label{section: purification settings}
Following the implementation of pixel space purification (including \citet{chen2023robust} with \href{https://github.com/huanranchen/DiffusionClassifier/blob/master/defenses/PurificationDefenses/DiffPure/LikelihoodMaximization/EDMLM.py}{code} and \citet{zhang2023enhancing} with \href{https://github.com/zzzhangboya/ScoreOpt/blob/fcaae8ad68041b468fd92fdbac1c36ba791703ee/defense.py}{code} ), CLIPure employs gradient ascent to maximize the log-likelihood of samples, which is formulated through: (1) cosine similarity (CLIPure-Cos) and (2) mean squared error (CLIPure-Diff). 
We conduct 10 purification steps on image embeddings obtained from CLIP's image encoder, with a step size of 30 for both CLIPure-Cos and CLIPure-Diff, to update directly on the vector direction of image embeddings. For CLIPure-Diff, we use the off-the-shelf DiffusionPrior model from DaLLE 2 \citep{ramesh2022hierarchical}. Experiments indicate that timesteps in the range of 900 to 1000 yield better defense outcomes when estimating likelihood via DaLLE2.Diffusion, hence we select this range for estimation.

For the blank template used in purification, we employ 80 diverse description templates combined with class names, such as ``a \textit{good} photo of $<$class-name$>$'' to enhance stability, following zero-shot classification strategies. Consistently, each class $c$'s text embedding, $\boldsymbol{z}^t_c$, as referred to in Eq. \ref{equation:zero-shot-classification}, is computed as the average text embedding combined with all templates.

\subsection{Datasets for Zero-Shot Performance}
\label{appendix: zero-shot datasets}
In order to evaluate the zero-shot performance of our CLIPure and adversarially trained CLIP model including FARE \citep{schlarmann2024robust} and TeCoA \citep{mao2022understanding}, we follow the settings and evaluation metrics used by FARE and outlined in the CLIP-benchmark\footnote{Available at \url{https://github.com/LAION-AI/CLIP_benchmark/}}. We evaluate the clean accuracy and adversarial robustness across 13 datasets against an $\ell_{\infty}$ threat model with $\epsilon=4/255$ and $\epsilon=2/255$. The datasets for zero-shot performance evaluation include CalTech101 \citep{griffin2007caltech}, StanfordCars \citep{krause20133d}, CIFAR10, CIFAR100 \citep{krizhevsky2009learning}, DTD \citep{cimpoi2014describing}, EuroSAT \citep{helber2019eurosat}, FGVC Aricrafts \citep{maji2013fine}, Flowers \citep{nilsback2008automated}, ImageNet-R \citep{hendrycks2021many}, ImageNet-S \citep{wang2019learning}, PCAM \citep{veeling2018rotation}, OxfordPets \citep{parkhi2012cats}, and STL10 \citep{coates2011analysis}.

\subsection{Baselines}
\label{sec: detailed baselines in appendix}
In this section, we discuss the detailed experimental settings of the baselines mentioned in Section~\ref{section: experimental settings}.
For the CIFAR-10 dataset results shown in Table~\ref{table: main result on CIFAR10}, we use the off-the-shelf Stable Diffusion model as provided by \citet{li2023your}, employing it as a zero-shot classifier. Our LM-StableDiffusion method adapts this model to a likelihood maximization approach. Other methods were applied directly according to the model in the original papers for CIFAR-10 without additional training.

For the ImageNet dataset experiments detailed in Table~\ref{table: main result on imagenet}, FARE \citep{schlarmann2024robust} and TeCoA \citep{mao2022understanding} involve testing checkpoints that were adversarially trained on the ImageNet dataset specifically for an $\ell_{\infty}$ threat model with $\epsilon=4/255$. The LM-DaLLE2.Decoder method utilizes the Decoder module from the ViT-L-14 model of DaLLE2 \citep{ramesh2022hierarchical} provided by OpenAI, adapted to the likelihood maximization method. DiffPure-DaLLE2.Decoder adapts the same Decoder module to the DiffPure \citep{nie2022diffusion} purification method. Other methods use models as provided and trained in the original papers.

Regarding the zero-shot performance across 13 datasets shown in Table~\ref{table: zero-shot results}, we opt for the adversarially trained models of TeCoA \citep{mao2022understanding} and FARE \citep{schlarmann2024robust} with an $\ell_{\infty}$ threat model at $\epsilon=4/255$, as they exhibited superior performance compared to those trained with $\epsilon=2/255$.

% In the results presented in Table~\ref{table: BPDA+EOT on CIFAR10} and Table~\ref{table: latent attack on CIFAR10}, we conduct an adversarial evaluation using models as specified in the literature without any additional fine-tuning.

\color{black}{}
\section{More Experimental Results}
\label{appendix: more expxperimental results}

\begin{table}[t]
% \captionsetup{labelfont={color=blue}}
\caption{Performance of CLIPure-Cos based on various versions of CLIP \citep{radford2021learning}, EVA2-CLIP \citep{sun2023eva}, CLIPA \citep{li2024inverse}, and CoCa \citep{yu2205coca} under the $\ell_{\infty}$ threat model ($\epsilon=4/255$) on the ImageNet dataset. "Param" denotes the number of parameters. The prefix "RN" refers to ResNet-based \citep{he2016deep} methods, while "ViT" indicates Vision Transformer-based \citep{alexey2020image} approaches.}
\label{table: performance across backbones}
\begin{center}
\setlength{\tabcolsep}{4pt}
\begin{tabular}{ccl|cccc}
% \hline 
% \multirow{4}{*}{\makecell[c]{\text{Adv.} \\ \text{Train}}} 
\toprule
\multirow{2}{*}{\textbf{Model}} & \multirow{2}{*}{\textbf{Version}} & \multirow{2}{*}{\textbf{Param (M)}} & \multicolumn{2}{c}{\textbf{w/o Defense}}  & \multicolumn{2}{c}{\textbf{CLIPure}} \\
& & & Acc (\%) & Rob (\%) & Acc (\%) & Rob (\%) \\ 
\midrule
\multirow{8}{*}{\makecell[c]{\text{CLIP} \\ \text{ \citep{radford2021learning}}}} & RN50 & 102 & 59.7& 0.0& 60.0& 52.9\\
 &RN101 & 119       & 61.6        & 0.0               & 61.9             & 55.5          \\
&RN50x64         & 623       & 72.0        & 0.0               & 72.3             & 69.5          \\
&ViT-B-16        & 149       & 68.1        & 0.0               & 68.2             & 63.0          \\
&ViT-B-32        & 151       & 62.0        & 0.0               & 62.0             & 58.1          \\
&ViT-L-14        & 427       & 74.9        & 0.0               & 76.3             & 72.6          \\
&ViT-H-14        & 986       & 77.2        & 0.0               & 77.4             & 74.4          \\
&ViT-bigG-14    & 2539      & 80.4        & 0.0               & 80.4             & 77.6          \\
\hline
\multirow{2}{*}{\makecell[c]{\text{EVA2-CLIP} \\ \text{ \citep{sun2023eva}}}}& ViT-B-16      & 149       & 74.6        & 0.0               & 74.7             & 71.7          \\
&ViT-L-14      & 427       & 81.0        & 0.0               & 80.7             & 78.7          \\
\hline
\multirow{2}{*}{\makecell[c]{\text{CLIPA} \\ \text{ \citep{li2024inverse}}}}&ViT-L-14 & 414 & 79.0 & 0.0 & 79.0 & 77.2 \\
&ViT-H-14  & 968       & 81.8        & 0.0               & \bf 81.5             & \bf 79.3          \\
\hline
\makecell[c]{\text{CoCa} \\ \text{ \citep{yu2205coca}}} & ViT-B-32 & 253 &64.2&0.0&63.8&59.8\\
\bottomrule
\end{tabular}
\end{center}
\end{table}

\begin{table}[t]
\caption{Zero-shot performance on 13 datasets against AutoAttack under $\ell_{\infty}$ threat model with $\epsilon=2/255$ and $\epsilon=4/255$. FARE \citep{schlarmann2024robust} and TeCoA \citep{mao2022understanding} are trained on $\ell_{\infty}$ threat model with $\epsilon=4/255$ for its generally better performance than $\epsilon=4/255$. Underlined results indicate the best robustness among baselines, while bold text denotes state-of-the-art (SOTA) performance across all methods. The term \textcolor{blue}{increase} represents the percentage improvement in robustness compared to the best baseline method.}
\label{table: zero-shot results}
\begin{center}
\setlength{\tabcolsep}{1.5pt} % Default value: 6pt
\renewcommand{\arraystretch}{1.1} % Default value: 1
\begin{tabular}{cc|ccccccccccccc|l}
% \hline
% \multicolumn{2}{c}{\textbf{Eval.}} & \multicolumn{14}{c|}{\textbf{Zero-shot datasets}} & \multirow{1}{*}{\textbf{Average Zero-shot}} \\ \cline{3-15}
% \multicolumn{2}{c|}{\textbf{Eval.}}
\toprule
\multirow{6}{*}{Eval.}& \multirow{6}{*}{Method} & &&&&&&&&&&&&& \multirow{6}{*}{\makecell[c]{\text{Average} \\ \text{Zero-shot}}} \\
% & \multicolumn{13}{c}{\textbf{Datasets}} &  \multirow{7}{*}{\makecell[c]{\text{Average} \\ \text{Zero-shot}}}\\
% \hline
% \cline{3-15}
& & \rotatebox{90}{CalTech} & \rotatebox{90}{Cars} & \rotatebox{90}{CIFAR10} & \rotatebox{90}{CIFAR100} & \rotatebox{90}{DTD} & \rotatebox{90}{EuroSAT} & \rotatebox{90}{FGVC} & \rotatebox{90}{Flowers} & \rotatebox{90}{ImageNet-R} & \rotatebox{90}{ImageNet-S} & \rotatebox{90}{PCAM} & \rotatebox{90}{OxfordPets} & \rotatebox{90}{STL-10} &  \\ 
\hline
\multirow{5}{*}{\rotatebox{90}{clean}} & CLIP & 83.3 & 77.9 & 95.2 & 71.1 & 55.2 & 62.6 & 31.8 & 79.2 & 87.9 & 59.6 & 52.0 & 93.2 & 99.3 & 73.1 \\
% & TeCoA$^{2}$ & 80.7 & 50.1 & 87.5 & 60.7 & 44.4 & 26.1 & 14.0 & 51.8 & 80.1 & 58.4 & 49.9 & 80.0 & 96.1 & 60.0 \\
% & FARE$^{2}$  & 84.8 & 70.5 & 89.5 & 69.1 & 50.0 & 25.4 & 26.7 & 70.6 & 91.1 & 59.7 & 50.0 & 91.1 & 98.5 & 67.0 \\
& TeCoA & 78.4 & 37.9 & 79.6 & 50.3 & 38.0 & 22.5 & 11.8 & 38.4 & 74.3 & 54.2 & 50.0 & 76.1 & 93.4 & 54.2 \\
& FARE & 84.7 & 63.8 & 77.7 & 56.5 & 43.8 & 18.3 & 22.0 & 58.1 & 80.2 & 56.7 & 50.0 & 87.1 & 96.0 & 61.1 \\
% & \bf LMz(adv) & 82.7 & 78.6 & 95.7 & 72.8&55.7& 63.5& 33.2&79.4& 87.5 & 58.4 &51.8 & 92.5 & 99.5 & 73.2\\
% & \bf LMz(COUP) & &&95.7&&&&&&&&&&&\\
% & \bf LMx-Diff&  &&& &&& && & &&&\\
&\cellcolor{mygray}\bf CLIPure-Diff&\cellcolor{mygray}79.9&\cellcolor{mygray}75.5&\cellcolor{mygray}94.9&\cellcolor{mygray}63.8&\cellcolor{mygray}55.2&\cellcolor{mygray}58.2&\cellcolor{mygray}29.3&\cellcolor{mygray}75.0&\cellcolor{mygray}87.5&\cellcolor{mygray}55.9&\cellcolor{mygray}56.8&\cellcolor{mygray}90.4&\cellcolor{mygray}98.4&\cellcolor{mygray}70.8\\
&\cellcolor{mygray}\bf CLIPure-Cos&\cellcolor{mygray}82.9&\cellcolor{mygray}78.6&\cellcolor{mygray}95.6&\cellcolor{mygray}73.0&\cellcolor{mygray}55.4&\cellcolor{mygray}63.3&\cellcolor{mygray}33.2&\cellcolor{mygray}79.3&\cellcolor{mygray}87.7&\cellcolor{mygray}58.3&\cellcolor{mygray}52.0&\cellcolor{mygray}92.5&\cellcolor{mygray}99.6&\cellcolor{mygray}73.2\\
\hline
\multirow{5}{*}{\rotatebox{90}{$\ell_{\infty} = 2/255$}} & CLIP & 0.0 & 0.0 & 0.0 & 0.0 & 0.0 & 0.1 & 0.0 & 0.0 & 0.0 & 0.0 & 0.0 & 0.0 & 0.0 & 0.0 \\
% & TeCoA$^{2}$ & 70.2 & 22.2 & 63.7 & 35.0 & 27.0 & 12.8 & 5.8 & 27.6 & 58.8 & 45.2 & 40.0 & 69.7 & 88.7 & 43.6 \\
% & FARE$^{2}$ & 73.0 & 26.0 & 60.3 & 35.6 & 26.7 & 6.2 & 5.9 & 31.2 & 68.3 & 38.3 & 41.9 & 68.3 & 90.1 & 43.1 \\
& TeCoA & 69.7 & 17.9 & 59.7 & 33.7 & 26.5 & 8.0 & 5.0 & 24.1 & 59.2 & 43.0 & 48.8 & 68.0 & 86.7 & 42.3 \\
& FARE & 76.7 & 30.0 & 57.3 & 36.5 & 28.3 & 12.8 & 8.2 & 31.6 & 61.6 & 41.6 & 48.4 & 72.4 & 89.6 & \underline{45.9}  \\
% & \bf LMx-Diff&  & & &  & & &  &  & & & & &\\
& \cellcolor{mygray}\bf CLIPure-Diff &\cellcolor{mygray}75.1& \cellcolor{mygray}65.9&  \cellcolor{mygray}92.5&\cellcolor{mygray}52.6&\cellcolor{mygray}45.9&\cellcolor{mygray}41.5&\cellcolor{mygray}20.8 &\cellcolor{mygray}65.8&\cellcolor{mygray}86.5 &\cellcolor{mygray}49.8&\cellcolor{mygray}51.4&\cellcolor{mygray}86.2&\cellcolor{mygray}97.9&\cellcolor{mygray}64.0 \textcolor{blue}{$\uparrow$}\textcolor{blue}{39.4\%}\\
& \cellcolor{mygray}\bf CLIPure-Cos&\cellcolor{mygray} 80.8&\cellcolor{mygray}73.9&\cellcolor{mygray}93.0&\cellcolor{mygray}65.0&\cellcolor{mygray}50.7&\cellcolor{mygray}49.1&\cellcolor{mygray}28.2&\cellcolor{mygray}75.3&\cellcolor{mygray}85.4&\cellcolor{mygray}54.3&\cellcolor{mygray}49.1&\cellcolor{mygray}91.2&\cellcolor{mygray}99.5&\cellcolor{mygray}68.8 \textcolor{blue}{$\uparrow$}\textcolor{blue}{45.9\%}\\
\hline
\multirow{5}{*}{\rotatebox{90}{$\ell_{\infty} = 4/255$}} & CLIP & 0.0 & 0.0 & 0.0 & 0.0 & 0.0 & 0.0 & 0.0 & 0.0 & 0.0 & 0.0 & 0.0 & 0.0 & 0.0 & 0.0 \\
% & TeCoA$^{2}$ & 57.4 & 6.5 & 31.0 & 17.8 & 14.7 & 7.7 & 1.1 & 9.8 & 36.7 & 32.8 & 16.0 & 50.3 & 69.2 & 27.0 \\
% & FARE$^{2}$ & 46.6 & 4.8 & 25.9 & 13.9 & 11.7 & 0.5 & 0.6 & 7.1 & 22.5 & 22.5 & 17.2 & 27.9 & 61.7 & 20.5 \\
& TeCoA & 60.9 & 8.4 & 37.1 & 21.5 & 16.4 & 6.6 & 2.1 & 12.4 & 41.9 & 34.2 & 44.0 & 55.2 & 74.3 & 31.9 \\
& FARE & 64.1 & 12.7 & 34.6 & 20.2 & 17.3 & 11.1 & 2.6 & 12.5 & 40.6 & 30.9 & 48.3 & 50.7 & 74.4 & \underline{32.4} \\
% & \bf LMz-Diff(adv) & 78.6 & 65.9 & 94.9 & 62.3 & 48.8& 46.6&22.7&68.0& 81.9&48.9&51.4& 81.9 & 98.8 & 65.4 \\
% & \bf LMx-Diff&  & & &  & & &  &  & & & & &\\
&\cellcolor{mygray}\bf CLIPure-Diff&\cellcolor{mygray}74.1&\cellcolor{mygray}66.3&\cellcolor{mygray}90.6&\cellcolor{mygray}51.6&\cellcolor{mygray}45.2&\cellcolor{mygray}42.9&\cellcolor{mygray}\cellcolor{mygray}18.8&\cellcolor{mygray}65.8&\cellcolor{mygray}83.8&\cellcolor{mygray}48.8&\cellcolor{mygray}52.0&\cellcolor{mygray}85.7&\cellcolor{mygray}97.2&\cellcolor{mygray}63.3 \textcolor{blue}{$\uparrow$}\textcolor{blue}{95.4\%}\\
&\cellcolor{mygray}\bf CLIPure-Cos&\cellcolor{mygray}80.1&\cellcolor{mygray}72.2&\cellcolor{mygray}91.1&\cellcolor{mygray}59.1&\cellcolor{mygray}50.1&\cellcolor{mygray}48.4&\cellcolor{mygray}26.1&\cellcolor{mygray}74.8&\cellcolor{mygray}84.6&\cellcolor{mygray}52.4&\cellcolor{mygray}48.9&\cellcolor{mygray}91.0&\cellcolor{mygray}99.4&\cellcolor{mygray}67.4 \textcolor{blue}{$\uparrow$}\textcolor{blue}{108.0\%}\\
\bottomrule
\end{tabular}
\end{center}
\end{table}

\subsection{Case Study}
\label{section: case study}
\begin{figure}[t]
\centering
\includegraphics[width=0.9\linewidth]{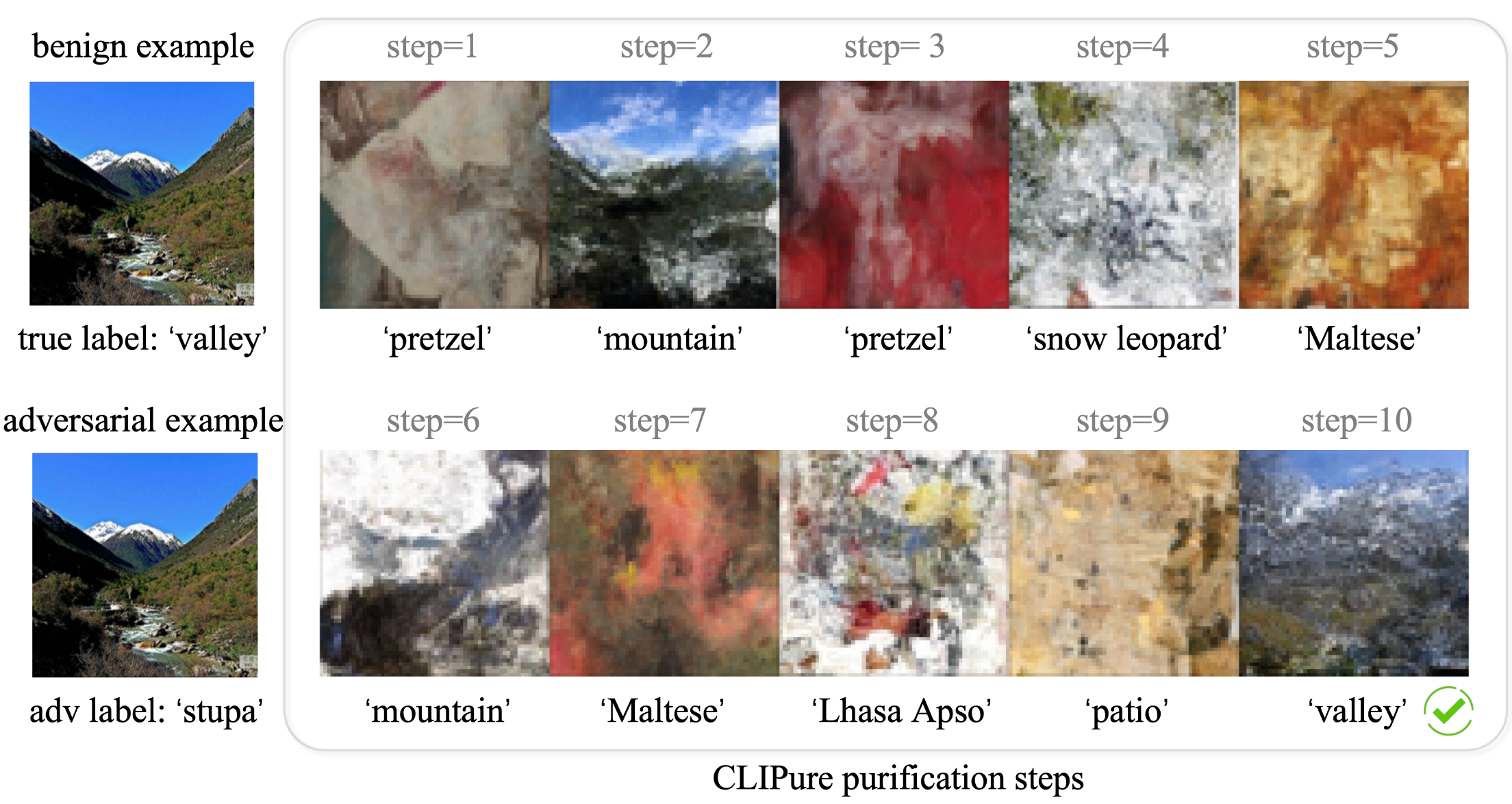} % Adjust the path and options as needed
\caption{Purification process of our CLIPure model for an image of a 'valley' adversarially perturbed to be classified as 'stupa'. Using the DaLLE2 Decoder \citep{ramesh2022hierarchical}, we visualize the stages of image embedding purification. The text below each picture annotates the classification results corresponding to each timestep in the purification process.}
\label{figure: purification process shown by generated images}
\end{figure}

To visualize the purification path of CLIPure, we employ the Decoder of DaLLE2 \citep{ramesh2022hierarchical}, which models the generation process from image embeddings to images through a diffusion model. As shown in Figure~\ref{figure: purification process shown by generated images}, starting from an adversarial example initially classified as a 'stupa,' CLIPure modifies the semantic properties of the image embedding. By the second step, the image is purified to 'mountain,' closely aligning with the semantics of the original image, which also contains mountainous features. Subsequently, after multiple purification steps, the image embedding is accurately classified as the 'valley' category.

\begin{figure}[t]
\centering
% \newlength{\commonheight}
% \setlength{\commonheight}{3.5cm}
\begin{subfigure}[b]{0.49\linewidth}
    \includegraphics[width=\linewidth]{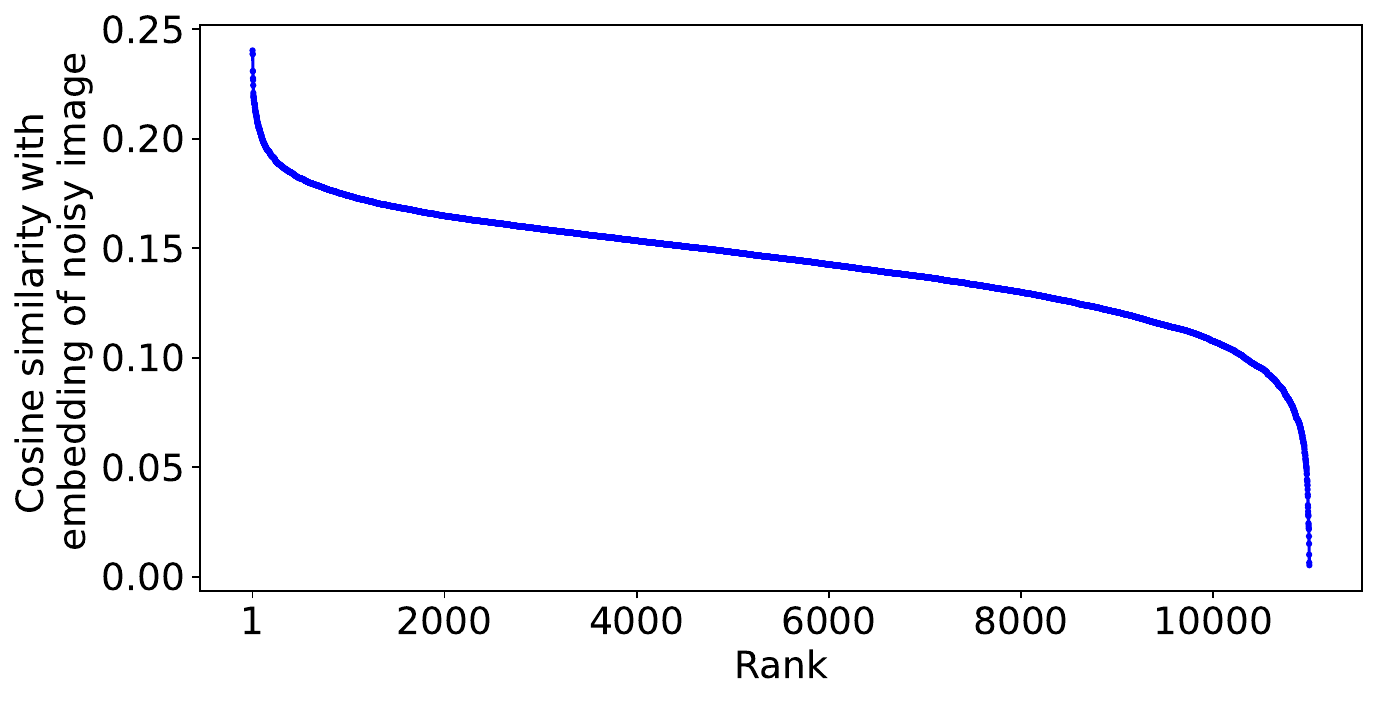}
    \caption{clean sample}
    \label{subfigure: vocab_distribution_clean}
\end{subfigure}
% \hfill  % Add space between subfigures
\begin{subfigure}[b]{0.49\linewidth}  % Width set to one third of text width
    \includegraphics[width=\linewidth]{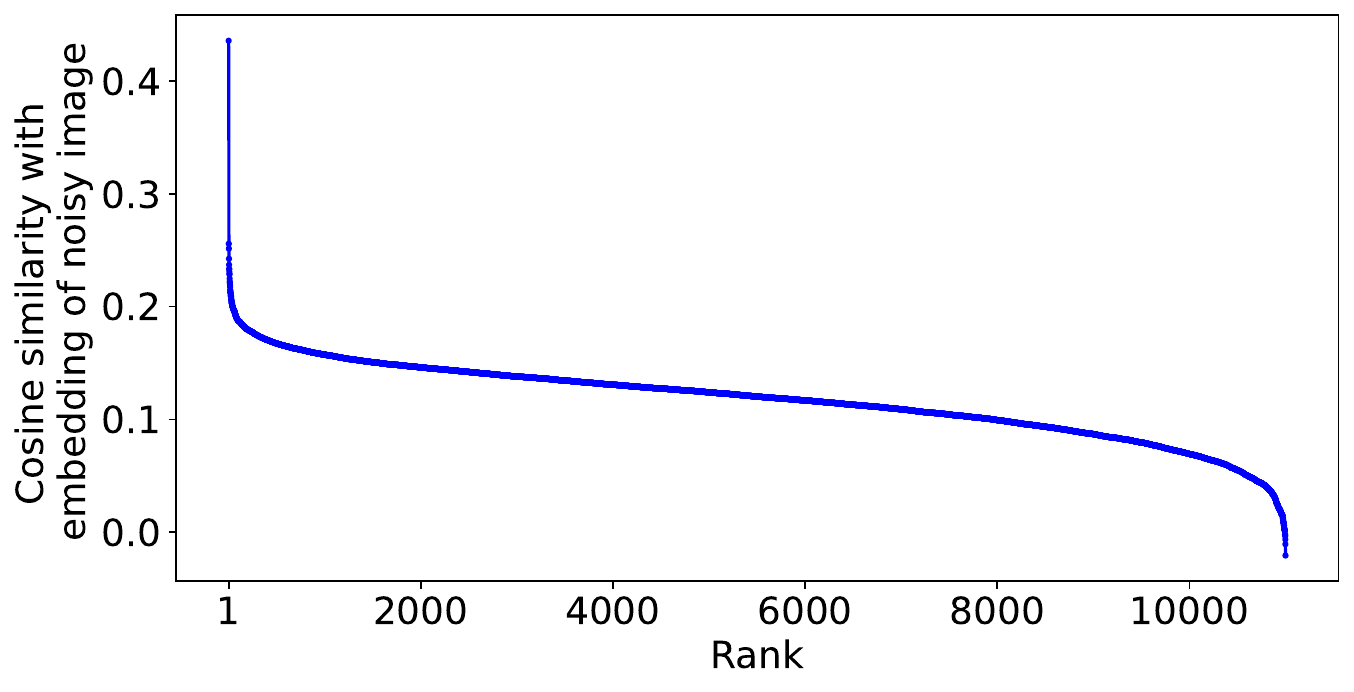}
    \caption{adversarial example}
    \label{subfigure: vocab_distribution_adv}
\end{subfigure}
\caption{Cosine similarity between words and the image embedding of (a) clean sample and (b) adversarial example of the case shown in Figure~\ref{figure: purification process shown by generated images} to illustrate the top-ranked words.}
\label{figure: cos with word distribution of clean and adv}
\end{figure}

In Figure~\ref{figure: cos with word distribution of clean and adv}, we analyze the distribution of cosine similarity between the embeddings of clean images and adversarial examples with words to understand the semantics of image embeddings. In Figure~\ref{figure: cosine similarity with words on clean and adv samples}, we also highlight the specific top-ranked categories for the benign and adversarial examples. We observe that the cosine similarity distribution for words close to the clean image is stable and resembles the prior distribution shown in Figure~\ref{subfigure: vocab_distribution_prior}. In contrast, the distribution of cosine similarity for the adversarial examples shows anomalies, with the top-ranked adversarial labels displaying abnormally high cosine similarities. This could potentially offer new insights into adversarial sample detection and defense strategies.

\begin{figure}[t]
% \captionsetup{labelfont={color=blue}}
\centering
\includegraphics[width=0.6\linewidth]{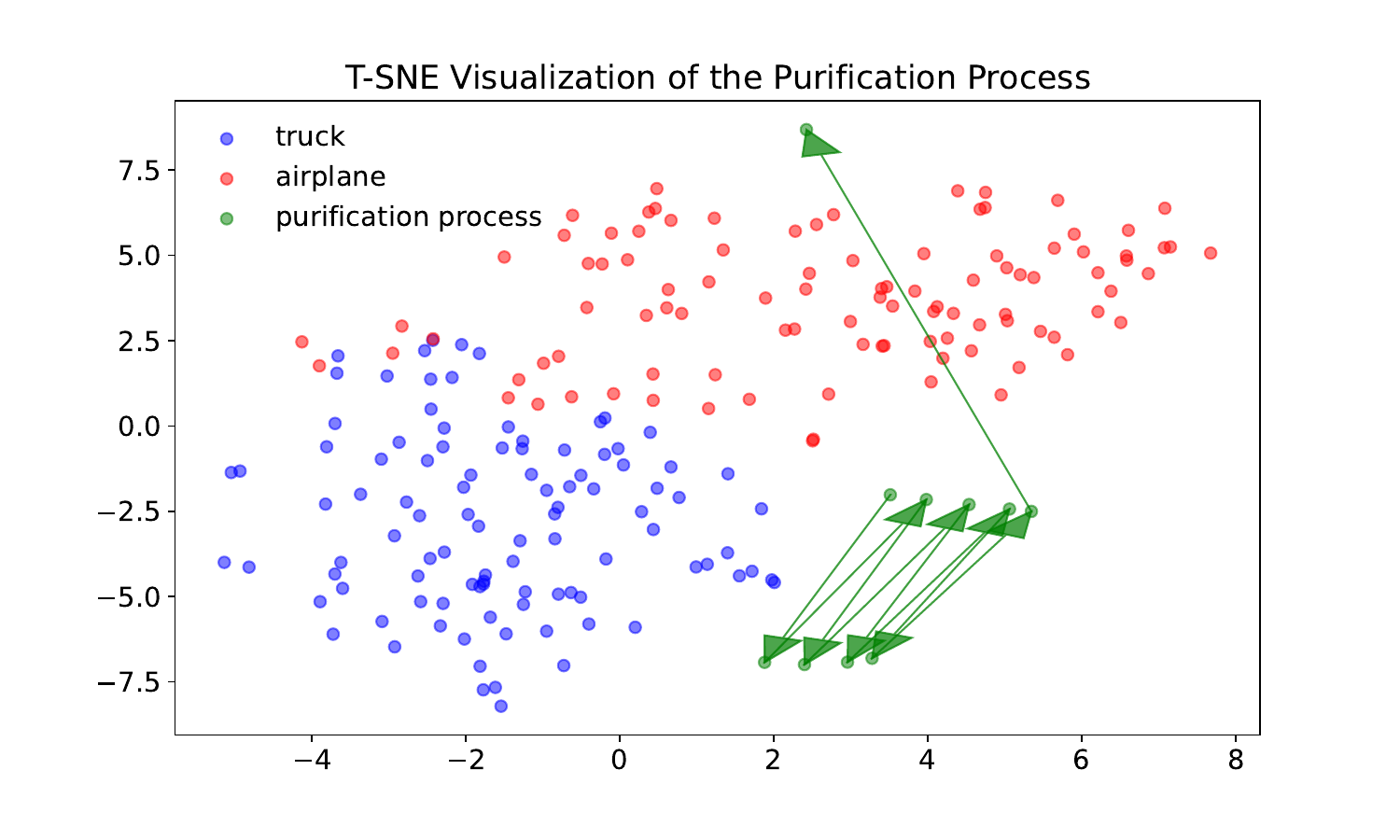} % Adjust the path and options as needed
\caption{Purification process of the adversarial example attacked from ground truth label 'airplane' to adversarial label 'truck' and purified by our CLIPure-Diff visualized by T-SNE. The image is randomly sampled from the CIFAR-10 dataset.}
\label{figure: purification process by T-SNE}
\end{figure}

Additionally, we illustrate the purification path using T-SNE. As depicted in Figure~\ref{figure: purification process by T-SNE}, an image sampled from CIFAR-10 originally categorized as 'airplane' was adversarially manipulated to resemble a 'truck,' making it an outlier. Through our CLIPure method, the image is purified towards a high-density area and ultimately reclassified accurately as the 'airplane' category.

% \textbf{Analysis on Semantically Similar Words.}
% \subsection{Analysis on Semantically Similar Words}

\subsection{Analysis from Textual Perspective}

\begin{figure}[t]
\centering
% \newlength{\commonheight}
% \setlength{\commonheight}{3.5cm}
\begin{subfigure}[b]{0.46\linewidth}
    \includegraphics[width=\linewidth]{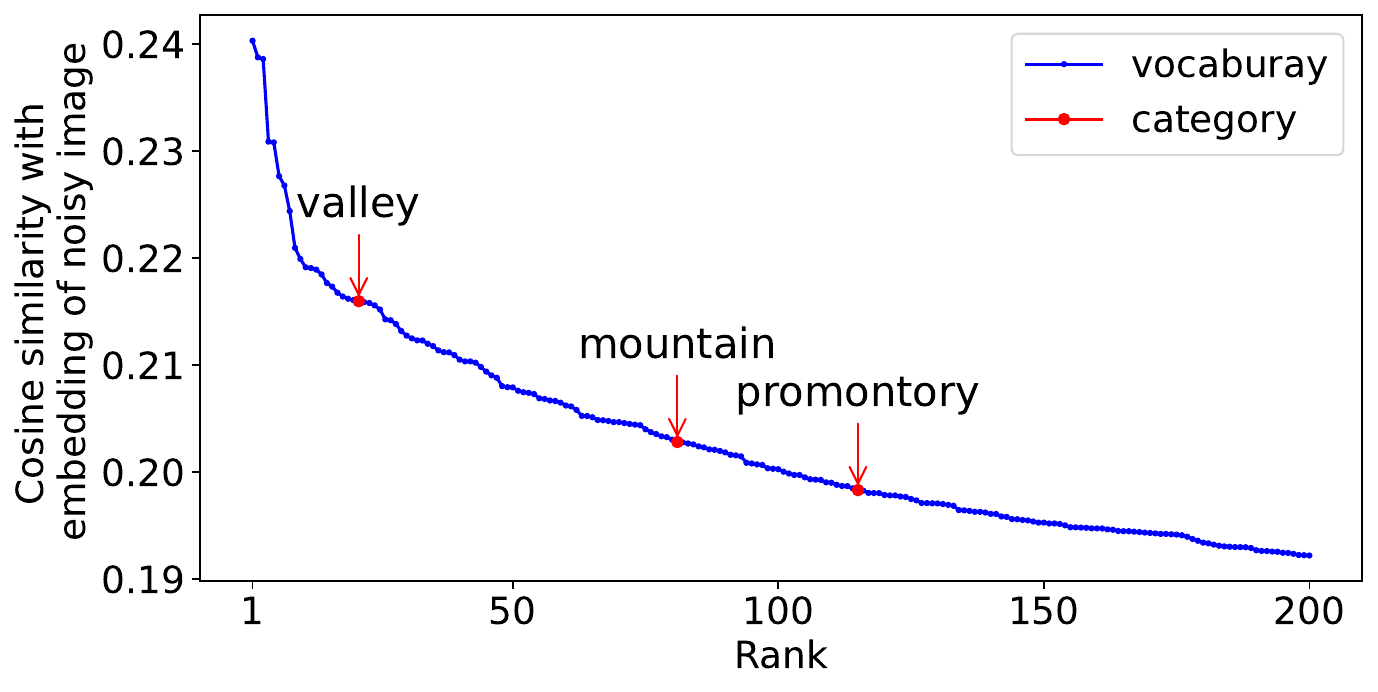}
    \caption{clean sample}
    \label{subfigure: vocab_topwords_clean}
\end{subfigure}
% \hfill  % Add space between subfigures
\begin{subfigure}[b]{0.46\linewidth}  % Width set to one third of text width
    \includegraphics[width=\linewidth]{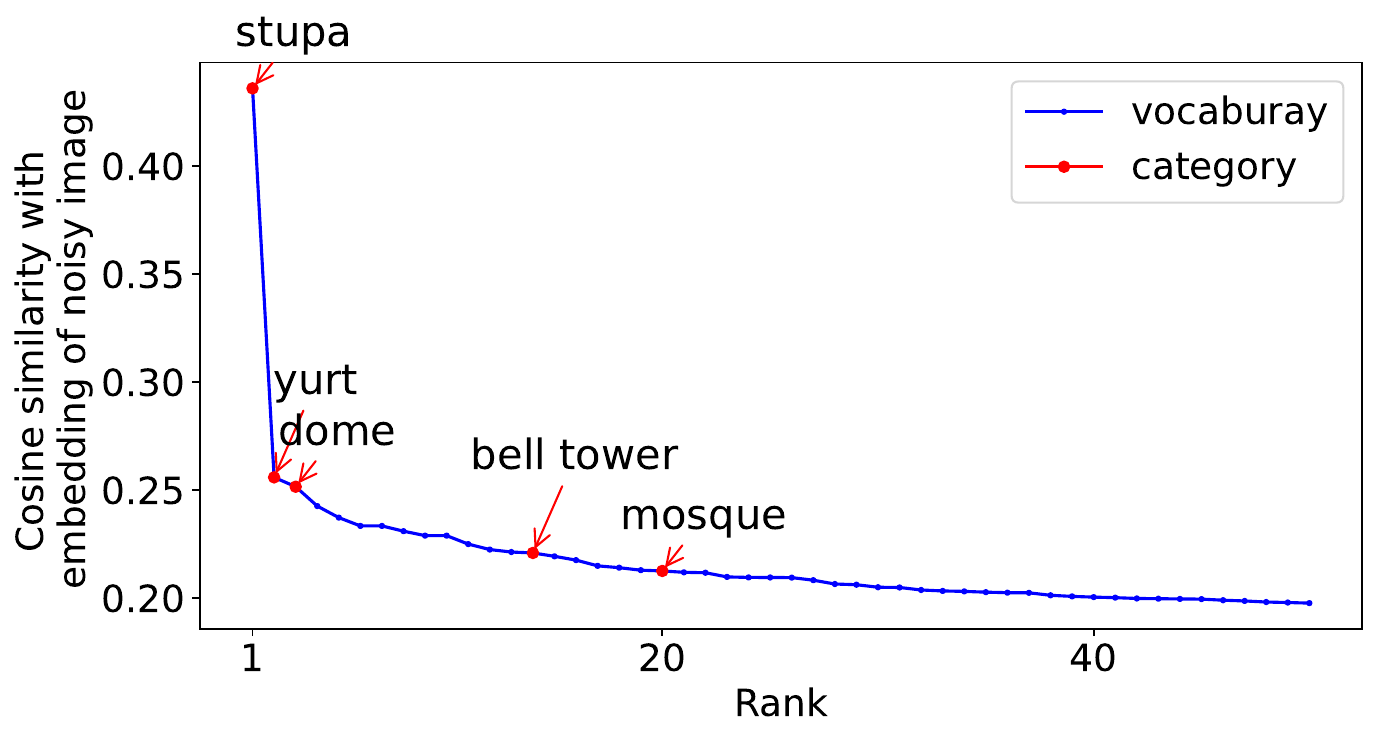}
    \caption{adversarial example}
    \label{subfigure: vocab_topwords_adv}
\end{subfigure}
\caption{Cosine similarity between word and image embeddings of (a) clean and (b) adversarial examples in Figure~\ref{figure: purification process shown by generated images} across different ranks. Blue dots denote 10,000 words sampled from Word2Vec vocabulary \citep{church2017word2vec}, and red dots denote words from the 1,000 ImageNet categories.}
\label{figure: cosine similarity with words on clean and adv samples}
\end{figure}

\begin{figure}[t]
\centering
% \newlength{\commonheight}
% \setlength{\commonheight}{3.5cm}
\begin{subfigure}[b]{0.49\linewidth}
    \includegraphics[width=\linewidth]{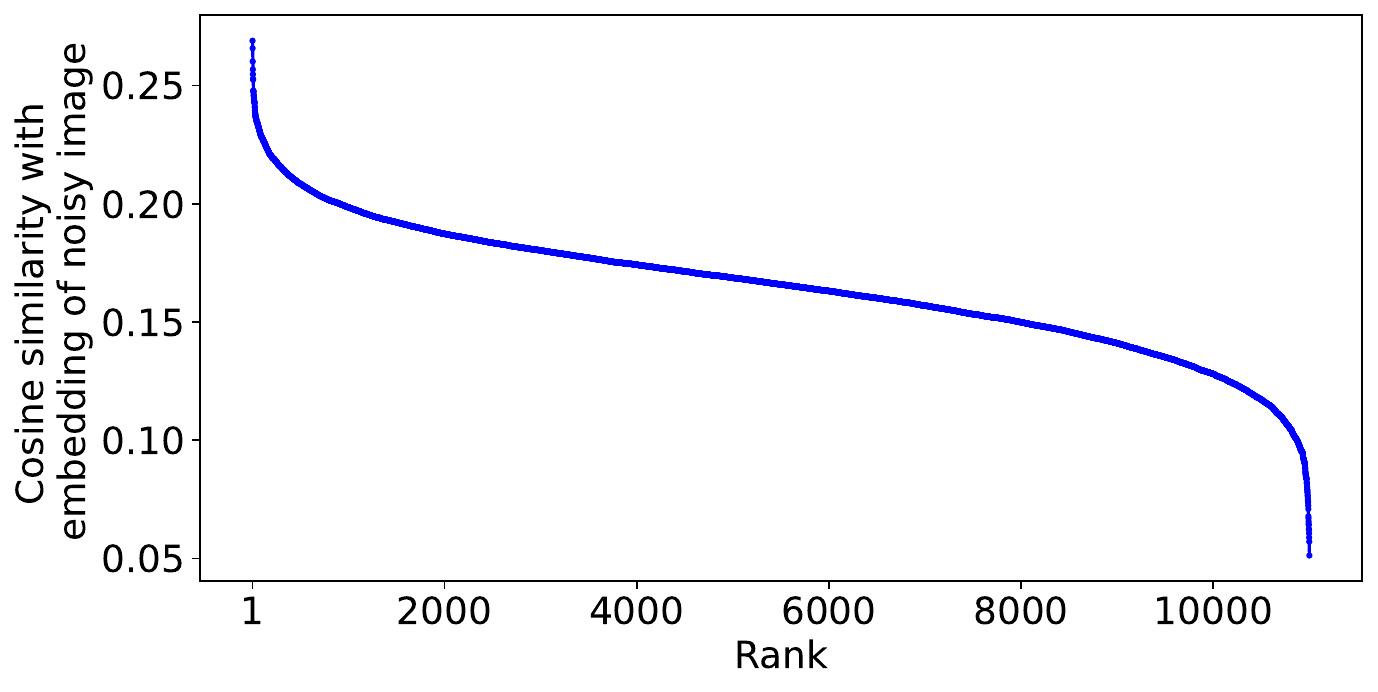}
    \caption{prior probability}
    \label{subfigure: vocab_distribution_prior}
\end{subfigure}
% \hfill  % Add space between subfigures
\begin{subfigure}[b]{0.49\linewidth}  % Width set to one third of text width
    \includegraphics[width=\linewidth]{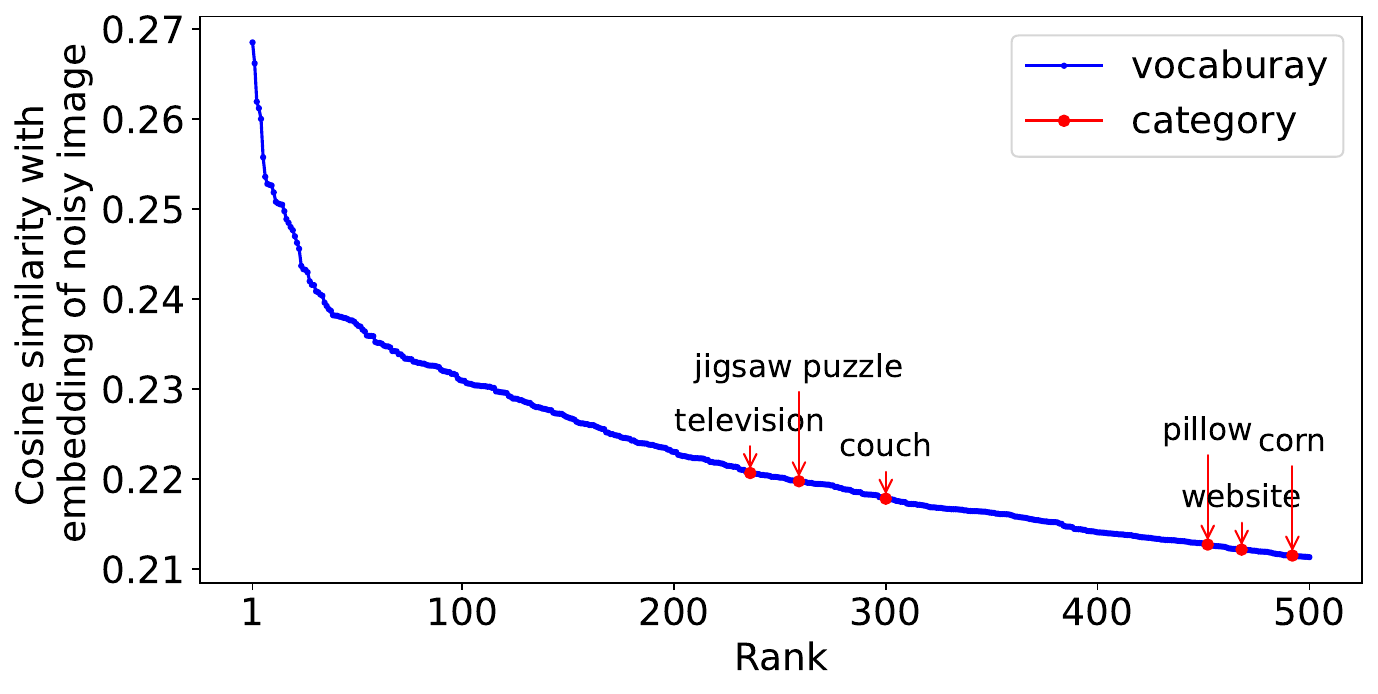}
    \caption{highlight top-ranked words}
    \label{subfigure: vocab_topwords_noise}
\end{subfigure}
\caption{(a)Cosine similarity between words and the image embeddings of a noisy example to illustrate the prior probability of the words. (b) Highlights the top-ranked words within the 1,000 ImageNet categories, emphasizing the most semantically similar words to the noisy image.}
\end{figure}

In this section, we provide additional analysis from a textual perspective as a supplement to the analysis. In Figure~\ref{figure: cosine similarity with words on clean and adv samples}, we take advantage of the text modality of the CLIP model to understand the semantics of clean and adversarial examples by matching them with closely related words. We selected words from ImageNet's categories and supplemented this with an additional 10,000 randomly drawn words from natural language vocabulary for a comprehensive list. We observe that the meanings of the category words ranking high are relatively similar. For clean samples, the distribution of cosine similarity across ranks is relatively stable, whereas the adversarial samples exhibit abnormally high cosine similarity for adversarial categories at top ranks. 
% Our vocabulary includes the 1,000 categories from ImageNet and 10,000 words randomly sampled from Word2Vec \citep{church2017word2vec}. 
Figure~\ref{subfigure: vocab_distribution_prior} displays the distribution of cosine similarities between the word embeddings and an image embedding generated from pure noise, across various ranks. We aim to use these similarities to represent the prior probabilities of the words. The results indicate that most samples cluster around a cosine similarity of approximately 0.15, with only a few extreme values either much higher or lower, suggesting a normal distribution pattern. This indicates that a majority of the samples have moderate prior probabilities, contrasting with the long-tail distribution often observed in textual data. In Figure~\ref{subfigure: vocab_topwords_noise}, we highlight the top-ranked words in the vocabulary that are closest to the noisy image embedding. Words like ``television'', which appear frequently in image data, rank high, aligning with our expectations.
This abnormal phenomenon in adversarial samples could potentially inspire adversarial detection and more robust adversarial defense methods.

\subsection{Combining CLIPure with Other Strategies}
\label{section: combination}
In this section, we conduct experiments that combine orthogonal strategies, including adversarial training \citep{schlarmann2024robust} and pixel space purification \citep{chen2023robust}, as well as incorporating classifier confidence guidance.
We carry out evaluations using a sample of 256 images from the ImageNet test set. The experimental setup in this section is aligned with the same configuration as presented in Table~\ref{table: main result on imagenet}, detailed in Section~\ref{section: experimental settings}.

\begin{figure}[t]
\centering
% \newlength{\commonheight}
% \setlength{\commonheight}{3.5cm}
\begin{subfigure}[b]{0.53\linewidth}  % Width set to one third of text width
    \includegraphics[width=\linewidth]{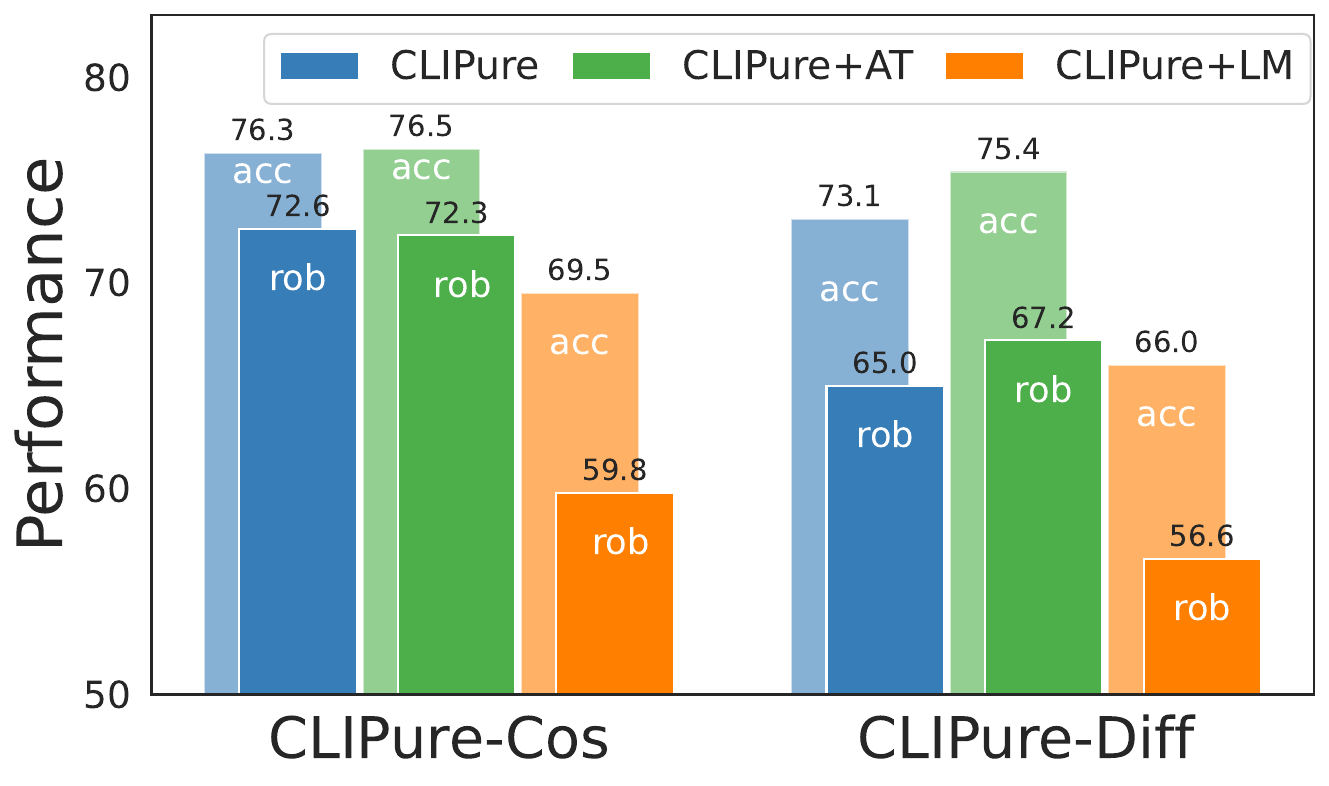}
    \caption{Combination methods}
    \label{subfigure: combination}
\end{subfigure}
% \hfill  % Add space between subfigures
\begin{subfigure}[b]{0.39\linewidth}
    \includegraphics[width=\linewidth]{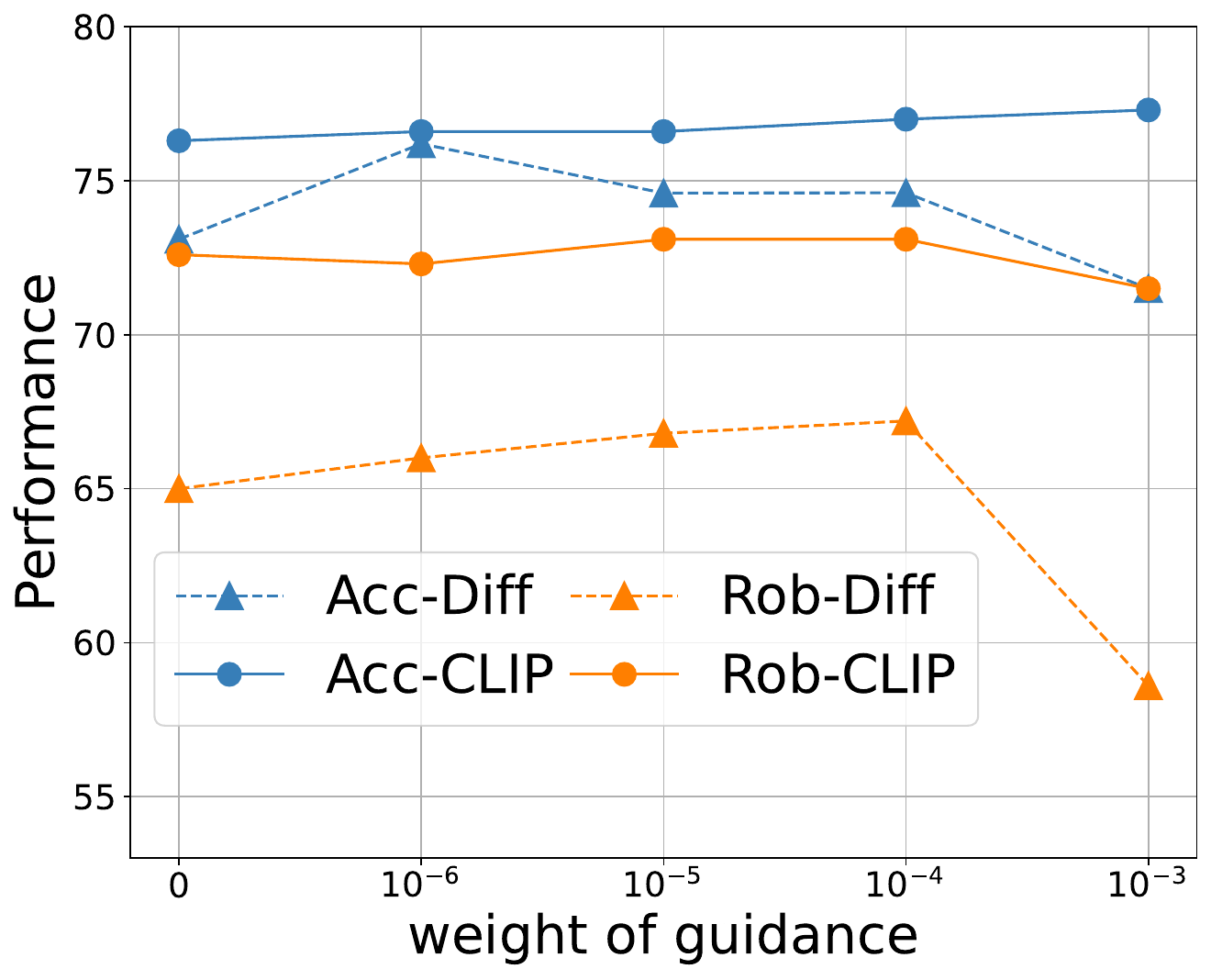}
    \caption{Impact of guidance weight}
    \label{subfigure: guidance}
\end{subfigure}
\caption{(a) Accuracy and robustness against AutoAttack with $\ell_{\infty}=4/255$ on ImageNet of CLIPure, CLIPure combined with Adversarial Training (AT), and CLIPure combined with pixel space Likelihood Maximization (LM). (b) Performance on ImageNet against AutoAttack with $\ell_{\infty}=4/255$ incorporating with classifier confidence guidance.}
\end{figure}

\textbf{Combination with Adversarial Training (CLIPure+AT).}
Orthogonal adversarial training (AT) methods like FARE \citep{schlarmann2024robust} has fintuned CLIP on ImageNet. We combine CLIPure and FARE by purifying image embedding in FARE's latent space and classification via FARE. The experimental results, depicted in Fig.~\ref{subfigure: combination}, indicate that the CLIPure-Diff+AT method enhances adversarial robustness in CLIPure-Diff. This improvement could be attributed to FARE providing more accurate likelihood estimates, as it is specifically trained on adversarial examples. When combined with CLIPure-Cos, the outcomes were comparable, likely because the inherent robustness of CLIPure-Cos is already high, leaving limited scope for further enhancement.

\textbf{Combination with Pixel Space Purification (CLIPure+LM).}
Purification in latent space and pixel space occurs at different stages of processing. Pixel space purification acts directly on the input samples, while latent space purification operates on the latent vectors encoded from the input picture. By combining both methods, we first purify the input samples in pixel space, then pass the purified samples through an encoder to obtain latent vectors, which are further purified in latent space. We mark this combination as CLIPure+LM. For the Likelihood Maximization (LM) method, we chose LM-EDM \citep{chen2023robust} due to its strong performance across pixel space purification methods.

Experimental results, as shown in Fig.~\ref{subfigure: combination}, indicate that combining with LM leads to a decrease in both clean accuracy and robustness. This decline may be associated with the challenges of information loss inherent to pixel space purification. Since LM operates directly on the pixels of the image, it is highly sensitive to the degree of purification. However, the adversarial perturbation varies across samples, potentially leading to over-purification thus a decrease in performance. In contrast, purification in latent space with polar coordinates simply involves changing vector directions—altering semantics without reducing semantic information. Thus, it avoids the problem of semantic information loss.

\textbf{Combination with Classifier Guidance}.
In Eq.~\ref{equation: reverse sde of untargeted attack}, we note that the reverse-time SDE associated with the attack method includes not only a likelihood maximization purification term but also a classifier guidance term, represented as $\nabla_{\boldsymbol{x}} \mathcal{L}(\theta; \boldsymbol{x}_t, y_{\text{true}})$. This approach aligns with the classifier confidence guidance proposed by \citet{zhang2024classifier}. We incorporate this classifier guidance term in two versions of CLIPure (CLIPure-Diff and CLIPure-Cos) as described in Algorithm~\ref{algorithm: latent purification by CLIP and DaLLE2.DiffusionPrior}. Since the ground truth label $y_{\text{true}}$ is unknown, we employ the predicted label by the CLIP model as a proxy for $y_{\text{true}}$. To address potential inaccuracies in early estimations, classifier guidance was only applied in the final 5 steps of the 10-step purification process.

We evaluate the performance varied with the weight of the guidance term, as depicted in Fig.~\ref{subfigure: guidance}. It shows that a guidance weight of $10^{-4}$ led to the largest improvement in robustness: CLIPure-Diff achieved a robustness of 67.2\% (an increase of +2.2\% over scenarios without guidance), and CLIPure-Cos attained a robustness of 73.1\% (an improvement of +0.5\% over no guidance).

\subsection{T-SNE visualization}
In Figure~\ref{subfigure: polar plot of embeddings}, we display the distribution of image embeddings $\boldsymbol{z}^i$ for both benign samples and adversarial examples from a subset of 30 samples on the CIFAR-10 dataset, after dimensionality reduction using t-SNE. The figure also includes embeddings for 10 category-associated texts (e.g., ``a photo of a dog.'') $\boldsymbol{z}^t$, as well as the text embedding for a blank template ``a photo of a .''.

We observe distinct clustering of text vectors, clean image embeddings, and adversarial image embeddings in the space. The blank template, which is used for purification, is positioned at the center of the category text clusters, representing a general textual representation. This arrangement demonstrates that clean and adversarial sample distributions occupy distinct regions in the embedding space, providing a foundational basis for the effectiveness of adversarial sample purification.

\subsection{Hyperparameters}

\subsubsection{Impact of Step Size}

\begin{figure}[t]
\centering
% \newlength{\commonheight}
% \setlength{\commonheight}{3.5cm}
\begin{subfigure}[b]{0.4\linewidth}
    \includegraphics[width=\linewidth]{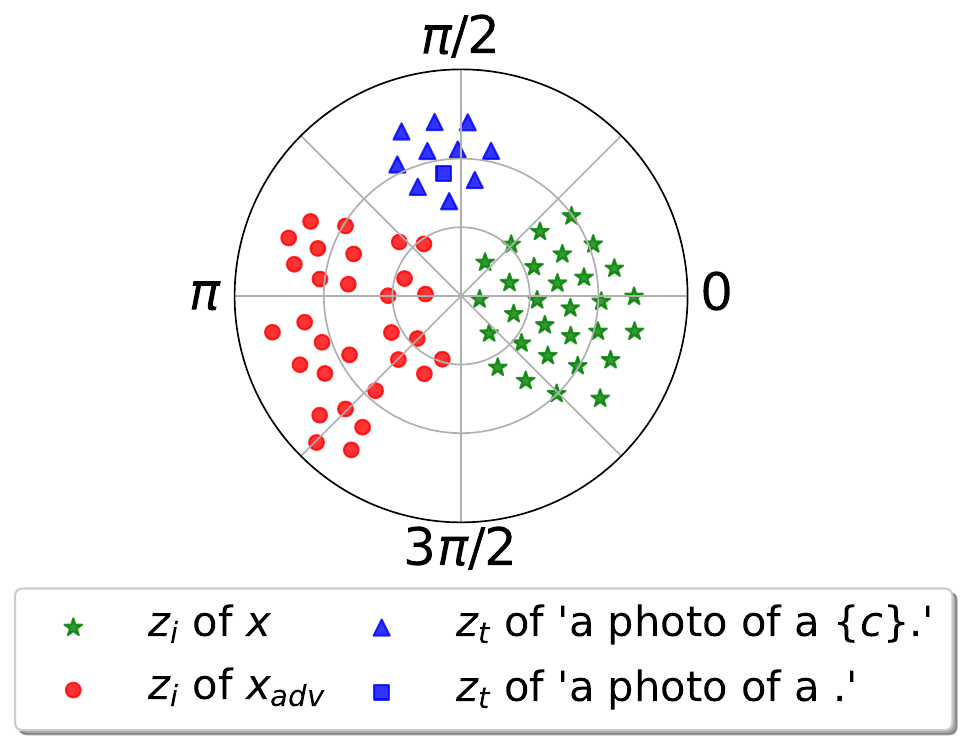}
    \caption{Embeddings in CLIP's latent space}
    \label{subfigure: polar plot of embeddings}
\end{subfigure}
\hfill  % Add space between subfigures
\begin{subfigure}[b]{0.54\linewidth}  % Width set to one third of text width
    \includegraphics[width=\linewidth]{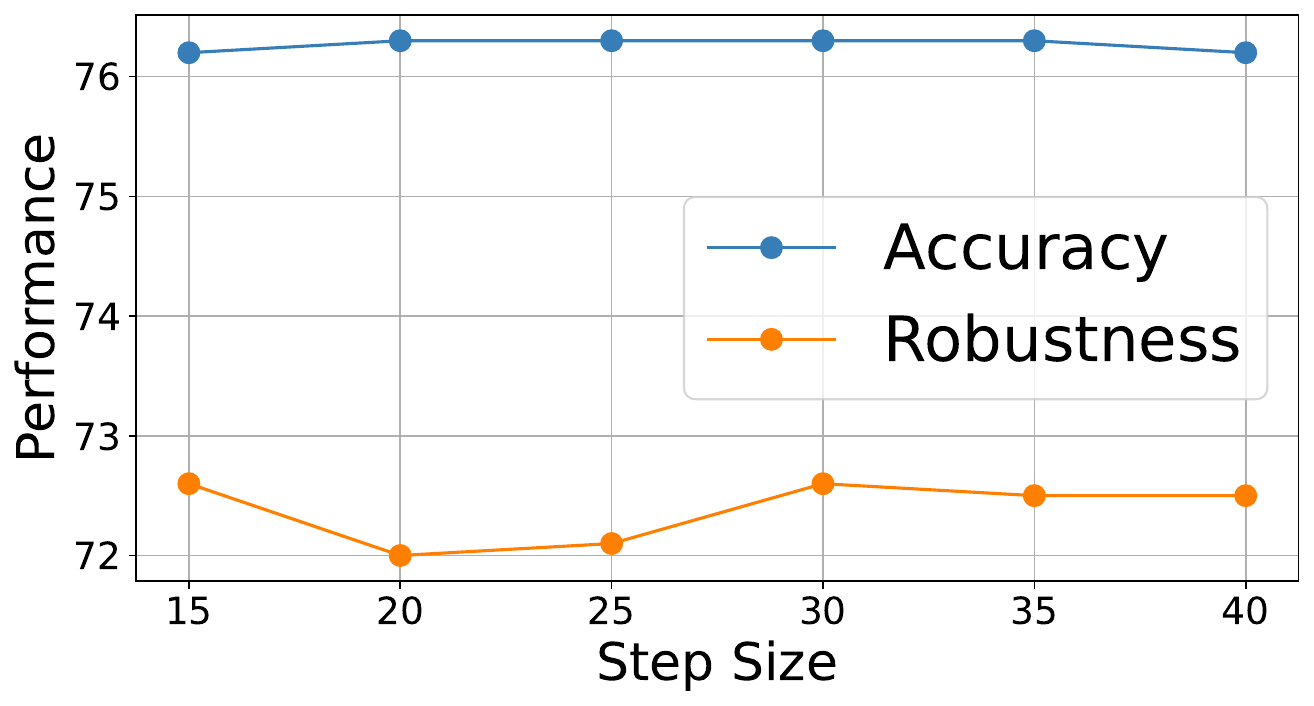}
    \caption{Impact of step size}
    \label{subfigure: hyperparam}
\end{subfigure}
\caption{(a) Visualization of image and text embeddings reduced via PCA and plotted in polar coordinates using t-SNE. $\boldsymbol{z}^i$ and $\boldsymbol{z}^t$ represent the image and text embeddings obtained through CLIP, respectively. Since vector direction signifies semantics, we depict the samples in polar coordinates to emphasize directional properties. (b) Impact of step size during a 10-step purification process using CLIPure-Cos on 1000 ImageNet samples against AutoAttack with $\ell_{\infty}$ threat model ($\epsilon=4/255$).}
\end{figure}

% \begin{figure}[t]
% \centering
% \includegraphics[width=0.75\linewidth]{figs/hyperparam_stepsize.pdf} % Adjust the path and options as needed
% \caption{hyperparameter.}
% \label{figure: hyperparameter stepsize}
% \end{figure}

\begin{figure}[t]
\centering
% \captionsetup{labelfont={color=blue}}
% \newlength{\commonheight}
% \setlength{\commonheight}{3.5cm}
\begin{subfigure}[b]{0.49\linewidth}
    \includegraphics[width=\linewidth]{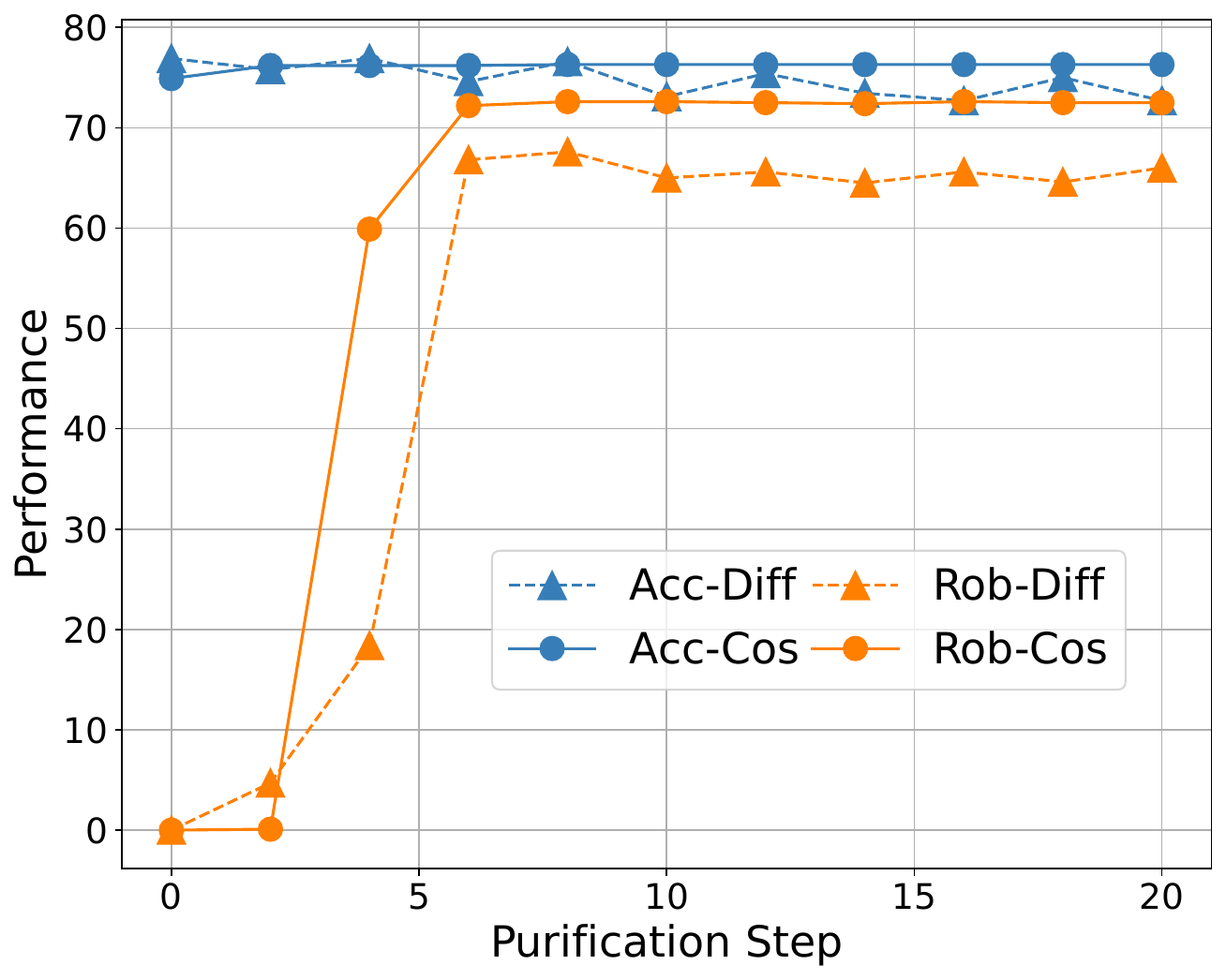}
    % \caption{\color{blue}{prior probability}}
    \label{subfigure: CLIPure_step}
\end{subfigure}
% \hfill  % Add space between subfigures
\begin{subfigure}[b]{0.49\linewidth}  % Width set to one third of text width
    \includegraphics[width=\linewidth]{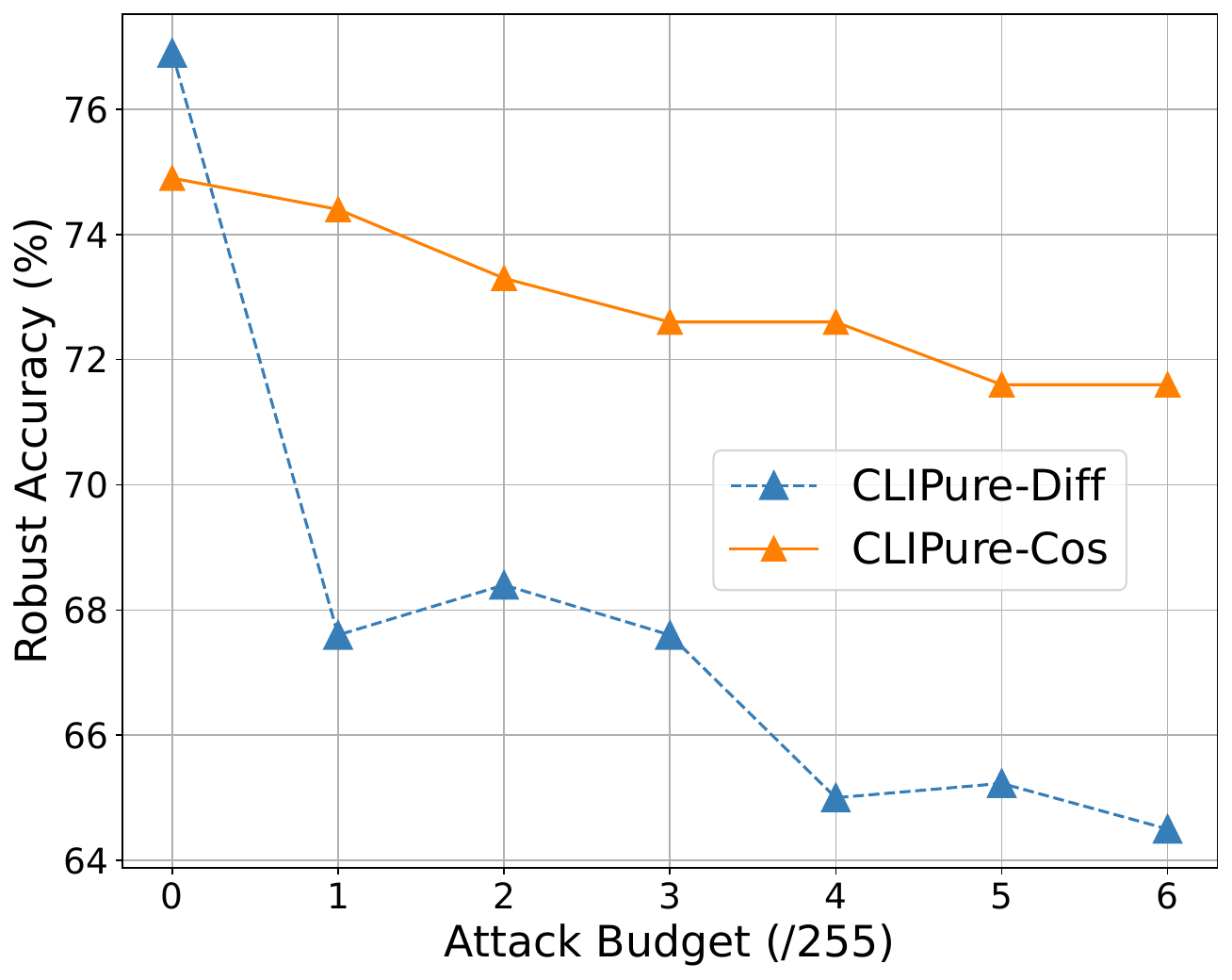}
    % \caption{\color{blue}{highlight top-ranked words}}
    \label{subfigure: CLIPure_eps}
\end{subfigure}
\caption{Clean accuracy (marked as "Acc") and robust accuracy (marked as "Rob") across different (Left) purification steps and (Right) attack budgets. "Diff" indicates CLIPure-Diff while "Cos" represents CLIPure-Cos.}
\label{figure: purification step and epsilon}
\end{figure}

We conducted experiments to assess the impact of different step sizes on the effectiveness of purification. Figure~\ref{subfigure: hyperparam} shows how the clean accuracy and robustness against AutoAttack ($\ell_\infty=4/255$) of our CLIPure-Cos method vary with changes in step size. These experiments were carried out on 1,000 samples from the ImageNet test set with 10-step purification. The results indicate that the performance of the method is relatively stable across different step sizes, consistently demonstrating strong effectiveness.

Limited to the computational complexity of CLIPure-Diff, we opt not to conduct a parameter search for this model. Instead, we directly apply the optimal parameters found for CLIPure-Cos.

\subsubsection{Impact of Purification Step}

As illustrated in Figure~\ref{figure: purification step and epsilon} (Left), we assess the performance of various purification steps in CLIPure-Cos and CLIPure-Diff. The results indicate that insufficient purification while maintaining acceptable clean accuracy, results in low robustness. As the number of purification steps increases, robustness gradually improves. Continuing to increase the purification steps stabilizes both accuracy and robustness, demonstrating a balance between the two metrics as the process evolves.

\subsubsection{Performance Across Attack Budget}
To further explore the adversarial defense capabilities of CLIPure, we evaluate the robustness of CLIPure-Diff and CLIPure-Cos against various attack budgets, following the settings used in Table~\ref{table: main result on imagenet}. As shown in Figure~\ref{figure: purification step and epsilon} (Right), CLIPure-Diff demonstrates superior clean accuracy (at $\epsilon$=0); however, its robustness decreases as the intensity of the attacks increases. In contrast, CLIPure-Cos exhibits a stronger ability to withstand adversarial perturbations.

\clearpage
\section{Supplementary Discussion on False Sense of CLIPure's Robustness Due to Implementation}
\label{appendix: supplementary for attack evaluation}

In this section, we provide a thorough analysis of an implementation oversight regarding computational graph retention during gradient-based white-box attacks on CLIPure. We demonstrate that omitting the \texttt{create\_graph=True} configuration led to artificial gradient obfuscation, masking the actual adaptive vulnerability of our initial defense. 

To rectify this, we clarify the theory of adaptive attacks and re-evaluate our framework using a gradient-free black-box attack. Our corrected evaluation reveals a stark divergence between the two variants: while the discriminative variant (\textit{CLIPure-Cos}) fails to provide effective defense against adversarial perturbations, the generative variant (\textit{CLIPure-Diff}) achieves successful latent purification, outperforming state-of-the-art baselines by a significant margin.

Finally, we provide a rigorous geometric and algebraic autopsy of why the discriminative variant (CLIPure-Cos) fails under true adaptive settings while preserving high clean accuracy through a unique perpendicular vector oscillation.

\subsection{Adaptive Attacks and Gradient Tracking Mechanics}
According to \citet{athalye2018obfuscated}, an \textbf{adaptive attack} is one that is constructed after a defense has been fully specified, where the adversary leverages complete knowledge of the defense mechanism and is restricted only by the threat model. This is further echoed by \citet{nie2022diffusion}, who define it as a scenario where the attacker possesses full knowledge of the defense method. 

In the context of purification-based defenses like CLIPure, a true white-box adaptive attack requires the attacker to backpropagate gradients through the entire defense pipeline. Since CLIPure introduces an optimization-based purification process prior to the CLIP classifier, the adversary must compute gradients of the final classification loss with respect to the input pixels $\vx$ by differentiating through the iterative purification steps.

% \subsection{Gradient Propagation through the Purification Bottleneck}
The core of CLIPure's purification involves updating the image embedding $\vz^i$ via gradient ascent to maximize $\log p(\vz^i)$ (reference to Algorithm~\ref{algorithm: latent purification by CLIP and DaLLE2.DiffusionPrior}). Let $\vz^i_0$ be the initial embedding and $\vz^i_N$ be the purified embedding after $N$ steps. The classification decision is made based on $\vz^i_N$. A gradient-based adaptive attacker (e.g., using PGD) seeks to compute:
\begin{equation}
    \nabla_{\vx} \mathcal{L}(\vz^i_N, y) = \frac{\partial \mathcal{L}}{\partial \vz^i_N} \cdot \frac{\partial \vz^i_N}{\partial \vz^i_{N-1}} \dots \frac{\partial \vz^i_1}{\partial \vz^i_0} \cdot \frac{\partial \vz^i_0}{\partial \vx}.
\end{equation}
This requires the computation of higher-order derivatives, as each step of purification already involves a gradient $\nabla_{\vz} \log p(\vz)$. 

In our implementation using the \texttt{torch.autograd.grad} API, two parameters are critical:
\begin{itemize}
    \item \textbf{\texttt{retain\_graph=True}}: Ensures the buffers used to compute the graph are not freed, allowing multiple passes.
    \item \textbf{\texttt{create\_graph=True}}: Instructs the engine to construct a graph of the derivative, enabling the computation of higher-order gradients.
\end{itemize}

During our initial experiments, we set \texttt{retain\_graph=True} to allow the attack to proceed, but \hl{erroneously relied on the default \texttt{create\_graph=False}. This led to a failure in constructing the second-order computational graph required to propagate gradients through the iterative purification updates. Consequently, the attacker was unable to obtain the full gradient information of the purified output relative to the input, resulting in a form of unintended gradient obfuscation.}

% \subsection{Discussion of Implementation of Adaptive Attacks of SOTA Purification Baselines}
Existing methods like \textit{DiffPure} and \textit{RDC} avoid this pitfall through specific architectural designs:
\begin{itemize}
    \item \textbf{DiffPure} \citep{nie2022diffusion}: Utilizes the score function $\nabla_{\vx} \log p(\vx)$ directly during the SDE-based denoising process. This formulation bypasses the need for explicit second-order differentiation through an optimization loop, as the purification process itself is defined by the first-order score.
    \item \textbf{RDC} \citep{chen2023robust}: Employs a custom \texttt{torch.autograd.Function} to implement a Backward Pass Differentiable Approximation (BPDA). By defining a custom backward pass:
    \begin{verbatim}
    @staticmethod
    def backward(ctx, grad):
        return grad
    \end{verbatim}
    RDC ensures that the gradient from the classifier is passed directly through the purification module (effectively treating the local Jacobian as an identity matrix), preventing gradient vanishing or obfuscation without the heavy overhead of $N$-step graph creation.
\end{itemize}

% \subsection{Analysis of Experimental Observations}
The omission of \texttt{create\_graph=True} explains some anomalies in our previous results. Specifically, in Table \ref{table: zero-shot results}, the CLIPure-Diff variant on the \textit{StanfordCars} dataset appeared to exhibit higher robustness at $\ell_\infty = 4/255$ than at $\ell_\infty = 2/255$. We initially attributed this to the stochasticity of the diffusion prior sampling. However, it is now clear that the primary cause was the disconnect in the computational graph, which rendered gradient-based attackers ineffective. 

Furthermore, our evaluation focused exclusively on white-box attacks. Had black-box attacks (such as \textit{Square Attack}) been included, the discrepancy where black-box methods potentially outperform white-box methods would have served as a diagnostic indicator of \textit{obfuscated gradients} \citep{athalye2018obfuscated}. The lack of a second-order graph effectively masked the true vulnerability of the purification loop to adaptive adversarial iterations.

\subsection{Robustness Evaluation via Square Attack and BPDA+EOT}
\label{sec: black_box_evaluation}

To circumvent the computational graph limitations imposed by the \texttt{create\_graph=False} setting in gradient-based white-box attacks, we conduct a rigorous and fair evaluation using the \textbf{Square Attack} \citep{andriushchenko2020square}, a query-efficient score-based black-box adversarial attack that operates independently of any model gradients or computational graph structures. We evaluate the adversarial robustness of CLIPure under an $\ell_\infty$ threat model with an attack budget of $\epsilon = 8/255$. 

Our evaluation spans 13 zero-shot benchmark datasets (including CIFAR-10, CIFAR-100, CalTech101, etc.) as well as the ImageNet dataset. To ensure a fair comparison against existing CLIP-based adversarially trained zero-shot classifiers, namely TeCoA \citep{mao2022understanding} and FARE \citep{schlarmann2024robust}, we follow the sampling protocol of \citet{schlarmann2024robust} and randomly sample 512 instances from each validation/test set for evaluation. Crucially, all evaluations for CLIPure are performed in a strict zero-shot manner utilizing the off-the-shelf pre-trained CLIP model without any dataset-specific fine-tuning or adversarial training.

Initially, we benchmarked both proposed variants of our framework. However, preliminary trials revealed that \textbf{CLIPure-Cos} exhibits a critical vulnerability under the black-box setting, achieving a robust accuracy of only $3.32\%$ on the CIFAR-10 dataset against the Square Attack ($\epsilon=8/255$). Consequently, we suspended further evaluation on the Cosine-similarity-based variant across the remaining datasets. A comprehensive theoretical and empirical autopsy detailing why CLIPure-Cos fails under true gradient-free black-box perturbations is elaborated in Section \ref{sec: reason Cos not work}. 

In light of this, we channel our primary evaluation into the generative variant, \textbf{CLIPure-Diff} (hereafter denoted simply as CLIPure in the black-box results). The comprehensive zero-shot performance across the 13 benchmark datasets is reported in Table \ref{table: zero-shot results}. 

The empirical results demonstrate a substantial and definitive breakthrough. While vanilla CLIP and its adversarially trained counterparts (FARE and TeCoA) completely collapse under the aggressive $\ell_\infty = 8/255$ Square Attack—yielding near-zero average robust accuracies ($3.61\%$, $3.62\%$, and $3.74\%$, respectively)—our generative purification framework retains remarkable resilience. CLIPure-Cos has a comparable robust accuracy of $3.56\%$. However, CLIPure-Diff achieves an exceptional average zero-shot robust accuracy of \textbf{59.89\%} across all 13 datasets, outperforming the best baseline (TeCoA) by an absolute margin of \textbf{56.15\%}. This represents a relative improvement of over $15x$ in robustness, thereby underscoring the formidable capability of multi-modal latent space purification when facing verified, gradient-obfuscation-free adversarial threat models.
However, when we re-evaluated CLIPure-Diff under the BPDA (Backward Pass Differentiable Approximation) setting, its robust accuracy plummeted to a mere 0.1\%, revealing that both variants of CLIPure are fundamentally vulnerable to BPDA-based and white-box attacks—although CLIPure-Diff remains effective against black-box threats.

\begin{table}[!htbp]
    \centering
    \caption{Zero-shot performance on 13 datasets against Square Attack with $\ell_\infty = 8/255$. Underlined results indicate the best robustness among baselines, while bold text denotes state-of-the-art (SOTA) performance across all methods.}
    \label{table: zero-shot results}
    \footnotesize
    \setlength{\tabcolsep}{1.5pt}
    \renewcommand{\arraystretch}{1.1}
    \begin{tabular}{cccccccccccccccc}
        \toprule
        \multirow{6}{*}{Evaluation} & \multirow{6}{*}{Method} & \multicolumn{13}{c}{Zero-Shot Datasets} & \multirow{6}{*}{\makecell[c]{Average}} \\
        \cmidrule{3-15}
        & & \rotatebox{90}{CalTech} & \rotatebox{90}{Cars} & \rotatebox{90}{CIFAR10} & \rotatebox{90}{CIFAR100} & \rotatebox{90}{DTD} & \rotatebox{90}{EuroSAT} & \rotatebox{90}{FGVC} & \rotatebox{90}{Flowers} & \rotatebox{90}{ImageNet-R} & \rotatebox{90}{ImageNet-S} & \rotatebox{90}{PCAM}  & \rotatebox{90}{OxfordPets} & \rotatebox{90}{STL-10} & \\
        \midrule
        \multirow{5}{*}{\rotatebox{90}{Clean}} & CLIP & 84.57 & 80.27 & 95.70 & 71.88 & 54.49 &65.43 & 32.81 & 79.69 & 88.09 & 58.40 & 52.73 & 91.99 & 99.22 &73.48 \\
        & FARE & 84.57 & 80.08 & 95.51 & 71.68  &  54.49& 65.43 & 32.81 & 79.88 & 88.09 & 58.20 & 52.93 & 91.99 & 99.22  &73.45 \\
        & TeCoA & 84.57 & 80.08 & 95.51 & 71.68 &  54.49& 65.43 & 32.81 & 79.88 & 88.09 & 58.20 & 52.93 & 91.99 & 99.22  & 73.45 \\
        & \cellcolor{gray!20}\textbf{CLIPure-Cos} & \cellcolor{gray!20}84.57 & \cellcolor{gray!20}80.08 & \cellcolor{gray!20}95.51 & \cellcolor{gray!20}71.68 & \cellcolor{gray!20}54.49 & \cellcolor{gray!20}65.43 & \cellcolor{gray!20}32.81 & \cellcolor{gray!20}79.88 & \cellcolor{gray!20}88.09 & \cellcolor{gray!20}58.20 & \cellcolor{gray!20}52.93 &
        \cellcolor{gray!20}91.99 &
        \cellcolor{gray!20}99.22 &
        \cellcolor{gray!20}73.45 \\
        & \cellcolor{gray!20}\textbf{CLIPure-Diff} & \cellcolor{gray!20}83.20 & \cellcolor{gray!20}77.15 & \cellcolor{gray!20}93.95 & \cellcolor{gray!20}63.09 & \cellcolor{gray!20}53.91 & \cellcolor{gray!20}59.57 & \cellcolor{gray!20}31.64 & \cellcolor{gray!20}73.44 & \cellcolor{gray!20}87.50 & \cellcolor{gray!20}58.01 & \cellcolor{gray!20}55.27 &
        \cellcolor{gray!20}89.84 &
        \cellcolor{gray!20}98.24 &
        \cellcolor{gray!20}71.14 \\
        \midrule
        \multirow{5}{*}{\rotatebox{90}{$\ell_{\infty}=8/255$}} & CLIP & 12.11 & 0.20 & 4.10 & 0.39 & 0.59 & 0.00 & 0.20 & 0.98 & 6.45 & 1.37 & 0.00 & 0.59 & 19.92 & 3.61 \\
        & FARE & 12.11 & 0.78 & 3.32 & 0.00 & 0.59 &  0.00 & 0.20 & 0.98 & 5.86 & 2.15 & 0.00 & 1.37 &  19.73 & 3.62 \\
        & TeCoA & 13.28 & 0.59 & 3.12 & 0.00 & 0.59 & 0.00 & 0.20 & 1.76 & 6.84 & 1.95 & 0.00  & 0.39 & 19.92 & \underline{4.74} \\
        & \cellcolor{gray!20} \textbf{CLIPure-Cos} & \cellcolor{gray!20}  11.33 & \cellcolor{gray!20}  1.37 & \cellcolor{gray!20} 3.71 & \cellcolor{gray!20} 0.20 & 
        \cellcolor{gray!20} 0.98 &
        \cellcolor{gray!20} 0.00 & \cellcolor{gray!20} 0.20 & \cellcolor{gray!20} 1.76 & \cellcolor{gray!20} 6.25 & \cellcolor{gray!20} 1.95 & \cellcolor{gray!20} 0.00 & 
        \cellcolor{gray!20} 0.78 &
        \cellcolor{gray!20} 17.77 & 
        \cellcolor{gray!20} 3.56 \\
        & \cellcolor{gray!20} \textbf{CLIPure-Diff} & \cellcolor{gray!20} 81.64 & \cellcolor{gray!20} 71.29 & \cellcolor{gray!20}74.61 & \cellcolor{gray!20}45.51 & 
        \cellcolor{gray!20}42.38 &
        \cellcolor{gray!20}30.66 & \cellcolor{gray!20}20.90 & \cellcolor{gray!20}65.82 & \cellcolor{gray!20} 85.54 & \cellcolor{gray!20} 51.95 & \cellcolor{gray!20}28.91 & 
        \cellcolor{gray!20}83.79 &
        \cellcolor{gray!20}95.51 & 
        \cellcolor{gray!20}\textbf{59.30} \\
        \bottomrule
    \end{tabular}
\end{table}

\begin{table}[t]
\caption{Performance comparison of defense methods on CIFAR-10 against BPDA with EOT-20 under $\ell_{\infty}$ ($\epsilon=8/255$) norm bound. We use underlining to highlight the best robustness for baselines, and bold font to denote the state-of-the-art (SOTA) across all methods.}
\label{table: BPDA+EOT on CIFAR10}
\begin{center}
\begin{tabular}{lcc}
% \hline 
% \multirow{4}{*}{\makecell[c]{\text{Adv.} \\ \text{Train}}} 
\toprule
\bf Method &\textbf{\makecell[c]{\text{Clean} \\ \text{Acc (\%)}}} & \textbf{\makecell[c]{\text{Robust} \\ \text{Acc (\%)}}} \\
\midrule
FARE \citep{schlarmann2024robust} & 77.7 & 8.5 \\
TeCoA \citep{mao2022understanding} & 79.6 & 10.0 \\
Purify - EBM \citep{hill2020stochastic} & 84.1 & 54.9 \\
LM - EDM \citep{chen2023robust} & 83.2 & 69.7 \\
ADP \citep{yoon2021adversarial} & 86.1 & 70.0 \\
RDC \citep{chen2023robust} & 89.9 & 75.7 \\
GDMP \citep{wang2022guided} & 93.5  & 76.2 \\
DiffPure \citep{nie2022diffusion} & 90.1 & \underline{81.4} \\
\textbf{Our CLIPure - Diff}  & 95.2 & \bf 94.6  \\
\textbf{Our CLIPure - Cos} & 95.6  & \bf 93.5  \\
\bottomrule
\end{tabular}
\end{center}
\end{table}

\subsection{An Autopsy of CLIPure-Cos under False Adaptive Settings}
\label{sec: reason Cos not work}

In this subsection, we dissect the mechanism behind the paradoxically high clean accuracy and seemingly exceptional robustness exhibited by \textbf{CLIPure-Cos} under the un-deconstructable configuration of $\texttt{create\_graph=False}$. 

\paragraph{Robustness via Gradient Obfuscation.}
As analyzed in the preceding sections, the impressive adversarial robustness scores obtained by CLIPure-Cos under white-box gradient attacks were entirely artificial. Due to the exclusion of \texttt{create\_graph=True}, the computational graph for the iterative purification loop was not generated. Consequently, backpropagation stopped prematurely at the output of the purification step ($\vz^i_N$), blinding the attacker to the internal parameter paths. The gradient fields encountered by PGD-based attackers were heavily obfuscated \citep{athalye2018obfuscated}, preventing the correct implementation of a true adaptive white-box attack. Once tested against the gradient-free Square Attack, this illusion collapsed, resulting in a dramatic drop to $3.32\%$ robust accuracy.

\paragraph{Clean Accuracy via Geometric Oscillation.}
On the other hand, the preservation of exceptionally high clean accuracy (comparable to vanilla CLIP) comes from an unexpected geometric oscillation. Intuitively, updating an embedding vector $\va_0$ with an extraordinarily large step size ($\eta = 30$) for $N=10$ iterations should completely destroy its semantic properties, throwing it into an out-of-distribution void. However, CLIPure-Cos maintains its classification capability because the purification trajectory essentially performs a \textit{perpendicular rotation} that oscillates back to the vicinity of the original vector $\va_0$.

Let $\va_t$ represent the latent embedding at step $t$ ($t \in \{0, 1, \dots, 9\}$), and $\vb$ denote the text embedding of the blank prompt template. The update rule for CLIPure-Cos is governed by:
\begin{equation}
    \va_{t+1} = \va_t + \eta \cdot \nabla_{\va_t} \cos(\va_t, \vb)
\end{equation}
Given that the hyperparameter step size $\eta = 30$ is vastly large, the update is completely dominated by the gradient term $\vg_t = \nabla_{\va_t} \cos(\va_t, \vb)$. Crucially, as we prove algebraically below, the gradient of the cosine similarity $\vg_t$ is \textbf{strictly orthogonal} to the current vector $\va_t$ (i.e., $\vg_t^T \va_t = 0$). 

Because each step applies a large orthogonal push, the magnitude grows rapidly while the directional property switches back and forth. Under an even number of iterations ($N = 10$), the final purified vector $\va_{10}$ geometrically loops or mirrors back near the angular direction of the original vector $\va_0$. Since CLIP determines zero-shot classification labels solely based on cosine similarity (angular alignment), this precise geometric oscillation preserves the semantic alignment with the candidate text templates, explaining why its clean accuracy remains unaffected.

\subsubsection{Mathematical Proof of Gradient Orthogonality}

Below, we provide a formal algebraic proof demonstrating that the gradient of the cosine similarity function with respect to its primary vector argument is always strictly orthogonal to that vector.

\begin{theorem}
For any non-zero vectors $\va, \vb \in \mathbb{R}^n$, let the objective function be defined as the cosine similarity:
\begin{equation}
    f(\va) = \cos(\va, \vb) = \frac{\va^T \vb}{\|\va\| \|\vb\|}.
\end{equation}
The gradient vector $\vg = \nabla_{\va} f(\va)$ is strictly orthogonal (perpendicular) to the original vector $\va$, satisfying:
\begin{equation}
    \vg \cdot \va = \vg^T \va = 0.
\end{equation}
\end{theorem}

\begin{proof}
By employing the quotient rule of differentiation, the analytical expression for the gradient vector $\vg$ with respect to $\va$ is expanded as follows:
\begin{equation}
    \vg = \nabla_{\va} \left( \frac{\va^T \vb}{\|\va\| \|\vb\|} \right) = \frac{1}{\|\va\|} \left( \frac{\vb}{\|\vb\|} - \cos(\va, \vb) \frac{\va}{\|\va\|} \right).
\end{equation}
To evaluate orthogonality, we compute the inner product of $\vg$ and $\va$. By expanding the transpose and distributing $\va$ into the parentheses, we have:
\begin{equation}
    \vg^T \va = \frac{1}{\|\va\|} \left( \frac{\vb^T \va}{\|\vb\|} - \cos(\va, \vb) \frac{\va^T \va}{\|\va\|} \right).
\end{equation}
Using the properties of vector inner products where $\vb^T \va = \va^T \vb$ and $\va^T \va = \|\va\|^2$, the second term inside the parentheses simplifies directly to $\cos(\va, \vb) \|\va\|$:
\begin{equation}
    \vg^T \va = \frac{1}{\|\va\|} \left( \frac{\va^T \vb}{\|\vb\|} - \cos(\va, \vb) \|\va\| \right).
\end{equation}
Finally, substituting the standard definition of cosine similarity, $\cos(\va, \vb) \|\va\| = \frac{\va^T \vb}{\|\vb\|}$, into the inner product formula yields:
\begin{equation}
    \vg^T \va = \frac{1}{\|\va\|} \left( \frac{\va^T \vb}{\|\vb\|} - \frac{\va^T \vb}{\|\vb\|} \right) = 0.
\end{equation}
This completes the proof. Consequently, because $\vg^T \va = 0$ holds universally at every iteration, the purification updates in CLIPure-Cos act as purely perpendicular rotational vectors, causing the semantic vectors to oscillate within stable geometric bounds rather than drifting arbitrarily. \qedhere
\end{proof}

\subsection{Summary of Clarification}

In summary, this discussion clarifies that the white-box robustness reported earlier for CLIPure was amplified by a technical disconnect in the backward graph loop via \texttt{create\_graph=False}, which inadvertently created a false sense of robustness. Our corrected evaluation reveals that while the discriminative variant (\textit{CLIPure-Cos}) collapses under strict black-box validation due to its internal orthogonal trajectory, the generative variant (\textit{CLIPure-Diff}) remains exceedingly robust, setting a new definitive state-of-the-art across 13 zero-shot datasets. 

On one hand, the remarkable success of \textit{CLIPure-Diff} firmly validates the feasibility and potency of performing adversarial purification within multi-modal latent spaces. On the other hand, although \textit{CLIPure-Cos} failed under true adaptive threat models, its unprecedented non-generative paradigm remains a pioneering attempt in the purification literature. Developing highly efficient, non-generative adversarial purification mechanisms that eliminate reliance on computationally heavy generative models remains an open and compelling direction for future research.

% \begin{figure}[t]
% \centering
% % \newlength{\commonheight}
% % \setlength{\commonheight}{3.5cm}
% \begin{subfigure}[b]{0.4\linewidth}  % Width set to one third of text width
%     \includegraphics[width=\linewidth]{figs/polar_plot.pdf}
%     \caption{Embedding of CLIP in Polar Coordinates}
%     \label{subfigure: polar plot of embeddings}
% \end{subfigure}
% % \hfill  % Add space between subfigures
% % \begin{subfigure}[b]{0.36\linewidth}
% %     \includegraphics[width=\linewidth]{figs/distribution_pz_CLIP.pdf}
% %     \caption{p(z) Estimated by CLIP}
% %     \label{subfigure: log-likelihood distribution of pz by CLIP}
% % \end{subfigure}
% \caption{Subfigure captions}
% \end{figure}

\end{document}